\documentclass{article}
\usepackage{pdflscape}

\usepackage{arxiv}

\usepackage[utf8]{inputenc} 
\usepackage[T1]{fontenc}    
\usepackage{url}            
\usepackage{booktabs}       
\usepackage{amsfonts}       
\usepackage{nicefrac}       
\usepackage{microtype}      
\usepackage{multirow}
\usepackage{xcolor}
\usepackage{graphicx}
\usepackage{orcidlink}
\usepackage{comment}
\usepackage{tcolorbox}
\newtcolorbox{promptbox}[1][]{%
  colback=gray!5,
  colframe=gray!50,
  fonttitle=\bfseries\small,
  title={#1},
  boxrule=0.5pt,
  arc=2pt,
  left=6pt,
  right=6pt,
  top=4pt,
  bottom=4pt,
  fontupper=\small\ttfamily
}
\usepackage{array}
\usepackage{todonotes}
\usepackage{float}
\usepackage{adjustbox}
\usepackage{csvsimple}
\newcolumntype{R}[1]{>{\raggedleft\arraybackslash}p{#1}}
\newcolumntype{L}[1]{>{\raggedright\arraybackslash}p{#1}}

\usepackage[authoryear,round,sort]{natbib}
\setcitestyle{authoryear,round,aysep={, },citesep={; },yysep={, }}
\usepackage{longtable}
\usepackage{caption}

\usepackage{titlesec}

\usepackage{placeins}


\usepackage{amsmath}
\usepackage{rotating}
\usepackage{cleveref}

\makeatletter
\renewcommand{\footnotesize}{\@setfontsize\footnotesize{8pt}{10pt}}
\makeatother

\makeatletter
\renewcommand{\scriptsize}{\@setfontsize\scriptsize{7pt}{9pt}}
\makeatother

\definecolor{VUB_blauw}{rgb}{0.1529, 0.2667, 0.5529}
\usepackage{hyperref}       
\hypersetup{
    colorlinks,%
    citecolor=VUB_blauw,%
    filecolor=VUB_blauw,%
    linkcolor=VUB_blauw,%
    urlcolor=VUB_blauw
}
\graphicspath{ {figures/} }
\newcommand{\customCor}[1]{%
  \includegraphics[height=1em]{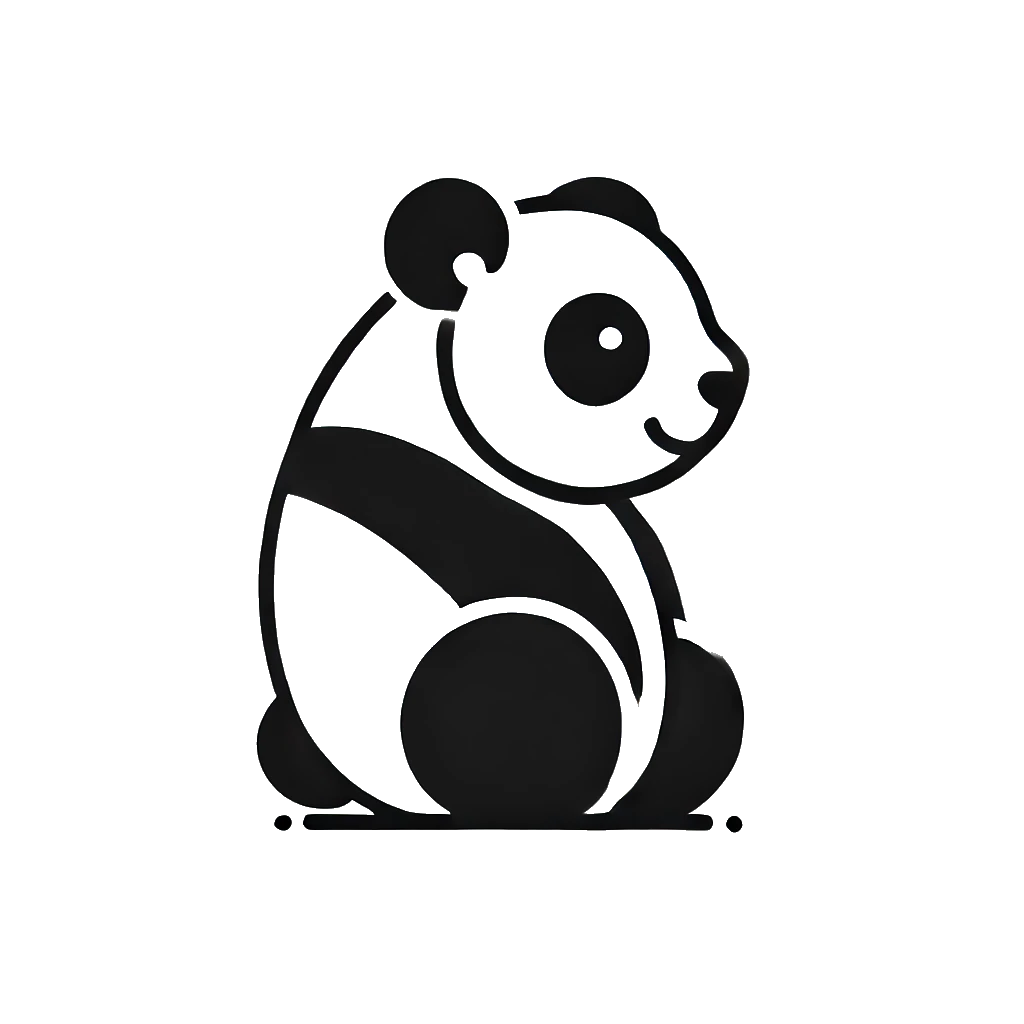} #1%
}

\AddToHook{shipout/background}{%
  \ifnum\value{page}=1
    \pagenumbering{Roman}
    \setcounter{page}{1}
  \fi
  \ifnum\value{page}=2
  \pagenumbering{arabic}
  \setcounter{page}{2}
  \fi
}

\title{Thinking Like a Scientist? A Structural Study of LLM-Generated Research Methods}
\runningtitle{}

\author{
 Francesca Carlon\textsuperscript{1, 2, \customCor{ }} \\
    \orcidlinkc{0009-0004-2152-2745} \And
  Brecht Verbeken\textsuperscript{1,2}\\
  \orcidlinkc{0000-0002-7506-3298}\\
  \And
    Vincent Ginis\textsuperscript{1,2,3} \\
 \orcidlinkc{0000-0003-0063-9608} \And
      Andres Algaba\textsuperscript{1,2} \\
 \orcidlinkc{0000-0002-0532-3066}\\
 \AND
  \textsuperscript{1}Data Analytics Lab, Vrije Universiteit Brussel, Pleinlaan 5, 1050 Brussels, Belgium \\
  \textsuperscript{2}imec-SMIT, Vrije Universiteit Brussel, Pleinlaan 9, 1050 Brussels, Belgium\\
  \textsuperscript{3}School of Engineering and Applied Sciences, Harvard University, Cambridge, Massachusetts 02138, USA
}

\begin{document}
\maketitle
\renewcommand{\thefootnote}{}
\footnotetext{\includegraphics[height=1em]{panda2.png} Corresponding author: \href{mailto:francesca.carlon@vub.be}{francesca.carlon@vub.be} \\}
\renewcommand{\thefootnote}{\arabic{footnote}}
\thispagestyle{plain}

\begin{abstract}
Large Language Models (LLMs) are increasingly used to guide research methodology, yet their default methodological tendencies under minimal prompting remain unclear. Here, we prompt GPT-5.1, Gemini 3 Pro, and DeepSeek-V3.2 with an LLM-extracted research question from each of 1,000 recent arXiv computer-science papers and compare the resulting methodology suggestions against a paper-derived experimental inventory. Since we provide only the research question, the differences we measure reflect initial suggestions and not how optimal those suggestions are. We extract structured method features from both sources, map them into a shared taxonomy, and quantify divergence across multiple taxonomy dimensions including model provider, dataset task type, and evaluation metric type. The strongest imbalance appears in provider choice, with Jensen--Shannon divergence about 3--5$\times$ larger than any other taxonomy dimension. Other/Academic single-occurrence models are underrepresented by 23--24 percentage points, while reused academic/community models are slightly overrepresented (4--6~pp). LLMs also suggest a much narrower range of methods overall: the effective number of model entities contracts from 1,232 to 59--96, and inter-LLM rank correlations ($0.55$--$0.68$) generally exceed LLM-to-paper correlations ($0.33$--$0.56$), so the distortions are largely shared across models. Popularity baselines, BM25 retrieval calibration, and paper-level similarity tests confirm that the outputs are query-specific responses, but filtered through a narrower set of options. Researchers who rely on LLM suggestions without cross-checking therefore risk narrowing their methodological search space toward a more concentrated default.
\end{abstract}


\section{Introduction}
\label{sec:Introduction}

\begin{figure}[!htbp]
\centering
\includegraphics[width=1\textwidth]{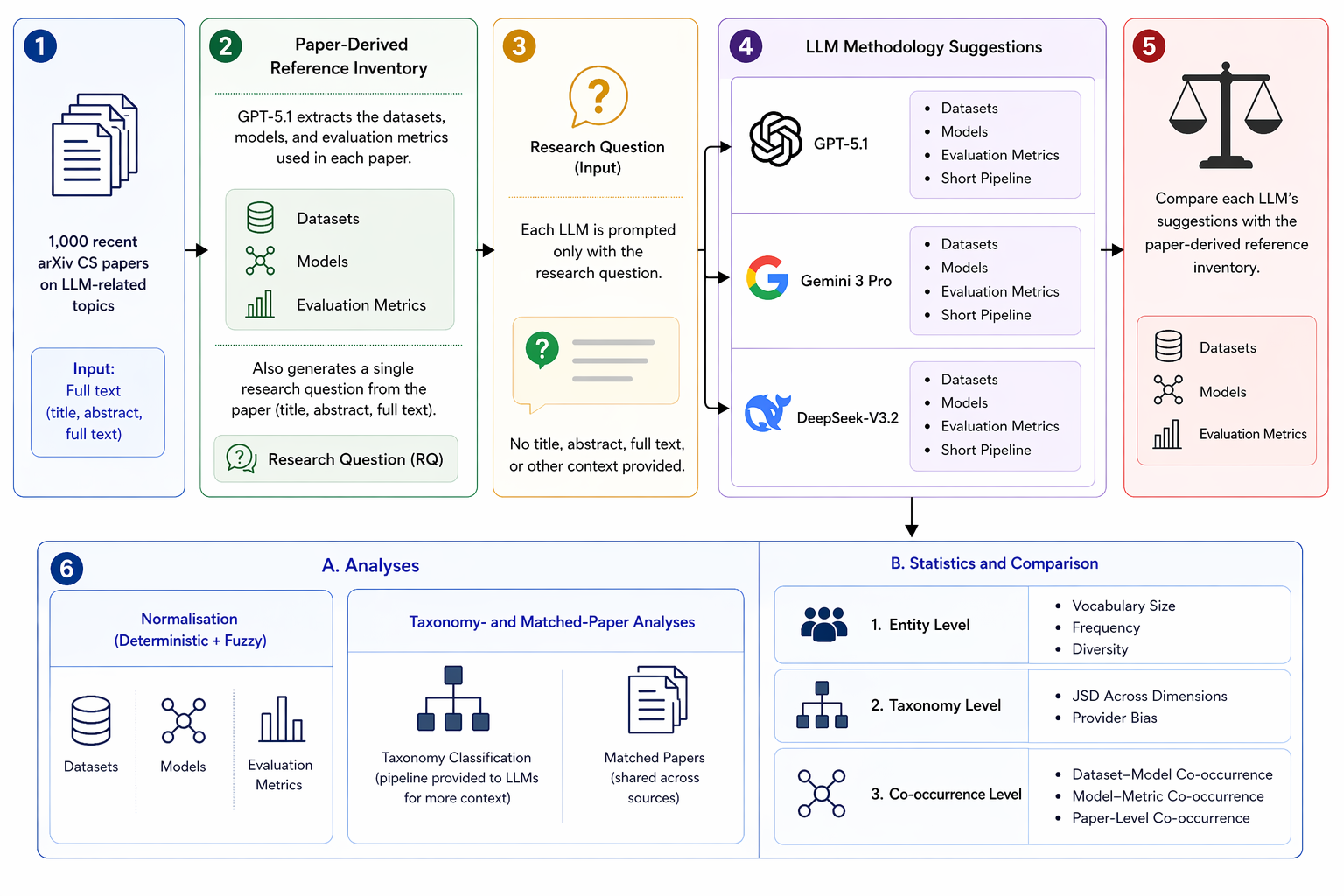}
\caption{\textbf{Overview of the study design.} We compare methodology suggestions from three LLMs, each prompted only with a research question, against paper-derived reference inventories for 1,000 recent arXiv computer-science papers on LLM-related topics. For each paper, GPT-5.1 extracts the datasets, models, and evaluation metrics actually used (the paper-derived reference inventory) and generates a single research question from the title, abstract, and full text. Each of GPT-5.1, Gemini~3~Pro, and DeepSeek-V3.2 then proposes datasets, models, metrics, and a short pipeline from that question alone. Normalised entity-name analyses use the full 1,000-paper outputs per source, while taxonomy- and matched-paper analyses use smaller classified or shared-paper subsets reported in the relevant captions and tables. This design isolates first-pass recommendation behaviour under deliberately sparse, research-question-only input.}
\label{fig:overview}
\end{figure}

Large Language Models (LLMs) are increasingly integrated into scientific workflows, not only as writing assistants~\citep{Lund_2023,10.1007/978-3-031-66329-1_42} but as tools that may reshape scientific practice more broadly~\citep{musslick2025automating}. Researchers now consult LLMs for paper feedback~\citep{liang2024can}, research-idea and hypothesis generation~\citep{si2025llmsgeneratenovelresearch,Baek2025ResearchAgent,qi2023largelanguagemodelszero}, knowledge synthesis and literature-guided research support~\citep{liao2024llmsresearchtoolslarge,skarlinski2024language}, and experimental design across natural language processing, computer vision, chemistry, materials
science, social science, and others~\citep{liao2024llmsresearchtoolslarge, 10.1007/978-3-031-66329-1_42, liang2024can, si2025llmsgeneratenovelresearch, Baek2025ResearchAgent, qi2023largelanguagemodelszero, boiko2023autonomous, skarlinski2024language, hewitt2024predicting, manning2024automated, kusumegi2025scientific, fortunato2018science}. Recent evidence suggests that scientific production itself is already shifting in response to LLM adoption~\citep{kusumegi2025scientific}. Several systems now automate larger parts of the research cycle, from autonomous idea generation and scientific discovery~\citep{lu2024aiscientistfullyautomated,yamada2025ai,mitchener2025kosmosaiscientistautonomous,gottweis2025towards,Elbadawi2024AIResearch} to multi-agent research workflows~\citep{li2024mlrcopilotautonomousmachinelearning,villaescusanavarro2025denarioprojectdeepknowledge,schmidgall-etal-2025-agent,li-etal-2025-chain-ideas,wang-etal-2024-scimon,gridach2025agenticaiscientificdiscovery,liu2025visionautoresearchllm}, literature screening~\citep{delgado2025transforming}, and research-code or methodology generation~\citep{gandhi2025researchcodeagentllmmultiagentautomated,novikov2025alphaevolve}.

The science-of-science question these developments raise is not whether LLMs answer correctly, but which methods they make salient by default. When a researcher asks a frontier model for first-pass guidance on datasets, models, or evaluation metrics, the resulting suggestions define an initial menu of methodological options that may anchor downstream decisions~\citep{musslick2025automating,fortunato2018science}. If that menu is systematically compressed or skewed, it could reshape the distribution of experimental designs across a field before deeper literature review even begins.

Prior work already suggests that reinforcement learning from human feedback can reduce output diversity~\citep{kirk2023understanding,luo2026quest}, that LLM-generated scholarship can reproduce unequal scientific recognition and citation patterns~\citep{liu2025unequal,algaba2025large,algaba2025deeplargelanguagemodels,mobini2026structurally}, and that LLM use can contribute to benchmark saturation and homogenization in downstream evaluation and creative tasks~\citep{ballon2026benchmarks,bommasani2022picking,doshi2024generative,anderson2024homogenization}. These concerns matter because science already operates under well-documented pressures around citation inequality and reproducibility~\citep{nielsen2021global,ioannidis2005most,open2015estimating,camerer2016evaluating}. However, it remains unclear how LLMs reshape methodological attention in a contemporary research field, and which specific providers, model families, benchmarks, and evaluation criteria become more or less salient when frontier models serve as first-pass intermediaries.

Here, we study one such case: recent computer-science papers on LLM-related topics, where researchers are especially likely to use LLMs to choose benchmarks, model families, and evaluation criteria. We present a large-scale empirical comparison of methodology suggestions from GPT-5.1 (documented in the GPT-5 system card; exact model ID reported in \Cref{sec:gtextraction})~\citep{singh2026openai}, Gemini~3~Pro~\citep{gemini3pro_modelcard}, and DeepSeek-V3.2~\citep{deepseekai2025v32} against paper-derived methodology inventories extracted from 1,000 recent arXiv papers. For each paper, we extract the datasets, models, and evaluation metrics used by the authors (which we refer to as the paper-derived reference inventory), generate a single research question from the paper's title, abstract, and full text, and ask each LLM to propose datasets, models, metrics, and a short pipeline from that question alone. We compare suggestions against the paper-derived inventories at three levels: individual entity frequencies, broad category distributions, and patterns of which methods appear together (\Cref{fig:overview}). Full methodological details are provided in \Cref{sec:Methods}. Ablations and robustness analyses are reported in \Cref{sec:Ablations,sec:StatisticalRobustness}.

We find two regularities. First, divergence concentrates in model provider. All three LLMs overweight a small set of major commercial providers, and under a broader regrouping the main deficit lies in the singleton-defined long tail rather than in reused academic or community models. Second, LLMs generate question-sensitive but sharply compressed method menus. Effective model diversity contracts by 13--21$\times$, inter-LLM rank correlations exceed LLM-to-paper correlations, and many exact-name misses collapse to family- or provider-level matches. We interpret divergence as redistribution of methodological attention relative to a paper-derived reference corpus, not deviation from a normative optimum, and we study these effects in a deliberately sparse regime where each model sees only a research question. Popularity baselines, BM25 calibration, and shuffled-paper tests rule out generic templating. Frontier LLMs respond to the question, but through a narrower and more provider-concentrated vocabulary.

\FloatBarrier

\section{Results}
\label{sec:Results}

We classify analyses by whether the taxonomy classifier sees only entity names [EL] or also the generated pipeline [WP]. Section~\ref{sec:taxonomy} uses the [EL] baseline, whereas Section~\ref{sec:cooccurrence} and the taxonomy-classified analyses in Sections~\ref{sec:established_only}--\ref{sec:provider_level1} use [WP]. Entity-name recall (Section~\ref{sec:established_only}) and normalisation-threshold sensitivity (Section~\ref{sec:normalisation_sensitivity}) are branch-independent. Section~\ref{sec:annotation_validation} contains branch-independent audits of extraction, normalisation, and introducedness, plus a [WP] audit of taxonomy labels; provider reliability is taken to apply to both [EL] and [WP] because provider labels barely move across branches (\Cref{sec:pipeline_ablation}). Figures based on normalised entity names (\Cref{fig:diversity_coverage,fig:entity}) precede classification and carry no branch tag. Provider remains the most divergent dimension in both settings (\Cref{fig:jsd_provider}a, \Cref{tab:effect_sizes}). Co-occurrence matrices are unavailable in the [EL] branch because it stores only aggregate category counts.

\subsection{Question-conditioned but compressed method menus}
\label{sec:entity}

\begin{table}[t]
\centering
\caption{\textbf{LLMs operate with a substantially reduced model and metric vocabulary while comparatively preserving more dataset diversity.} Computed from deterministically normalised entity inventories after default fuzzy clustering ($T = 90$) for 1,000 papers per source. Unique entity names, total mentions, and the ratio of unique names to total mentions for the paper-derived reference inventory and each LLM. Total mentions exceed the number of papers because each paper may contain multiple entities of each type.}
\label{tab:vocab}
\begin{adjustbox}{max width=\textwidth}
\begin{tabular}{ll R{1.7cm} R{1.7cm} R{1.7cm}}
\toprule
\textbf{Entity type} & \textbf{Source} & \textbf{Unique names} & \textbf{Total mentions} & \textbf{Unique/total (\%)} \\
\midrule
\multirow{4}{*}{Datasets}
 & Reference & 2,656 & 4,007 & 66.3 \\
 & GPT-5.1 & 1,598 & 3,225 & 49.6 \\
 & Gemini~3~Pro & 1,659 & 3,056 & 54.3 \\
 & DeepSeek-V3.2 & 1,211 & 2,218 & 54.6 \\
\addlinespace
\multirow{4}{*}{Models}
 & Reference & 3,546 & 8,361 & 42.4 \\
 & GPT-5.1 & 523 & 3,285 & 15.9 \\
 & Gemini~3~Pro & 594 & 3,349 & 17.7 \\
 & DeepSeek-V3.2 & 387 & 2,529 & 15.3 \\
\addlinespace
\multirow{4}{*}{Metrics}
 & Reference & 2,453 & 5,232 & 46.9 \\
 & GPT-5.1 & 943 & 4,125 & 22.9 \\
 & Gemini~3~Pro & 990 & 3,357 & 29.5 \\
 & DeepSeek-V3.2 & 761 & 2,875 & 26.5 \\
\bottomrule
\end{tabular}
\end{adjustbox}
\end{table}

\begin{figure}[!htbp]
\centering
\includegraphics[width=1\textwidth]{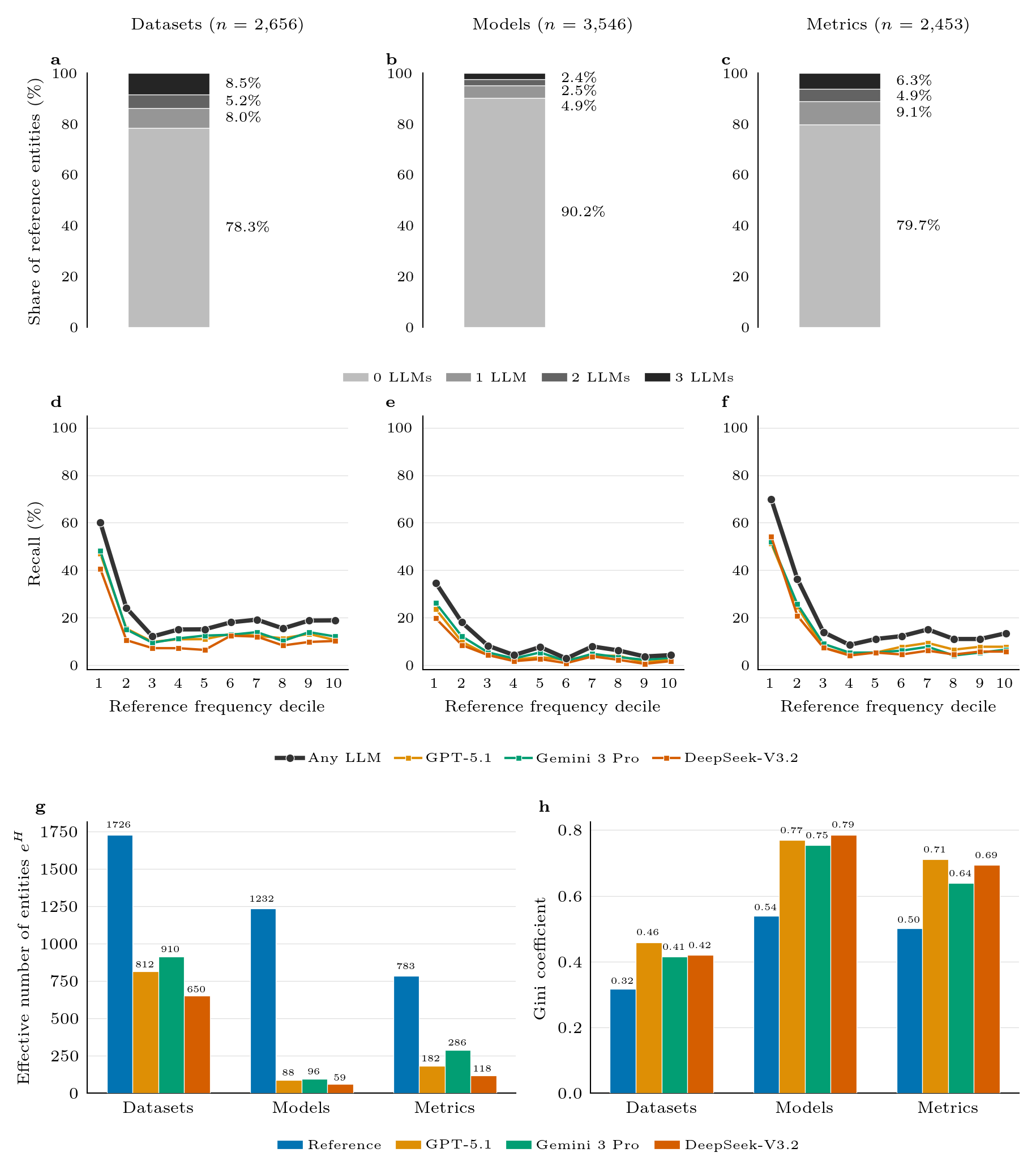}
\caption{\textbf{LLMs cover only a small fraction of the entities in paper-derived reference inventories and do so through markedly more concentrated distributions.} Panels use normalised entity inventories (\Cref{sec:normalisation}) from 1,000 papers per source. (\textbf{a}--\textbf{c})~Share of reference-inventory entities covered by 0, 1, 2, or all 3 LLMs. (\textbf{d}--\textbf{f})~Recall of reference-inventory entities by frequency decile (decile~1 = most frequent) for datasets, models, and metrics. (\textbf{g})~Effective number of entities ($\exp(H)$) for each source across entity types. (\textbf{h})~Gini coefficient measuring frequency inequality. The coverage gap is heavily concentrated in the long tail, with around 90\% of reference models receiving no LLM coverage.}
\label{fig:diversity_coverage}
\end{figure}

We first compare individual dataset, model, and evaluation metric names between the paper-derived reference inventory and the LLM suggestions. Unless otherwise noted, entity-name analyses use deterministic normalisation followed by fuzzy clustering at the default threshold $T = 90$ (\Cref{sec:normalisation}), and \Cref{tab:vocab} summarises vocabulary size and total mentions by entity type. The clearest pattern is vocabulary compression. LLMs operate with an order-of-magnitude smaller model vocabulary than the reference inventory (\Cref{tab:vocab,tab:normalisation_sensitivity}), and coverage is correspondingly sparse. Roughly four out of five reference datasets and metrics, and nine out of ten reference models, are not suggested by any LLM (\Cref{fig:diversity_coverage}a--c). These vocabulary totals are pipeline-specific estimates rather than exact counts, because the blinded audit finds only moderate agreement on extraction (inter-model ICC$(2,1) = 0.469$) and the perfect precision/recall/F1 in \Cref{tab:annotation_extraction} applies only to the consensus subset. LLMs also introduce many corpus-novel entities, so they simultaneously omit entities present in full paper inventories and substitute alternatives not used in the target literature.

\begin{figure}[!htbp]
\centering
\includegraphics[width=\textwidth]{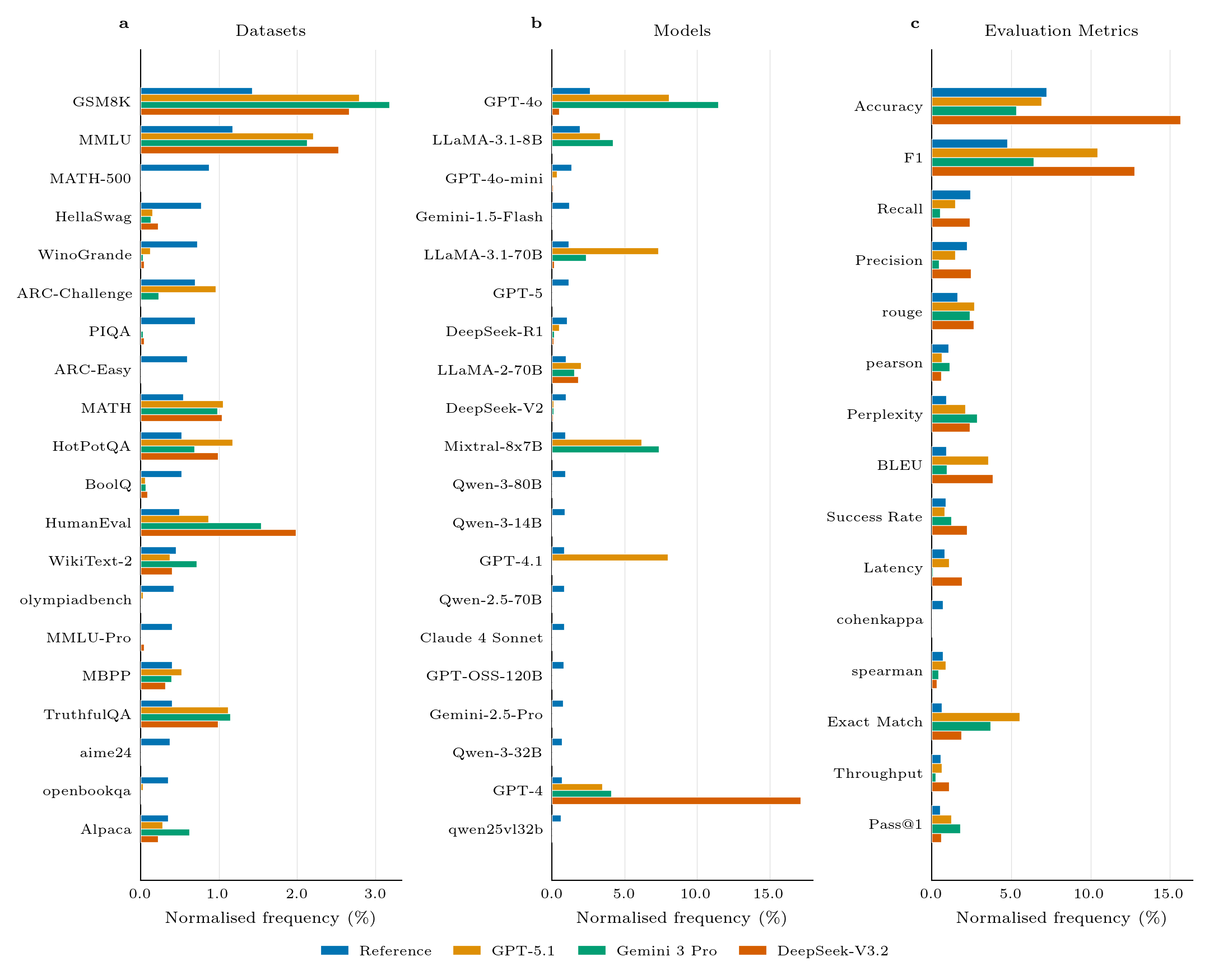}
\caption{\textbf{The top-ranked datasets, models, and metrics in paper-derived reference inventories are overrepresented in LLM suggestions, while the long tail is truncated.} Computed from normalised entity inventories (\Cref{sec:normalisation}) aggregated over 1,000 papers per source. Normalised frequency (\%) of the top-20 datasets (\textbf{a}), models (\textbf{b}), and top-15 evaluation metrics (\textbf{c}), sorted by reference-inventory frequency. Bars show the reference inventory (blue), GPT-5.1 (orange), Gemini~3~Pro (green), and DeepSeek-V3.2 (red). For models, exact-name comparisons should be interpreted with caution because the paper corpus (June--December 2025) partly postdates the training-data cutoffs of the evaluated LLMs. The figure makes visible how LLM suggestions cluster on well-known benchmarks and major commercial models while under-covering the long tail present in full paper-level inventories.}
\label{fig:entity}
\end{figure}

Information-theoretic diversity measures show that this gap reflects concentration as well as missing coverage. The effective number of model entities contracts by $13$--$21\times$ between the reference inventory and LLM suggestions, a reduction that is stable across fuzzy-clustering thresholds (\Cref{fig:diversity_coverage}g,h; \Cref{tab:normalisation_sensitivity}), and Gini coefficients rise in the same direction, indicating heavy reliance on a small set of names. Coverage is also strongly frequency-dependent. The union of all three LLMs recovers most top-decile datasets and metrics but only a small fraction of bottom-decile entities (\Cref{fig:diversity_coverage}d--f). Among shared entities, log-log regressions of the LLM-to-reference frequency ratio on reference rank yield significantly positive slopes (all $p < 0.001$), indicating that LLMs amplify the rarer entities they do cover. Representative entity-level examples are shown in \Cref{fig:entity}.

Distortions are correlated across the three systems. Inter-LLM Spearman correlations reach $0.55$--$0.68$, generally exceeding reference-to-LLM correlations of $0.33$--$0.56$ (\Cref{tab:inter-llm-agreement}), so the three LLMs compress the methodological vocabulary in similar ways rather than making independent errors. Because GPT-5.1 is also used for research-question generation, paper-side entity extraction, and taxonomy classification (\Cref{sec:Methods}), reference--GPT-5.1 overlap comparisons are not independent and may overstate GPT-5.1's apparent closeness to the reference corpus relative to Gemini~3~Pro and DeepSeek-V3.2. The inter-LLM comparisons do not share this confound.

Compression does not, however, imply fixed output. Paper-level similarity tests (\Cref{sec:content_sensitivity}) show that same-paper taxonomy similarity between the reference inventory and LLM suggestions significantly exceeds shuffled baselines ($p < 0.001$ across all nine entity type $\times$ LLM combinations), with effect sizes spanning $0.075$--$0.266$ from models to metrics (\Cref{tab:content_sensitivity}). Suggestions are conditioned on individual research questions, but each question is answered through a narrower vocabulary than the paper ecosystem provides.

\begin{table}[t]
\centering
\caption{\textbf{Inter-LLM rank correlations generally exceed paper-to-LLM correlations, indicating shared distortions rather than independent errors.} Computed from normalised entity-count series (\Cref{sec:normalisation}) aggregated over 1,000 papers per source. Spearman rank correlation~\citep{spearman1904proof} ($\rho$) is computed on shared support (the intersection of entities with non-zero counts in both sources) and Jaccard similarity~\citep{jaccard1912distribution} at $K{=}20$ for each pair of sources across datasets, models, and evaluation metrics. Reference--GPT-5.1 pairs are not fully symmetric because GPT-5.1 also generates the research question and the paper-side inventories (\Cref{sec:Methods}); these values may overstate GPT-5.1's closeness to the paper-derived reference inventory relative to Gemini~3~Pro and DeepSeek-V3.2, whereas inter-LLM pairs are unaffected.}
\label{tab:inter-llm-agreement}
\begin{adjustbox}{max width=\textwidth}
\begin{tabular}{l R{1.0cm} R{1.0cm} R{1.0cm} R{1.0cm} R{1.0cm} R{1.0cm}}
\toprule
 & \multicolumn{3}{c}{Spearman $\rho$} & \multicolumn{3}{c}{Jaccard@20} \\
\cmidrule(lr){2-4} \cmidrule(lr){5-7}
Pair & Datasets & Models & Metrics & Datasets & Models & Metrics \\
\midrule
Reference--GPT-5.1 & 0.51 & 0.56 & 0.54 & 0.25 & 0.21 & 0.48 \\
Reference--Gemini~3~Pro & 0.51 & 0.48 & 0.50 & 0.25 & 0.18 & 0.33 \\
Reference--DeepSeek-V3.2 & 0.54 & 0.33 & 0.54 & 0.18 & 0.05 & 0.48 \\
GPT-5.1--Gemini~3~Pro & 0.63 & 0.63 & 0.60 & 0.43 & 0.43 & 0.43 \\
GPT-5.1--DeepSeek-V3.2 & 0.67 & 0.58 & 0.60 & 0.29 & 0.14 & 0.48 \\
Gemini~3~Pro--DeepSeek-V3.2 & 0.68 & 0.57 & 0.55 & 0.25 & 0.25 & 0.54 \\
\bottomrule
\end{tabular}
\end{adjustbox}
\end{table}

We turn next to whether these shared distortions reflect systematic concentration in specific method categories (model provider, dataset modality, evaluation type) that entity-level analysis alone cannot reveal.

\FloatBarrier

\subsection{Provider concentration dominates taxonomy-level divergence}
\label{sec:taxonomy}

\begin{figure}[!htbp]
\centering
\includegraphics[width=\textwidth]{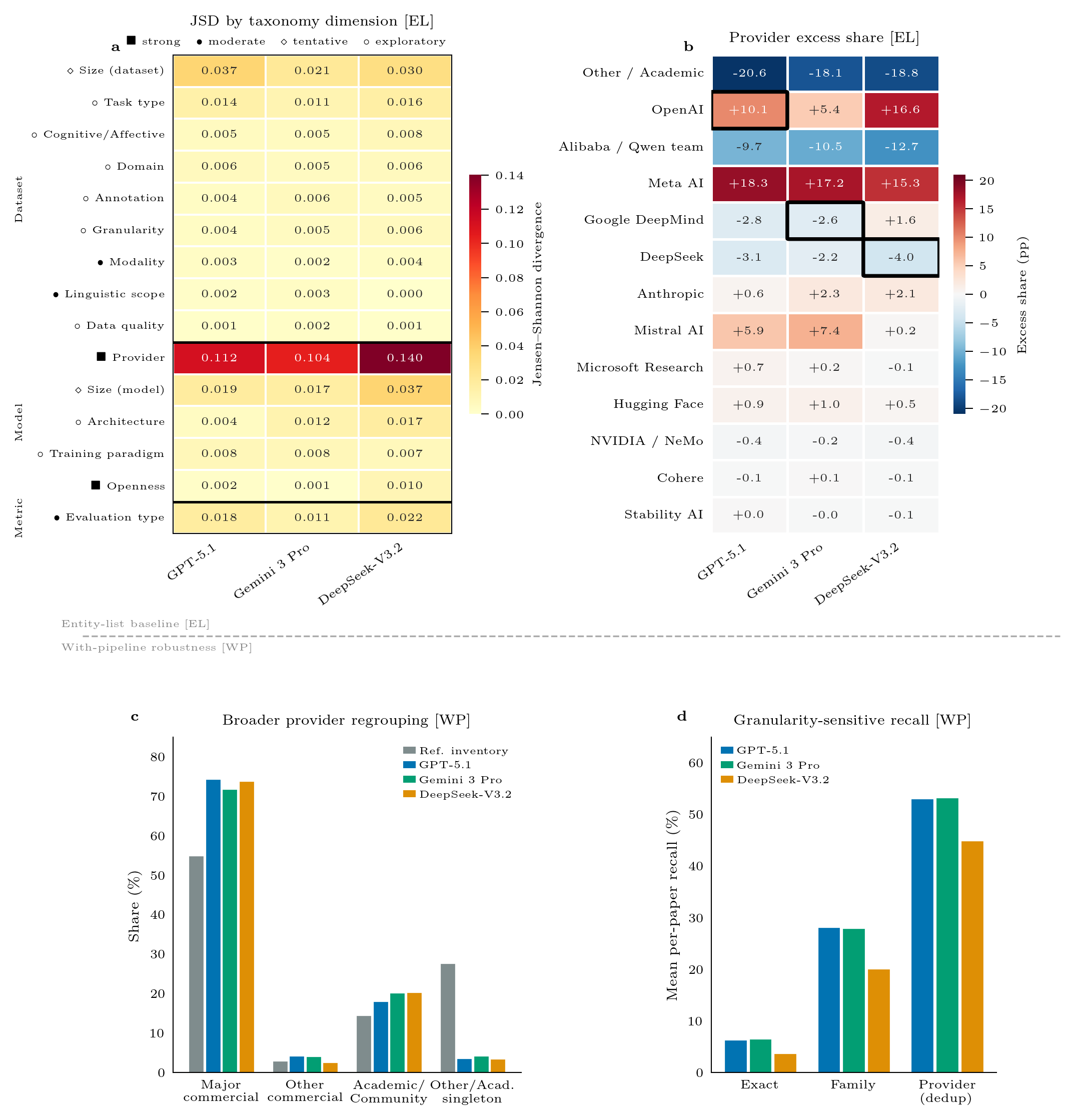}
\caption{\textbf{Model-provider choice is the single most divergent taxonomy dimension, with the deficit concentrated in the singleton-defined long tail rather than in reused academic/community models.} Composite figure combining one baseline branch with two robustness summaries; a dashed separator marks the branch switch. (\textbf{a}--\textbf{b})~[EL, mixed strong/moderate/tentative/exploratory dimensions]: (\textbf{a})~Jensen--Shannon divergence (JSD, base-2) between the reference inventory and each LLM for all 15 taxonomy dimensions, grouped by entity type; computed in the entity-list-only baseline using classified outputs from n$=1{,}000$ reference-inventory papers, 1{,}000 GPT-5.1 papers, 981 Gemini~3~Pro papers, and 998 DeepSeek-V3.2 papers. (\textbf{b})~Excess share (percentage-point difference from reference inventory) for each model provider $\times$ LLM. Black borders mark ``own provider'' cells; all three LLMs overrepresent Meta~AI and OpenAI while underrepresenting the aggregate Other/Academic category. (\textbf{c}--\textbf{d})~[WP, strong: provider]: (\textbf{c})~Provider distribution under a broader four-way provider regrouping (\Cref{sec:provider_level1}), computed on the all-three-LLM shared model-paper subset (n$=915$). (\textbf{d})~Mean per-paper model recall under exact-name, family-level, and provider-level matching (provider recall deduplicated per paper; \Cref{sec:family_matching}), on the same n$=915$ shared-paper subset. The substantial rise from exact to family and provider matching shows that many LLM misses are errors of granularity, not complete failures to recover the relevant model family.}
\label{fig:jsd_provider}
\end{figure}

We classify all entities into structured category schemes (\Cref{sec:taxonomy_methods}) and compare the resulting distributions across sources. Unless otherwise noted, the summary category results in this subsection use labels assigned from the entity lists alone and are based on the classified outputs available at analysis time (n$=1{,}000$ reference-inventory papers, 1{,}000 GPT-5.1 papers, 981 Gemini~3~Pro papers, 998 DeepSeek-V3.2 papers; the lower counts for Gemini~3~Pro and DeepSeek-V3.2 arise from classification failures during processing, and entries with broken or invalid classifications were excluded). We keep that entity-list-only baseline separate from the richer setting in which the classifier also sees the suggested pipeline. Appendix~B measures how much labels move when pipeline or web search are added, and Appendix~C recomputes robustness statistics under the richer labelling setting. Across all 15 category dimensions, provider is the dominant divergence. In the entity-list-only baseline (\Cref{fig:jsd_provider}a) and in the with-pipeline robustness setting alike, provider carries the largest JSD, about $3$--$5\times$ the next-largest, and also the largest Cram\'{e}r's $V$, though the Cram\'{e}r's $V$ margin is smaller (\Cref{tab:effect_sizes}). By contrast, low-JSD dimensions such as openness and data quality show much closer aggregate distributions, but only openness reaches the strong audit tier. Data quality remains exploratory because it falls below the $\kappa$ threshold.

As detailed in the blinded audit (\Cref{sec:annotation_validation}), the 15 taxonomy dimensions fall into four validation tiers. Strong dimensions (provider, openness) anchor our main claims. Moderate dimensions (modality, evaluation type, linguistic scope) meet the $\kappa \geq 0.5$ and accuracy $\geq 75\%$ thresholds with narrower margins, and tentative size is adequate overall but drops on the model-side LLM stratum, so size-dependent claims carry more uncertainty. All remaining dimensions are exploratory and provide descriptive context rather than definitive structural claims. Full tier statistics, $\kappa$ values, and accuracies are reported in \Cref{sec:annotation_validation} and \Cref{tab:annotation_classification}.

All three LLMs share nearly the same provider-concentration profile (\Cref{fig:jsd_provider}b). They overrepresent Meta~AI and OpenAI while underrepresenting the aggregate Other/Academic category and Alibaba/Qwen. In the paper-derived reference inventory, academic and independent models together account for roughly two-fifths of model mentions, and all three LLMs roughly halve that share while inflating Meta~AI (panel~b gives per-provider percentage-point deltas, \Cref{tab:effect_sizes} gives $\chi^2$ and effect sizes). The fine-grained Other/Academic category aggregates reused academic/community models with singletons that appear in only one paper. Under a broader four-way regrouping that separates these two groups, the deficit falls in the singleton-defined long tail, and reused academic/community models are modestly overrepresented (\Cref{fig:jsd_provider}c, \Cref{sec:provider_level1}). The deficit is therefore major-provider concentration plus long-tail suppression, not a blanket suppression of academic or community models. Only GPT-5.1 shows clear own-provider self-preferencing, so the shared concentration toward a small set of commercial providers is difficult to explain as simple self-promotion.

Most other dimensions are closer to the reference inventory, but the remaining deviations are directionally consistent. Dataset and model size rank next by JSD, and evaluation type shifts toward accuracy-like metrics while user-experience and efficiency metrics decline (\Cref{fig:taxonomy_appendix}, \Cref{tab:effect_sizes}). LLMs consistently overweight the top model-size bucket, though size labels are coarse analysis buckets rather than exact parameter-count bands, especially for closed models whose sizes are estimated from public information. In the with-pipeline robustness setting (\Cref{tab:effect_sizes}), provider remains dominant and size forms the next tier. Model architecture appears elevated in some summaries but stays exploratory and label-sensitive under the audit and ablation checks (\Cref{sec:annotation_validation,sec:websearch_ablation}).

The remaining dataset dimensions (domain, annotation type, linguistic scope, cognitive/affective properties, granularity, data quality) all exhibit low aggregate divergence (mean JSD $=$ 0.002--0.006), with small shifts toward self-supervised annotation, multilingual datasets, and away from education-domain content (\Cref{fig:taxonomy_appendix} caption). These deviations are an order of magnitude smaller than the provider distortions documented above.

The observed distributional divergences are substantially smaller than what a deterministic popularity baseline produces, and larger than what a popularity-proportional sampler produces. For provider, model size, and metric evaluation type alike, each LLM sits between a stochastic sampled baseline (near-zero JSD) and a deterministic top-$k$ baseline (one to two orders of magnitude higher JSD), always closer to the sampled floor than to the top-$k$ ceiling (\Cref{tab:popularity_baseline}). LLM suggestions therefore track paper-level content without collapsing to a fixed popularity ranking, even though they exhibit distributional biases that a popularity-proportional sampler would not. A complementary calibration using leave-one-out BM25 retrieval over generated research questions improves substantially over global popularity but remains more divergent than the LLMs on all three main dimensions (\Cref{tab:local_retrieval_baseline}). Equal-count comparisons that match entities per paper between reference and LLM preserve the hierarchy, with provider the dominant model dimension (\Cref{sec:cardinality_matched}). These results, together with the paper-level question-sensitivity tests in \Cref{sec:entity}, suggest that provider concentration is a systematic distributional property of LLM suggestions, not an artefact of fixed template outputs.

\FloatBarrier

\begin{figure}[!htbp]
\centering
\includegraphics[width=\textwidth]{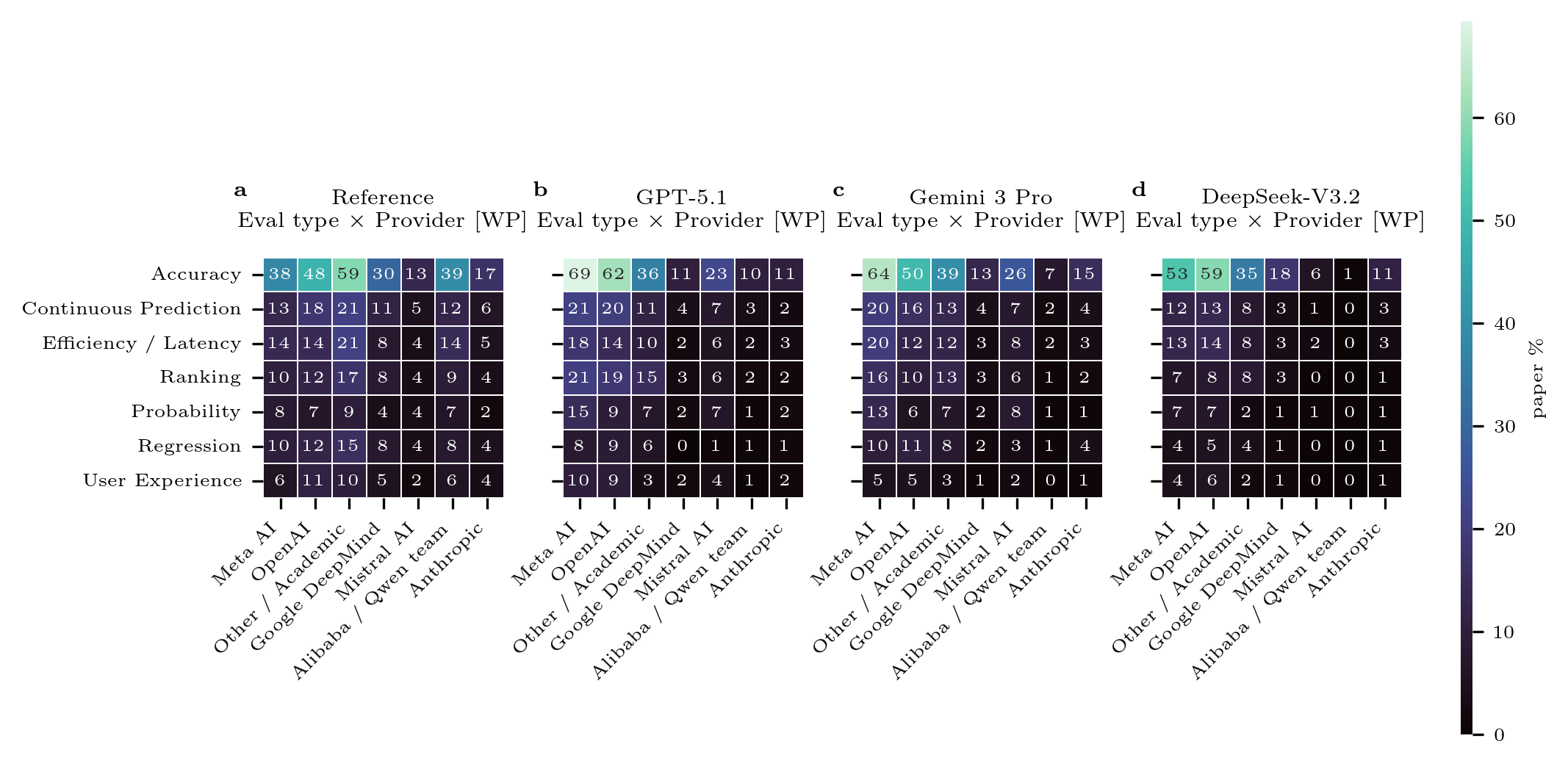}
\caption{\textbf{LLM suggestions shift evaluation-type $\times$ provider co-occurrences toward Meta~AI and OpenAI and away from the aggregated Other/Academic category~[WP].} Both axes are validated dimensions (provider: strong, 89.7--94.1\% accuracy, $\kappa = 0.923$--$1.000$; evaluation type: moderate, 89.7\% LLM accuracy, $\kappa = 0.588$), though evaluation-type reliability is asymmetric and rests on the LLM stratum (reference-stratum $\kappa = 0.061$ on $n = 17$ rows). Panels show the percentage of papers (within each source) in which a given evaluation-type row co-occurs with a given provider column, computed from the with-pipeline classification branch; source files contain n$=1{,}000$ reference-inventory papers, 1{,}000 GPT-5.1 papers, 981 Gemini~3~Pro papers, and 998 DeepSeek-V3.2 papers, with each panel restricted to papers with non-empty labels for both entity types. Heatmaps show the top 7 row and column categories. A complementary exploratory figure for dataset task type $\times$ model provider (task-type labels have only 10--32\% audit accuracy in the primary strata) is in \Cref{fig:cooccurrence_tasktype_appendix}. Among the reported co-occurrence analyses, this evaluation type $\times$ provider pair is the most interpretable because provider is strongly validated and evaluation type is moderately validated in the LLM stratum; reference-stratum reliability for evaluation type is the only weak point ($\kappa = 0.061$, $n = 17$).}
\label{fig:cooccurrence}
\end{figure}

\subsection{LLMs narrow provider-centred co-occurrence patterns}
\label{sec:cooccurrence}

Beyond frequency, we test whether LLMs reproduce method combinations. The analyses in this subsection use the with-pipeline classification branch (\Cref{sec:cooccurrence_methods}) and serve as a structural robustness check, not a direct numeric continuation of the entity-list-only category distributions in \Cref{sec:taxonomy}. \Cref{fig:cooccurrence} shows the evaluation type $\times$ model provider co-occurrence across sources. Evaluation-type patterns are broadly distributed across providers in the reference inventory, whereas in LLM suggestions they concentrate on Meta~AI and OpenAI, and the aggregated Other/Academic category declines. Much of that decline reflects the singleton-defined long tail identified in \Cref{sec:provider_level1}. Among the reported co-occurrence analyses, this pair is the most interpretable because provider is strongly validated and evaluation type is moderately validated in the LLM stratum; reference-stratum reliability for evaluation type is the only weak point (\Cref{sec:annotation_validation}). A complementary exploratory analysis of dataset task type $\times$ model provider is reported in \Cref{fig:cooccurrence_tasktype_appendix}; task-type labels have only 10--32\% audit accuracy in the primary strata and provide descriptive context rather than precise semantic claims.

Across all 12 taxonomy pairs, provider-based combinations show the largest structural divergence (\Cref{fig:cooccurrence_jsd_appendix}). Some architecture-based pairs also show non-trivial divergence, but architecture labels are exploratory and sensitive to classification context (\Cref{sec:annotation_validation,sec:websearch_ablation}), so those patterns are descriptive only. Openness-based pairs are the most faithfully preserved. Averaged across all 12 pairs, Gemini~3~Pro is closest to the reference inventory and DeepSeek-V3.2 farthest, though the ordering varies by pair.

Residual analysis that holds row and column totals fixed shows that LLMs do not invent an entirely new interaction structure, though they do sharpen it. Residual correlations for the focal provider pairs are moderate, and sign flips are rare (\Cref{tab:residual_summary}). LLMs therefore usually preserve the direction of associations while amplifying a narrower subset of provider-centred combinations. This narrowing is not uniform across entity types. The inter-paper Jaccard decomposition (\Cref{tab:content_sensitivity}) shows that most of the increased inter-paper overlap for datasets is accounted for by marginal frequency shifts, that the pattern for metrics is mixed, and that excess homogenisation beyond vocabulary compression is concentrated in model suggestions, where all three LLMs show significantly positive $\Delta_{\mathrm{excess}}$ (\Cref{sec:content_sensitivity}). The co-occurrence narrowing is therefore mainly a model and provider effect, not an equally strong flattening across all entity types, with the practical consequence that fewer provider-centred combinations remain salient on the suggested menu.

\FloatBarrier

\subsection{Robustness: granularity, long tails, and label reliability}
\label{sec:boundary}

Three complementary analyses sharpen our interpretation of the compression and concentration patterns documented above (\Cref{sec:established_only,tab:established_robustness,sec:provider_level1,tab:provider_level1,sec:family_matching,tab:family_matching}, \Cref{fig:robustness_dashboard}). They clarify where the measured divergence reflects granularity mismatches, where it reflects long-tail underweighting, and where the evidence is strongest.

Exact-name model recall is low (4--7\% per paper), but relaxed matching substantially improves it. Mean per-paper recall rises to 20--28\% at family level and 45--53\% at provider level (deduplicated per paper, \Cref{sec:family_matching}). These are per-paper robustness scores on the shared-paper model subset, not direct re-estimates of the raw 1,000-paper entity-frequency branch. They show that many apparent misses are errors of granularity, in which an LLM names a different variant from the same model family, not complete failures to recover the relevant model family.

Under the broader four-way provider regrouping (\Cref{sec:provider_level1}), the dominant deficit falls in the singleton-defined long tail (roughly $-$24~pp across LLMs), and reused academic/community models are modestly overrepresented ($+$4--6~pp). The pattern is therefore concentration on major commercial providers alongside long-tail suppression, not academic erasure. Removing singleton entities from the reference inventory improves recall but also removes many established methods. The singleton filter has only 7.5\% precision as a proxy for paper-specificity (\Cref{tab:annotation_heuristic_calibration,sec:annotation_validation}), so singleton exclusion is a long-tail sensitivity analysis rather than a principled fairness correction, and the compression remains large even after these adjustments.

Finally, the blinded cross-model audit (\Cref{sec:annotation_validation}) classifies the 15~taxonomy dimensions into four validation tiers. Provider and openness are strong. Modality, evaluation type, and linguistic scope are moderate, while size is tentative (adequate overall but weaker on the model-side LLM stratum). All remaining dimensions are exploratory and provide descriptive context rather than definitive structural claims. Full tier statistics are reported in \Cref{tab:annotation_classification,tab:annotation_summary}.

\FloatBarrier

\section{Discussion}
\label{sec:Discussion}

Two findings emerge from these analyses about how LLMs redistribute methodological attention under research-question-only prompting.

First, LLM suggestions are commercially concentrated and suppress Other/Academic singleton models. Provider divergence is about 3--5$\times$ larger than the next-largest taxonomy dimension, and the same pattern survives broader provider regrouping, popularity-baseline comparisons, BM25 retrieval calibration, and cross-model robustness checks in which Claude Opus~4.6 re-extracts entities from a 94-paper subset and re-classifies 200~papers' entities in place of GPT-5.1 (\Cref{sec:annotation_validation}). When providers are grouped into four broad categories, the shortfall concentrates in singleton Other/Academic models (those appearing in exactly one paper in the reference corpus, $-$23.5 to $-$24.3~pp), while established academic and community models are modestly overrepresented ($+$3.6 to $+$5.8~pp). LLMs therefore favour well-known providers rather than suppressing non-commercial work across the board. The concentration also propagates to provider-centred method combinations. Evaluation type $\times$ provider is the most interpretable co-occurrence pair among those reported because provider is strongly validated and evaluation type is moderately validated in the LLM stratum (LLM $\kappa = 0.588$, 89.7\% accuracy); the weak point is reference-stratum reliability for evaluation type ($\kappa = 0.061$ on $n = 17$ pairwise rows). Excess homogenisation beyond vocabulary compression is significantly positive for models (\Cref{sec:cooccurrence}).

Second, LLMs narrow first-pass method menus more broadly. Model diversity contracts from an effective 1,232 entities in the paper-derived inventory to 59--96 in LLM suggestions (these effective numbers inherit extraction and normalisation uncertainty, \Cref{sec:annotation_validation,sec:normalisation_sensitivity}). At the corpus level, 78--90\% of paper-side entities receive no coverage from any model, though many exact-name misses are errors of granularity. Per-paper recall rises to 20--28\% at family level and 45--53\% at provider level (per-paper means on the $n = 915$ shared-paper subset, not directly comparable to the corpus-level exact-name statistics above). Inter-LLM rank correlations generally exceed LLM-to-paper correlations, so the compressions are shared rather than idiosyncratic, with the three models converging on similar subsets of the methodology space.

First-pass suggestions can shape the initial menu of options researchers consider before deeper literature review, so these patterns matter for downstream design. If multiple frontier models share similar provider-centred and popularity-weighted tendencies, over-reliance on them may narrow the subset of experimental designs that receive early consideration~\citep{bommasani2022picking,doshi2024generative, anderson2024homogenization,fortunato2018science, musslick2025automating}. These results do not show that LLMs ignore the question. Popularity baselines, BM25 retrieval calibration, equal-count comparisons, and paper-by-paper similarity tests all show that suggestions still vary with the input problem (\Cref{sec:entity}). LLMs respond to the task, but through a compressed and commercially concentrated vocabulary. This conclusion rests primarily on the strong provider dimension, and claims involving moderate or lower-tier dimensions should be weighted accordingly (\Cref{sec:annotation_validation}).

\paragraph{Limitations.} This study isolates the recommendation behaviour of LLMs under research-question-only prompting in one field and evaluates it against a paper-derived reference corpus rather than a normative optimum. It therefore identifies output-level redistribution of methodological attention, not the mechanism producing it, and it does not by itself establish generality beyond recent arXiv computer-science papers on LLM-related topics. Because research-question generation, paper-side entity extraction, and taxonomy classification all use GPT-5.1, reference-inventory--GPT-5.1 comparisons are not independent. Our main claims rest on patterns shared across all three models rather than on GPT-5.1-specific closeness, and a model-swap robustness check using Claude Opus~4.6 for both extraction and classification confirms the same provider-concentration pattern (\Cref{sec:annotation_validation}). Exact-name model misses are also partly inflated by temporal mismatch, because the paper corpus (June--December 2025) postdates the known training-data cutoffs of GPT-5.1 and Gemini~3~Pro. The suggestion prompt explicitly asks for ``one or more'' specific names and a ``structured and short'' pipeline (\Cref{sec:prompt_suggestion}), so part of the observed compression reflects this concise, single-shot elicitation rather than a property intrinsic to LLM-assisted methodology search more broadly. Adding richer context (abstracts, full papers, iterative follow-up, or retrieval augmentation) might reduce the observed compression, so the results bound a lower-context baseline rather than the full range of LLM-assisted research workflows~\citep{Baek2025ResearchAgent, li2024mlrcopilotautonomousmachinelearning, skarlinski2024language}.

Audit confidence varies across dimensions. Provider and openness are strong ($\kappa \geq 0.923$). Modality, evaluation type, and linguistic scope are moderate ($\kappa = 0.588$--$0.756$). Size is tentative (model-side LLM accuracy drops to 47.1\%), and all others are exploratory (\Cref{sec:annotation_validation}). These confidence tiers apply to all consensus-row diagnostics, which themselves are best-case checks covering 59--95\% of audited rows rather than corpus-level accuracy estimates. These constraints bound the interpretation but do not alter the central descriptive result, which rests on the strongly validated provider dimension.

The architecture schema itself is a source of overlap. It combines backbone types (Transformer, CNN, RNN/LSTM, GNN), functional roles (Generative), and training paradigms (Reinforcement Learning), and this mixing contributes to the label sensitivity of the architecture category and helps explain why results shift sharply under the web-search ablation. The schema notes already acknowledge leakage between task types and adjacent semantic categories. The resulting low reliability reflects a structural limitation of the taxonomy, which conflates distinct dimensions, rather than a pure annotation issue. Making the limitation explicit clarifies where and why the measurement becomes unreliable.

More broadly, LLM-based research support should be evaluated not only for relevance or correctness, but also for how it redistributes methodological attention across providers, families, and long-tail alternatives. The vocabulary compression and provider concentration documented here are output-level distributional regularities. Our study does not distinguish whether they arise from training-data exposure, benchmark popularity, or internal model preferences, so disentangling these mechanisms is an important direction for future work.

\paragraph{Future directions.} The most direct extension is to test whether richer suggestion-time context, iterative prompting, or retrieval reduces the convergence patterns we observe here. Replicating the analysis in other scientific domains would show whether the same distortions generalise beyond computer science, and longitudinal analyses could test whether they weaken or reinforce as training data changes. Causal studies of how LLM advice changes researchers' actual methodological choices would help clarify whether the descriptive concentration patterns documented here translate into measurable shifts in experimental practice.

\clearpage
\section*{Acknowledgements}
This research was supported by funding from the Flemish Government under the ``Onderzoeksprogramma Artifici\"{e}le Intelligentie (AI) Vlaanderen'' program.
Andres Algaba acknowledges support from the Francqui Foundation (Belgium) through a Francqui Start-Up Grant and a fellowship from the Research Foundation Flanders (FWO) under Grant No. 1286924N.
Vincent Ginis acknowledges support from Research Foundation Flanders (FWO) under Grant Nos. G032822N and G0K9322N.
The resources and services used in this work were provided by the VSC (Flemish Supercomputer Center), funded by Research Foundation Flanders (FWO) and the Flemish Government.

\section*{Data and code availability statement}
The code for the full pipeline, including data collection, arXiv license-metadata collection, paper-side entity extraction, LLM suggestion generation, entity normalisation, taxonomy classification, and figure generation, is available at \url{https://github.com/francescacarlon/Thinking-Like-a-Scientist}. The processed analysis outputs, including derived entity inventories, frequency counts, ratio tables, taxonomy classifications, co-occurrence matrices, and per-paper arXiv license metadata, are included in the repository. We do not redistribute arXiv PDFs, extracted full texts, substantial expressive excerpts from papers, or API batch/input/output files containing paper full text. Raw arXiv papers can be retrieved from arXiv using the collection script, subject to arXiv's API terms and the license associated with each paper. Local raw PDFs and extracted full-text files, and provider-side batch/input/output file objects under our control, were deleted after the structured annotations had been generated and verified.

\clearpage
\bibliographystyle{plainnat}
\bibliography{references}

@article{lu2024aiscientistfullyautomated,
  title={Towards end-to-end automation of AI research},
  author={Lu, Chris and Lu, Cong and Lange, Robert Tjarko and Yamada, Yutaro and Hu, Shengran and Foerster, Jakob and Ha, David and Clune, Jeff},
  journal={Nature},
  volume={651},
  number={8107},
  pages={914--919},
  year={2026},
  publisher={Nature Publishing Group UK London}
}

@article{li2024mlrcopilotautonomousmachinelearning,
  title={Mlr-copilot: Autonomous machine learning research based on large language models agents},
  author={Li, Ruochen and Patel, Teerth and Wang, Qingyun and Du, Xinya},
  journal={arXiv preprint arXiv:2408.14033},
  year={2024}
}

@article{liao2024llmsresearchtoolslarge,
  title={Llms as research tools: A large scale survey of researchers' usage and perceptions},
  author={Liao, Zhehui and Antoniak, Maria and Cheong, Inyoung and Cheng, Evie Yu-Yen and Lee, Ai-Heng and Lo, Kyle and Chang, Joseph Chee and Zhang, Amy X},
  journal={arXiv preprint arXiv:2411.05025},
  year={2024}
}

@InProceedings{10.1007/978-3-031-66329-1_42,
author="Jain, Rishab
and Jain, Aditya",
editor="Arai, Kohei",
title="Generative AI in Writing Research Papers: A New Type of Algorithmic Bias and Uncertainty in Scholarly Work",
booktitle="Intelligent Systems and Applications",
year="2024",
publisher="Springer Nature Switzerland",
address="Cham",
pages="656--669",
isbn="9783031663291",
  doi = {10.1007/978-3-031-66329-1_42},
  issn = {2367-3389},
  url = {https://doi.org/10.1007/978-3-031-66329-1_42},
}

@inproceedings{schmidgall-etal-2025-agent,
    title = "Agent Laboratory: Using {LLM} Agents as Research Assistants",
    author = "Schmidgall, Samuel  and
      Su, Yusheng  and
      Wang, Ze  and
      Sun, Ximeng  and
      Wu, Jialian  and
      Yu, Xiaodong  and
      Liu, Jiang  and
      Moor, Michael  and
      Liu, Zicheng  and
      Barsoum, Emad",
    editor = "Christodoulopoulos, Christos  and
      Chakraborty, Tanmoy  and
      Rose, Carolyn  and
      Peng, Violet",
    booktitle = "Findings of the Association for Computational Linguistics: EMNLP 2025",
    month = nov,
    year = "2025",
    address = "Suzhou, China",
    publisher = "Association for Computational Linguistics",
    url = "https://aclanthology.org/2025.findings-emnlp.320/",
    doi = "10.18653/v1/2025.findings-emnlp.320",
    pages = "5977--6043",
    ISBN = "979-8-89176-335-7"
}

@article{gridach2025agenticaiscientificdiscovery,
  title={Agentic ai for scientific discovery: A survey of progress, challenges, and future directions},
  author={Gridach, Mourad and Nanavati, Jay and Abidine, Khaldoun Zine El and Mendes, Lenon and Mack, Christina},
  journal={arXiv preprint arXiv:2503.08979},
  year={2025}
}

@inproceedings{si2025llmsgeneratenovelresearch,
  title={Can LLMs Generate Novel Research Ideas? A Large-Scale Human Study with 100+ NLP Researchers},
  author={Chenglei Si and Diyi Yang and Tatsunori Hashimoto},
  booktitle={The Thirteenth International Conference on Learning Representations},
  year={2025},
  url={https://proceedings.iclr.cc/paper_files/paper/2025/hash/ea94957d81b1c1caf87ef5319fa6b467-Abstract-Conference.html},
}

@inproceedings{li-etal-2025-chain-ideas,
    title = "Chain of Ideas: Revolutionizing Research Via Novel Idea Development with {LLM} Agents",
    author = "Li, Long  and
      Xu, Weiwen  and
      Guo, Jiayan  and
      Zhao, Ruochen  and
      Li, Xingxuan  and
      Yuan, Yuqian  and
      Zhang, Boqiang  and
      Jiang, Yuming  and
      Xin, Yifei  and
      Dang, Ronghao  and
      Rong, Yu  and
      Zhao, Deli  and
      Feng, Tian  and
      Bing, Lidong",
    editor = "Christodoulopoulos, Christos  and
      Chakraborty, Tanmoy  and
      Rose, Carolyn  and
      Peng, Violet",
    booktitle = "Findings of the Association for Computational Linguistics: EMNLP 2025",
    month = nov,
    year = "2025",
    address = "Suzhou, China",
    publisher = "Association for Computational Linguistics",
    url = "https://aclanthology.org/2025.findings-emnlp.477/",
    doi = "10.18653/v1/2025.findings-emnlp.477",
    pages = "8971--9004",
    ISBN = "979-8-89176-335-7"
}

@article{Lund_2023,
   title={ChatGPT and a new academic reality: Artificial intelligence-written research papers and the ethics of large language models in scholarly publishing},
   volume={74},
   ISSN={2330-1643},
   url={http://dx.doi.org/10.1002/asi.24750},
   DOI={10.1002/asi.24750},
   number={5},
   journal={Journal of the Association for Information Science and Technology},
   publisher={Wiley},
   author={Lund, Brady D. and Wang, Ting and Mannuru, Nishith Reddy and Nie, Bing and Shimray, Somipam and Wang, Ziang},
   year={2023},
   month=mar, pages={570--581} }

@article{villaescusanavarro2025denarioprojectdeepknowledge,
  title={The Denario project: Deep knowledge AI agents for scientific discovery},
  author={Villaescusa-Navarro, Francisco and Bolliet, Boris and Villanueva-Domingo, Pablo and Bayer, Adrian E and Acquah, Aidan and Amancharla, Chetana and Barzilay-Siegal, Almog and Bermejo, Pablo and Bilodeau, Camille and Ram{\'\i}rez, Pablo C{\'a}rdenas and others},
  journal={arXiv preprint arXiv:2510.26887},
  year={2025}
}

@article{algaba2025deeplargelanguagemodels,
  title={How deep do large language models internalize scientific literature and citation practices?},
  author={Algaba, Andres and Holst, Vincent and Tori, Floriano and Mobini, Melika and Verbeken, Brecht and Wenmackers, Sylvia and Ginis, Vincent},
  journal={arXiv preprint arXiv:2504.02767},
  year={2025}
}

@article{qi2023largelanguagemodelszero,
  title={Large language models are zero shot hypothesis proposers},
  author={Qi, Biqing and Zhang, Kaiyan and Li, Haoxiang and Tian, Kai and Zeng, Sihang and Chen, Zhang-Ren and Zhou, Bowen},
  journal={arXiv preprint arXiv:2311.05965},
  year={2023}
}

@inproceedings{Baek2025ResearchAgent,
  title={ResearchAgent: Iterative Research Idea Generation over Scientific Literature with Large Language Models},
  author={Jinheon Baek and Sujay Kumar Jauhar and Silviu Cucerzan and Sung Ju Hwang},
  booktitle={Proceedings of the 2025 Conference of the Nations of the Americas Chapter of the Association for Computational Linguistics: Human Language Technologies (Volume 1: Long Papers)},
  year={2025},
  url={https://doi.org/10.18653/v1/2025.naacl-long.342},
  doi = {10.18653/v1/2025.naacl-long.342},
  pages = {6709--6738},
  publisher = {Association for Computational Linguistics},
}

@inproceedings{wang-etal-2024-scimon,
    title = "{S}ci{MON}: Scientific Inspiration Machines Optimized for Novelty",
    author = "Wang, Qingyun  and
      Downey, Doug  and
      Ji, Heng  and
      Hope, Tom",
    editor = "Ku, Lun-Wei  and
      Martins, Andre  and
      Srikumar, Vivek",
    booktitle = "Proceedings of the 62nd Annual Meeting of the Association for Computational Linguistics (Volume 1: Long Papers)",
    month = aug,
    year = "2024",
    address = "Bangkok, Thailand",
    publisher = "Association for Computational Linguistics",
    url = "https://aclanthology.org/2024.acl-long.18/",
    doi = "10.18653/v1/2024.acl-long.18",
    pages = "279--299"
}

@article{mitchener2025kosmosaiscientistautonomous,
  title={Kosmos: An ai scientist for autonomous discovery},
  author={Mitchener, Ludovico and Yiu, Angela and Chang, Benjamin and Bourdenx, Mathieu and Nadolski, Tyler and Sulovari, Arvis and Landsness, Eric C and Barabasi, Daniel L and Narayanan, Siddharth and Evans, Nicky and others},
  journal={arXiv preprint arXiv:2511.02824},
  year={2025}
}

@article{liu2025visionautoresearchllm,
  title={A vision for auto research with llm agents},
  author={Liu, Chengwei and Wang, Chong and Cao, Jiayue and Ge, Jingquan and Wang, Kun and Zhang, Lyuye and Cheng, Ming-Ming and Zhao, Penghai and Li, Tianlin and Jia, Xiaojun and others},
  journal={arXiv preprint arXiv:2504.18765},
  year={2025}
}

@incollection{gandhi2025researchcodeagentllmmultiagentautomated,
  title={ResearchCodeAgent: An LLM Multi-Agent System for Automated Codification of Research Methodologies},
  author={Shubham Gandhi and Dhruv Shah and Manasi Patwardhan and Lovekesh Vig and Gautam Shroff},
  booktitle={AI for Research and Scalable, Efficient Systems},
  year={2025},
  pages={3--37},
  publisher={Springer Nature Singapore},
  doi={10.1007/978-981-96-8912-5_1},
  url={https://doi.org/10.1007/978-981-96-8912-5_1},
  issn={1865-0937},
  isbn={9789819689125},
}

@article{Elbadawi2024AIResearch,
  author  = {Elbadawi, Moe and Li, Hanxiang and Basit, Abdul W. and Gaisford, Simon},
  title   = {The role of artificial intelligence in generating original scientific research},
  journal = {International Journal of Pharmaceutics},
  year    = {2024},
  volume  = {652},
  pages   = {123741},
  month   = {March},
  doi     = {10.1016/j.ijpharm.2023.123741},
  pmid    = {38181989},
  note    = {Epub 2024 Jan 3}
}

@article{ioannidis2005most,
  title={Why most published research findings are false},
  author={Ioannidis, John PA},
  journal={PLoS medicine},
  volume={2},
  number={8},
  pages={e124},
  year={2005},
  publisher={Public Library of Science (PLoS)},
  doi = {10.1371/journal.pmed.0020124},
  issn = {1549-1676},
  month = {aug},
  url = {https://doi.org/10.1371/journal.pmed.0020124},
}

@article{open2015estimating,
  title={Estimating the reproducibility of psychological science},
  author={{Open Science Collaboration}},
  journal={Science},
  volume={349},
  number={6251},
  pages={aac4716},
  year={2015},
  publisher={American Association for the Advancement of Science (AAAS)},
  doi = {10.1126/science.aac4716},
  issn = {1095-9203},
  month = {aug},
  url = {https://doi.org/10.1126/science.aac4716},
}

@article{kirk2023understanding,
  title={Understanding the effects of {RLHF} on {LLM} generalisation and diversity},
  author={Kirk, Robert and Mediratta, Ishita and Nalmpantis, Christoforos and Luketina, Jelena and Hambro, Eric and Grefenstette, Edward and Raileanu, Roberta},
  journal={arXiv preprint arXiv:2310.06452},
  year={2023}
}

@article{luo2026quest,
  title={Inducing Sustained Creativity and Diversity in Large Language Models},
  author={Luo, Queenie and King, Gary and Puett, Michael and Smith, Michael D},
  journal={arXiv preprint arXiv:2603.19519},
  year={2026}
}

@article{nielsen2021global,
  title={Global citation inequality is on the rise},
  author={Nielsen, Mathias Wullum and Andersen, Jens Peter},
  journal={Proceedings of the National Academy of Sciences},
  volume={118},
  number={7},
  pages={e2012208118},
  year={2021},
  publisher={Proceedings of the National Academy of Sciences},
  doi = {10.1073/pnas.2012208118},
  issn = {1091-6490},
  month = {feb},
  url = {https://doi.org/10.1073/pnas.2012208118},
}

@article{liang2024can,
  title={Can large language models provide useful feedback on research papers? {A} large-scale empirical analysis},
  author={Liang, Weixin and Zhang, Yuhui and Cao, Hancheng and Wang, Binglu and Ding, Daisy Yi and Yang, Xinyu and Vodrahalli, Kailas and He, Siyu and Smith, Daniel Scott and Yin, Yian and McFarland, Daniel A. and Zou, James},
  journal={NEJM AI},
  volume={1},
  number={8},
  pages={AIoa2400196},
  year={2024},
  publisher={Massachusetts Medical Society},
  doi = {10.1056/AIoa2400196},
  issn = {2836-9386},
  month = {jul},
  url = {https://doi.org/10.1056/AIoa2400196},
}

@article{skarlinski2024language,
  title={Language agents achieve superhuman synthesis of scientific knowledge},
  author={Skarlinski, Michael D and Cox, Sam and Laurent, Jon M and Braza, James D and Hinks, Michaela and Hammerling, Michael J and Ponnapati, Manvitha and Rodriques, Samuel G and White, Andrew D},
  journal={arXiv preprint arXiv:2409.13740},
  year={2024}
}

@article{boiko2023autonomous,
  title={Autonomous chemical research with large language models},
  author={Boiko, Daniil A and MacKnight, Robert and Kline, Ben and Gomes, Gabe},
  journal={Nature},
  volume={624},
  number={7992},
  pages={570--578},
  year={2023},
  doi = {10.1038/s41586-023-06792-0},
  issn = {1476-4687},
  month = {dec},
  publisher = {Springer Science and Business Media LLC},
  url = {https://doi.org/10.1038/s41586-023-06792-0},
}

@article{ballon2026benchmarks,
  title={Benchmarks Saturate When The Model Gets Smarter Than The Judge},
  author={Ballon, Marthe and Algaba, Andres and Verbeken, Brecht and Ginis, Vincent},
  journal={arXiv preprint arXiv:2601.19532},
  year={2026}
}

@article{kusumegi2025scientific,
  title={Scientific production in the era of large language models},
  author={Kusumegi, Keigo and Yang, Xinyu and Ginsparg, Paul and de Vaan, Mathijs and Stuart, Toby and Yin, Yian},
  journal={Science},
  volume={390},
  number={6779},
  pages={1240--1243},
  year={2025},
  doi = {10.1126/science.adw3000},
  issn = {1095-9203},
  month = {dec},
  publisher = {American Association for the Advancement of Science (AAAS)},
  url = {https://doi.org/10.1126/science.adw3000},
}

@inproceedings{liu2025unequal,
  title={Unequal scientific recognition in the age of {LLMs}},
  author={Liu, Yixuan and Elekes, {\'A}bel and Lu, Jianglin and Dorantes-Gilardi, Rodrigo and Barab{\'a}si, Albert-L{\'a}szl{\'o}},
  booktitle={Findings of the Association for Computational Linguistics: EMNLP 2025},
  pages={23558--23568},
  year={2025},
  doi={10.18653/v1/2025.findings-emnlp.1279},
  publisher={Association for Computational Linguistics},
  url={https://doi.org/10.18653/v1/2025.findings-emnlp.1279},
}

@article{mobini2026structurally,
  title={Structurally Human, Semantically Biased: Detecting {LLM}-Generated References with Embeddings and {GNNs}},
  author={Mobini, Melika and Holst, Vincent and Tori, Floriano and Algaba, Andres and Ginis, Vincent},
  journal={arXiv preprint arXiv:2601.20704},
  year={2026}
}

@inproceedings{algaba2025large,
  title={Large language models reflect human citation patterns with a heightened citation bias},
  author={Algaba, Andres and Mazijn, Carmen and Holst, Vincent and Tori, Floriano and Wenmackers, Sylvia and Ginis, Vincent},
  booktitle={Findings of the Association for Computational Linguistics: NAACL 2025},
  year={2025},
  doi = {10.18653/v1/2025.findings-naacl.381},
  pages = {6844--6879},
  publisher = {Association for Computational Linguistics},
  url = {https://doi.org/10.18653/v1/2025.findings-naacl.381},
}

@article{delgado2025transforming,
  title={Transforming literature screening: The emerging role of large language models in systematic reviews},
  author={Delgado-Chaves, Fernando M and Jennings, Matthew J and Atalaia, Antonio and Wolff, Justus and Horvath, Rita and Mamdouh, Zeinab M and Baumbach, Jan and Baumbach, Linda},
  journal={Proceedings of the National Academy of Sciences},
  volume={122},
  number={2},
  pages={e2411962122},
  year={2025},
  publisher={Proceedings of the National Academy of Sciences},
  doi = {10.1073/pnas.2411962122},
  issn = {1091-6490},
  month = {jan},
  url = {https://doi.org/10.1073/pnas.2411962122},
}

@article{yamada2025ai,
  title={The {AI} Scientist-v2: Workshop-level automated scientific discovery via agentic tree search},
  author={Yamada, Yutaro and Lange, Robert Tjarko and Lu, Cong and Hu, Shengran and Lu, Chris and Foerster, Jakob and Clune, Jeff and Ha, David},
  journal={arXiv preprint arXiv:2504.08066},
  year={2025}
}

@article{novikov2025alphaevolve,
  title={{AlphaEvolve}: A coding agent for scientific and algorithmic discovery},
  author={Alexander Novikov and Ng{\^a}n V{\~u} and Marvin Eisenberger and Emilien Dupont and Po-Sen Huang and Adam Zsolt Wagner and Sergey Shirobokov and Borislav Kozlovskii and Francisco J. R. Ruiz and Abbas Mehrabian and M. Pawan Kumar and Abigail See and Swarat Chaudhuri and George Holland and Alex Davies and Sebastian Nowozin and Pushmeet Kohli and Matej Balog},
  journal={arXiv preprint arXiv:2506.13131},
  year={2025}
}

@article{fortunato2018science,
  title={Science of science},
  author={Fortunato, Santo and Bergstrom, Carl T. and B{\"o}rner, Katy and Evans, James A. and Helbing, Dirk and Milojevi{\'c}, Sta{\v{s}}a and Petersen, Alexander M. and Radicchi, Filippo and Sinatra, Roberta and Uzzi, Brian and Vespignani, Alessandro and Waltman, Ludo and Wang, Dashun and Barab{\'a}si, Albert-L{\'a}szl{\'o}},
  journal={Science},
  volume={359},
  number={6379},
  pages={eaao0185},
  year={2018},
  publisher={American Association for the Advancement of Science (AAAS)},
  doi={10.1126/science.aao0185},
  issn={1095-9203},
  month={mar},
  url={https://doi.org/10.1126/science.aao0185}
}

@article{musslick2025automating,
  title={Automating the practice of science: Opportunities, challenges, and implications},
  author={Musslick, Sebastian and Bartlett, Laura K. and Chandramouli, Suyog H. and Dubova, Marina and Gobet, Fernand and Griffiths, Thomas L. and Hullman, Jessica and King, Ross D. and Kutz, J. Nathan and Lucas, Christopher G. and Mahesh, Suhas and Pestilli, Franco and Sloman, Sabina J. and Holmes, William R.},
  journal={Proceedings of the National Academy of Sciences},
  volume={122},
  number={5},
  pages={e2401238121},
  year={2025},
  publisher={Proceedings of the National Academy of Sciences},
  doi = {10.1073/pnas.2401238121},
  issn = {1091-6490},
  month = {jan},
  url = {https://doi.org/10.1073/pnas.2401238121},
}

@article{jost2006entropy,
  title={Entropy and diversity},
  author={Jost, Lou},
  journal={Oikos},
  volume={113},
  number={2},
  pages={363--375},
  year={2006},
  publisher={Wiley},
  doi={10.1111/j.2006.0030-1299.14714.x},
  issn={1600-0706},
  month={may},
  url={https://doi.org/10.1111/j.2006.0030-1299.14714.x}
}

@inproceedings{bommasani2022picking,
  title={Picking on the same person: Does algorithmic monoculture lead to outcome homogenization?},
  author={Bommasani, Rishi and Creel, Kathleen A and Kumar, Ananya and Jurafsky, Dan and Liang, Percy},
  booktitle={Advances in Neural Information Processing Systems},
  volume={35},
  pages={3663--3678},
  year={2022}
}

@article{doshi2024generative,
  title={Generative AI enhances individual creativity but reduces the collective diversity of novel content},
  author={Doshi, Anil R and Hauser, Oliver P},
  journal={Science Advances},
  volume={10},
  number={28},
  pages={eadn5290},
  year={2024},
  doi={10.1126/sciadv.adn5290}
}

@inproceedings{anderson2024homogenization,
  title={Homogenization Effects of Large Language Models on Human Creative Ideation},
  author={Anderson, Barrett R. and Shah, Jash Hemant and Kreminski, Max},
  booktitle={Proceedings of the 16th Conference on Creativity and Cognition},
  pages={413--425},
  year={2024},
  organization={ACM},
  doi={10.1145/3635636.3656204}
}

@article{camerer2016evaluating,
  title={Evaluating replicability of laboratory experiments in economics},
  author={Camerer, Colin F. and Dreber, Anna and Forsell, Eskil and Ho, Teck-Hua and Huber, J{\"u}rgen and Johannesson, Magnus and Kirchler, Michael and Almenberg, Johan and Altmejd, Adam and Chan, Taizan and Heikensten, Emma and Holzmeister, Felix and Imai, Taisuke and Isaksson, Siri and Nave, Gideon and Pfeiffer, Thomas and Razen, Michael and Wu, Hang},
  journal={Science},
  volume={351},
  number={6280},
  pages={1433--1436},
  year={2016},
  publisher={American Association for the Advancement of Science (AAAS)},
  doi = {10.1126/science.aaf0918},
  issn = {1095-9203},
  month = {mar},
  url = {https://doi.org/10.1126/science.aaf0918},
}

@misc{hewitt2024predicting,
  title={Predicting Results of Social Science Experiments Using Large Language Models},
  author={Hewitt, Luke and Ashokkumar, Ashwini and Ghezae, Isaias and Willer, Robb},
  year={2024},
  note={Working paper},
  url={https://ai4pb.stanford.edu/projects/predicting-results-of-social-science-experiments-using-large-language-models},
}

@techreport{manning2024automated,
  title={Automated Social Science: Language Models as Scientist and Subjects},
  author={Manning, Benjamin S and Zhu, Kehang and Horton, John J},
  year={2024},
  institution={National Bureau of Economic Research},
  type={Working Paper},
  number={32381},
  doi={10.3386/w32381},
  url={https://www.nber.org/papers/w32381},
}

@article{gottweis2025towards,
  title={Towards an AI co-scientist},
  author={Juraj Gottweis and Wei-Hung Weng and Alexander Daryin and Tao Tu and Anil Palepu and Petar Sirkovic and Artiom Myaskovsky and Felix Weissenberger and Keran Rong and Ryutaro Tanno and Khaled Saab and Dan Popovici and Jacob Blum and Fan Zhang and Katherine Chou and Avinatan Hassidim and Burak Gokturk and Amin Vahdat and Pushmeet Kohli and Yossi Matias and Andrew Carroll and Kavita Kulkarni and Nenad Tomasev and Yuan Guan and Vikram Dhillon and Eeshit Dhaval Vaishnav and Byron Lee and Tiago R D Costa and Jos{\'e} R Penad{\'e}s and Gary Peltz and Yunhan Xu and Annalisa Pawlosky and Alan Karthikesalingam and Vivek Natarajan},
  journal={arXiv preprint arXiv:2502.18864},
  year={2025}
}

@article{singh2026openai,
  title={{OpenAI} {GPT}-5 System Card},
  author={Singh, Aaditya and Fry, Adam and Perelman, Adam and Tart, Adam and Ganesh, Adi and El-Kishky, Ahmed and McLaughlin, Aidan and Low, Aiden and Ostrow, AJ and Ananthram, Akhila and Nathan, Akshay and Luo, Alan and Helyar, Alec and Madry, Aleksander and Efremov, Aleksandr and others},
  journal={arXiv preprint arXiv:2601.03267},
  year={2025},
  doi={10.48550/arXiv.2601.03267},
  url={https://arxiv.org/abs/2601.03267},
}

@article{shannon1948mathematical,
  title={A mathematical theory of communication},
  author={Shannon, Claude Elwood},
  journal={The Bell system technical journal},
  volume={27},
  number={3},
  pages={379--423},
  year={1948},
  publisher={Nokia Bell Labs}
}

@article{lin1991divergence,
  title={Divergence measures based on the {Shannon} entropy},
  author={Lin, Jianhua},
  journal={IEEE Transactions on Information Theory},
  volume={37},
  number={1},
  pages={145--151},
  year={1991},
  doi={10.1109/18.61115}
}

@article{benjamini1995controlling,
  title={Controlling the false discovery rate: a practical and powerful approach to multiple testing},
  author={Benjamini, Yoav and Hochberg, Yosef},
  journal={Journal of the Royal Statistical Society: Series B (Methodological)},
  volume={57},
  number={1},
  pages={289--300},
  year={1995},
  doi={10.1111/j.2517-6161.1995.tb02031.x}
}

@book{cramer1946mathematical,
  title={Mathematical Methods of Statistics},
  author={Cram{\'e}r, Harald},
  year={1946},
  publisher={Princeton University Press},
  address={Princeton, NJ}
}

@book{cohen1988statistical,
  title={Statistical Power Analysis for the Behavioral Sciences},
  author={Cohen, Jacob},
  edition={2nd},
  year={1988},
  publisher={Lawrence Erlbaum Associates},
  address={Hillsdale, NJ}
}

@article{spearman1904proof,
  title={The proof and measurement of association between two things},
  author={Spearman, Charles},
  journal={American Journal of Psychology},
  volume={15},
  number={1},
  pages={72--101},
  year={1904},
  doi={10.2307/1412159}
}

@article{jaccard1912distribution,
  title={The distribution of the flora in the alpine zone},
  author={Jaccard, Paul},
  journal={New Phytologist},
  volume={11},
  number={2},
  pages={37--50},
  year={1912},
  doi={10.1111/j.1469-8137.1912.tb05611.x}
}

@book{gini1912variabilita,
  title={Variabilit{\`a} e mutabilit{\`a}: contributo allo studio delle distribuzioni e delle relazioni statistiche},
  author={Gini, Corrado},
  year={1912},
  publisher={Tipografia di Paolo Cuppini},
  address={Bologna},
  url={https://www.byterfly.eu/islandora/object/librib%3A680892},
}

@article{efron1993introduction,
  title={An introduction to the bootstrap},
  author={Tibshirani, Robert J and Efron, Bradley},
  journal={Monographs on statistics and applied probability},
  volume={57},
  number={1},
  pages={1--436},
  year={1993}
}

@article{pearson1900criterion,
  title={On the criterion that a given system of deviations from the probable in the case of a correlated system of variables is such that it can be reasonably supposed to have arisen from random sampling},
  author={Pearson, Karl},
  journal={Philosophical Magazine},
  volume={50},
  number={302},
  pages={157--175},
  year={1900},
  doi={10.1080/14786440009463897}
}

@article{levenshtein1966binary,
  title={Binary codes capable of correcting deletions, insertions and reversals},
  author={Levenshtein, Vladimir I.},
  journal={Soviet Physics Doklady},
  volume={10},
  number={8},
  pages={707--710},
  year={1966},
  note={English translation of the 1965 Russian original}
}

@article{deepseekai2025v32,
  title={{DeepSeek-V3.2}: Pushing the Frontier of Open Large Language Models},
  author={{DeepSeek-AI} and Aixin Liu and Aoxue Mei and Bangcai Lin and Bing Xue and Bingxuan Wang and Bingzheng Xu and Bochao Wu and Bowei Zhang and Chaofan Lin and Chen Dong and Chengda Lu and Chenggang Zhao and Chengqi Deng and Chenhao Xu and others},
  journal={arXiv preprint arXiv:2512.02556},
  year={2025},
  doi={10.48550/arXiv.2512.02556},
  url={https://arxiv.org/abs/2512.02556},
}

@misc{deepseek_v32_release,
  title        = {DeepSeek-V3.2 Release},
  author       = {{DeepSeek}},
  year         = {2025},
  howpublished = {\url{https://api-docs.deepseek.com/news/news251201}},
  note         = {Published 2025-12-01; accessed 2026-06-01}
}

@misc{thefuzz2023,
  title={thefuzz: Fuzzy String Matching in {Python}},
  author={{SeatGeek}},
  year={2023},
  howpublished={\url{https://github.com/seatgeek/thefuzz}},
  note={Software library}
}

@article{cohen1960coefficient,
  title={A coefficient of agreement for nominal scales},
  author={Cohen, Jacob},
  journal={Educational and Psychological Measurement},
  volume={20},
  number={1},
  pages={37--46},
  year={1960},
  publisher={Sage Publications}
}

@article{shrout1979intraclass,
  title={Intraclass correlations: Uses in assessing rater reliability},
  author={Shrout, Patrick E and Fleiss, Joseph L},
  journal={Psychological Bulletin},
  volume={86},
  number={2},
  pages={420--428},
  year={1979},
  publisher={American Psychological Association}
}

@techreport{gemini3pro_modelcard,
  title        = {Gemini 3 Pro Model Card},
  author       = {{Google DeepMind}},
  year         = {2025},
  institution  = {Google DeepMind},
  url          = {https://storage.googleapis.com/deepmind-media/Model-Cards/Gemini-3-Pro-Model-Card.pdf}
}

@misc{arxiv_license_info,
  title        = {arXiv License Information},
  author       = {{arXiv}},
  year         = {2026},
  howpublished = {\url{https://info.arxiv.org/help/license/index.html}},
  note         = {Accessed 2026-06-01}
}

@misc{arxiv_api_terms,
  title        = {Terms of Use for arXiv APIs},
  author       = {{arXiv}},
  year         = {2026},
  howpublished = {\url{https://info.arxiv.org/help/api/tou.html}},
  note         = {Accessed 2026-06-01}
}

@misc{eu_dsm_directive,
  title        = {Directive (EU) 2019/790 of the European Parliament and of the Council on Copyright and Related Rights in the Digital Single Market},
  author       = {{European Parliament and Council of the European Union}},
  year         = {2019},
  howpublished = {\url{https://eur-lex.europa.eu/eli/dir/2019/790/oj/eng}}
}

@misc{belgian_dsm_transposition,
  title        = {European Directive on Copyright and Related Rights in the Digital Single Market -- Transposition in Belgian Law},
  author       = {{Belgian Federal Public Service Economy}},
  year         = {2022},
  howpublished = {\url{https://economie.fgov.be/en/themes/intellectual-property/intellectual-property-rights/copyright-and-related-rights/copyright/european-directive-copyright}},
  note         = {Accessed 2026-06-01}
}

@misc{openai_data_controls,
  title        = {Data Controls in the OpenAI Platform},
  author       = {{OpenAI}},
  year         = {2026},
  howpublished = {\url{https://platform.openai.com/docs/guides/your-data}},
  note         = {Accessed 2026-06-01}
}

@misc{openai_services_agreement,
  title        = {OpenAI Services Agreement},
  author       = {{OpenAI}},
  year         = {2026},
  howpublished = {\url{https://openai.com/policies/services-agreement/}},
  note         = {Effective 2026-01-01; accessed 2026-06-01}
}

@misc{google_gemini_api_terms,
  title        = {Gemini API Additional Terms of Service},
  author       = {{Google}},
  year         = {2026},
  howpublished = {\url{https://ai.google.dev/gemini-api/terms}},
  note         = {Effective 2026-03-23; accessed 2026-06-01}
}

@misc{deepseek_terms,
  title        = {DeepSeek Terms of Use},
  author       = {{DeepSeek}},
  year         = {2026},
  howpublished = {\url{https://cdn.deepseek.com/policies/en-US/deepseek-terms-of-use.html}},
  note         = {Last updated 2026-03-27; accessed 2026-06-01}
}

@misc{anthropic_commercial_terms,
  title        = {Anthropic Commercial Terms of Service},
  author       = {{Anthropic}},
  year         = {2026},
  howpublished = {\url{https://www.anthropic.com/legal/commercial-terms}},
  note         = {Effective 2025-06-17; accessed 2026-06-01}
}

@misc{anthropic_retention,
  title        = {How Long Do You Store My Data?},
  author       = {{Anthropic}},
  year         = {2026},
  howpublished = {\url{https://privacy.claude.com/en/articles/10023548-how-long-do-you-store-my-data}},
  note         = {Article dated 2026-03-16; accessed 2026-06-01}
}

\clearpage
\appendix

\section{Methods}
\label{sec:Methods}

\subsection{Data collection}
\label{sec:datacollection}

We retrieved 1,000 papers from arXiv by running an automated collection script on December 4, 2025. The script queried arXiv for Computer Science (CS) papers with ``LLM'' or ``Large Language Model'' in the title, appended a runtime \texttt{submittedDate} filter covering the preceding 182 days (approximately June--December 2025), sorted results by submission date, filtered the returned records to papers published in 2025, and then randomly sampled 1,000 papers from that pool. We chose this time frame so that the papers would post-date the known training-data cutoffs of GPT-5.1 (September~30, 2024) and Gemini~3~Pro (January 2025), yielding suggestions that are unlikely to rely on knowledge of the papers themselves. No official cutoff has been published for DeepSeek-V3.2, so prior exposure cannot be ruled out for that model. We filtered for papers in the Computer Science (CS) category with ``LLM'' or ``Large Language Model'' in the title, using the core query:
\begin{verbatim}
(ti:"LLM*" OR ti:"Large Language Model*") AND cat:cs.*
\end{verbatim}
with the date-window constraint added programmatically at runtime. The subcategories (e.g., Computation and Language, Artificial Intelligence, Computer Vision) were left unrestricted within the CS main category. For each paper, we downloaded the arXiv-hosted PDF and extracted full text using PyMuPDF. We also recorded the license URI declared in the arXiv metadata for each paper. The per-paper license metadata are included in the released derived-data table, and the aggregate license distribution is reported in \Cref{tab:arxiv_license_distribution}.

\subsection{Copyright, licensing, and text-and-data mining}
\label{sec:copyright_tdm}
We did not treat availability on arXiv, nor the arXiv default perpetual non-exclusive distribution license, as a general public license to redistribute or republish full text. arXiv states that e-prints remain subject to copyright protection, that redistribution requires permission unless a permissive license applies, and that its default non-exclusive license gives arXiv limited distribution rights while limiting reuse by other entities or individuals~\citep{arxiv_license_info,arxiv_api_terms}. For this reason, we restricted our handling of the PDFs to non-public computational processing. This distinction is also consistent with arXiv's API terms, which permit users to retrieve, store, and use arXiv e-print content for research purposes, while prohibiting users from storing and serving PDFs, source files, or other e-print content from their own servers unless authorised by the copyright holder or permitted by the e-print's license~\citep{arxiv_api_terms}.

The full-text collection and processing were conducted at Vrije Universiteit Brussel, a Belgian research organisation, for scientific research purposes, using works lawfully accessible through arXiv. We relied on the scientific text-and-data-mining exception in Article~3 of Directive~(EU)~2019/790, as transposed into Belgian law, specifically Article XI.191/1, \S~1, 7$^\circ$ of the Belgian Code of Economic Law, which permits reproductions and extractions by research organisations for scientific TDM over works to which they have lawful access and requires retained copies to be stored with an appropriate level of security~\citep{eu_dsm_directive,belgian_dsm_transposition}. The Directive defines research organisations to include universities, defines TDM as automated analysis of digital text/data to generate information, treats freely available online content as lawful access, and recognises that research organisations may rely on private partners or their technological tools to carry out TDM~\citep{eu_dsm_directive}. Article~4's reservation mechanism for general TDM does not affect Article~3 scientific TDM, and contractual provisions contrary to the Article~3 exception are unenforceable under Article~7~\citep{eu_dsm_directive}. Third-party API processing followed the controls summarised in \Cref{sec:api_data_handling}: OpenAI API submissions were not used for model training or improvement absent opt-in; Anthropic API/commercial use was not used for model training absent opt-in, and model-improvement was disabled where applicable for Claude Code; and Gemini calls were made from VUB/Belgium/EEA, where Google's Gemini API terms apply paid-services data-use protections to all Gemini API/AI Studio services, including unpaid quota~\citep{openai_data_controls,openai_services_agreement,google_gemini_api_terms,anthropic_commercial_terms,anthropic_retention}. The TDM outputs analysed and released in this study are derived annotations and aggregate statistics. We do not redistribute arXiv PDFs, extracted full texts, substantial expressive excerpts from the papers, or API batch/input/output files containing paper full text.

{\color{red}
\begin{table}[H]
\centering
\caption{\textbf{arXiv license distribution in the 1,000-paper corpus.} License metadata were retrieved from the arXiv metadata record for each paper. The table is reported for transparency and to delimit downstream reuse. Inclusion in the computational analysis did not rely on permissive Creative Commons licensing, but on the scientific TDM basis described in \Cref{sec:copyright_tdm}.}
\label{tab:arxiv_license_distribution}
\begin{tabular}{lrr}
\toprule
\textbf{arXiv-declared license} & \textbf{Number of papers} & \textbf{Percentage} \\
\midrule
arXiv.org perpetual, non-exclusive license & 385 & 38.5\% \\
CC BY 4.0 & 451 & 45.1\% \\
CC BY-SA 4.0 & 24 & 2.4\% \\
CC BY-NC-SA 4.0 & 58 & 5.8\% \\
CC BY-NC-ND 4.0 & 51 & 5.1\% \\
CC0 1.0 & 8 & 0.8\% \\
Other / missing / custom notice & 23 & 2.3\% \\
\bottomrule
\end{tabular}
\end{table}
}

\subsection{Paper-side entity extraction and research question generation}
\label{sec:gtextraction}

For each of these 1,000 papers, we prompted GPT-5.1 with the title, abstract, and full text to generate a main research question (RQ) and to extract the paper-side reference entities, under the copyright/TDM and API data-handling procedures described in \Cref{sec:copyright_tdm,sec:api_data_handling}. This was an inference-only use of the OpenAI Batch API. The submitted texts were not used for model training or fine-tuning, and the OpenAI batch input/output file objects under our control were deleted after processing. The resulting RQ was then used as the input prompt for the suggestion step described below.

\paragraph{Paper-side entity extraction.} We prompted GPT-5.1 via the OpenAI Batch API, providing it with the title, abstract, and full text, to extract:
\begin{enumerate}
    \item \textbf{Datasets} used in the paper's experiments (excluding those only mentioned in related work).
    \item \textbf{Models} and architectures used experimentally.
    \item \textbf{Evaluation metrics} reported in experiments.
\end{enumerate}
We requested JSON with a \texttt{research\_question} field and \texttt{GroundTruth...} entity lists, with each entity name limited to 1--3 words for consistency. In the saved raw batch outputs, some responses instead used generic \texttt{datasets/models/metrics} keys; downstream parsing harmonized both key variants into the standardized paper-side schema used for analysis. The full extraction prompt is provided in \Cref{sec:prompt_extraction}.

\paragraph{LLM suggestion generation.} Using the generated RQ, we prompted three LLMs to suggest methodology components given only the research question. Gemini~3~Pro and DeepSeek-V3.2 did not receive the arXiv PDFs, abstracts, or extracted full texts:
\begin{itemize}
    \item \textbf{GPT-5.1} (OpenAI), exact experimental model ID \texttt{gpt-5.1-2025-11-13}, documented in the GPT-5 system card~\citep{singh2026openai}, knowledge cutoff September~30, 2024, accessed via the OpenAI Batch API (\texttt{/v1/responses} endpoint, reasoning effort set to \texttt{medium}).
    \item \textbf{Gemini~3~Pro} (Google), model ID \texttt{gemini-3-pro-preview}, knowledge cutoff January 2025, accessed via the Google GenAI Batch API.
    \item \textbf{DeepSeek-V3.2} (DeepSeek)~\citep{deepseekai2025v32,deepseek_v32_release}, accessed via the DeepSeek API alias \texttt{model="deepseek-reasoner"} on the OpenAI-compatible \texttt{/v1/chat/completions} endpoint. DeepSeek announced DeepSeek-V3.2 as live on its App, Web, and API on December~1, 2025, and documented V3.2 support for thinking/tool-use mode. Our API calls ran between December 2025 and January 2026. Because DeepSeek compatibility aliases are time-dependent and may later be remapped on the live Models \& Pricing page, we report the version used at execution time according to our run logs and API configuration.
\end{itemize}
Each model was asked to suggest suitable datasets, models, evaluation metrics, and a structured experimental pipeline describing how these components would be used together to address the RQ. Responses were collected in JSON format: GPT-5.1 via the OpenAI Batch API, Gemini~3~Pro via the Google GenAI Batch API, and DeepSeek-V3.2 via asynchronous calls to the standard DeepSeek \texttt{/v1/chat/completions} endpoint. All API calls were executed between December 2025 and January 2026. The suggestion prompt, identical across all three models except for LLM-specific JSON key prefixes, is provided in \Cref{sec:prompt_suggestion}.

\subsection{\texorpdfstring{Comparison scope and analysis denominators}{Comparison scope and analysis denominators}}
\label{sec:comparison_scope}

The comparison is intentionally asymmetric: the paper-derived reference inventory is an automated extraction of the datasets, models, and metrics used experimentally in each paper (with extraction reliability quantified by ICC$(2,1) = 0.469$; \Cref{sec:annotation_validation}), whereas the suggestion prompt asks for a short set of suitable first-pass choices and a concise ``straightforward'' pipeline from the research question alone. The analysis therefore targets compression and reweighting of first-pass methodological menus rather than exact reconstruction of full paper-level experimental inventories. Throughout the narrative and in visible table labels we use ``paper-derived reference inventory'' (or ``reference inventory'', ``Reference''); the tag ``GT'' or ``Ground Truth'' persists only in pipeline artefacts (JSON schema keys such as \texttt{GroundTruthDatasets}, CSV column headers, filenames, and a small number of pre-generated figure panels) that would be fragile to rename. This reference corpus is itself an automated extraction with only moderate inter-model agreement on counts (ICC$(2,1) = 0.469$; \Cref{sec:annotation_validation}). The measured object is therefore the compound pipeline \textit{paper} $\rightarrow$ \textit{GPT-5.1-generated research question} $\rightarrow$ \textit{model suggestion} $\rightarrow$ \textit{GPT-5.1-derived ontology}, not an independent model-only benchmark. Our prompt explicitly asks for ``one or more'' specific names and a ``structured and short'' straightforward pipeline (\Cref{sec:prompt_suggestion}), so the observed compression likely reflects both model-side methodological priors and a concise-assistant elicitation frame. Unless otherwise noted, normalised entity-name analyses use the full 1,000-paper outputs for each source. Taxonomy-based analyses use the classified files available at analysis time (n$=1{,}000$ reference, 1{,}000 GPT-5.1, 981 Gemini~3~Pro, 998 DeepSeek-V3.2), and analyses requiring non-empty entities in both sources use smaller intersections reported in the corresponding captions and tables.

\subsection{Third-party API data handling}
\label{sec:api_data_handling}
All third-party model calls were inference-only calls. We did not fine-tune, train, or ask any provider to train models on the arXiv corpus. Provider-side file objects under our control were deleted after processing, while provider security, abuse-monitoring, or safety logs may be retained according to the applicable provider terms. Table~\ref{tab:api_data_handling} summarises the data submitted to each provider and the relevant data-handling controls. No provider was granted any independent right to redistribute the corpus, to expose full texts to users, or to use the paper full texts for model training.

\begin{table}[H]
\centering
\caption{\textbf{Third-party API/model data handling.} The table distinguishes full-text processing from downstream suggestion and validation calls.}
\label{tab:api_data_handling}
\small
\begin{adjustbox}{width=\textwidth}
\begin{tabular}{p{0.19\textwidth} p{0.28\textwidth} p{0.35\textwidth} p{0.18\textwidth}}
\toprule
\textbf{Provider/model} & \textbf{Data submitted} & \textbf{Training/model-improvement control} & \textbf{Post-processing action} \\
\midrule
OpenAI GPT-5.1 via Batch API; OpenAI GPT-5.4 via Codex/OpenAI tooling & GPT-5.1 extraction/RQ generation received title, abstract, and extracted full text for the 1,000 papers. OpenAI models also received extracted entity names, generated pipeline text, and sampled audit/classification inputs where applicable. & OpenAI states that API data are not used to train or improve OpenAI models unless the customer explicitly opts in; its endpoint table lists \texttt{/v1/responses}, \texttt{/v1/files}, and \texttt{/v1/batches} as ``No'' for data used for training, and OpenAI's service terms state that customer content is not used to develop or improve services unless the customer explicitly agrees~\citep{openai_data_controls,openai_services_agreement}. & No model-improvement opt-in; no fine-tuning; batch/input/output file objects under our control deleted after processing. \\
\addlinespace
Google Gemini~3~Pro via GenAI Batch API & Generated research question only; no arXiv PDF, abstract, or extracted full text. & Because our Gemini calls were made from VUB/Belgium/EEA, Google states that the paid-services data-use terms apply to all Gemini API/AI Studio services, including unpaid quota; under those paid-services terms, prompts, files, and responses are not used to improve Google products~\citep{google_gemini_api_terms}. & No paper full text submitted; outputs retained only as derived model suggestions. \\
\addlinespace
DeepSeek-V3.2 via DeepSeek API & Generated research question only; no arXiv PDF, abstract, or extracted full text. & DeepSeek's terms state that Inputs and Outputs may be used to improve its services unless users turn off ``Improve the model for everyone''~\citep{deepseek_terms}. DeepSeek was therefore not used for full-text TDM processing. & No paper full text submitted; DeepSeek was used only for downstream suggestion generation from generated research questions. \\
\addlinespace
Anthropic Claude Opus~4.6 via Claude Code and the Anthropic API & Reconstructed full text for the 94-paper robustness re-extraction; sampled audit/classification inputs for validation. & Anthropic's commercial terms state that customers retain inputs, own outputs, and that Anthropic may not train models on Customer Content. For Claude Code used under consumer Claude plans, Anthropic states that turning off model improvement prevents previous and new chats or coding sessions from being used for future model training, and that deleted conversations are removed from backend storage within 30 days~\citep{anthropic_commercial_terms,anthropic_retention}. & Model-improvement/training disabled before use where applicable; sessions/files under our control deleted after processing. \\
\bottomrule
\end{tabular}
\end{adjustbox}
\end{table}

\subsection{Entity normalisation}
\label{sec:normalisation}

Raw entity names from different sources exhibit surface-level variation (e.g., ``GPT-4o'' vs.\ ``GPT4o'', ``F1-score'' vs.\ ``F1''). We applied a two-stage normalisation:

\paragraph{Stage 1: Deterministic normalisation.} We applied a sequence of regex-based transformations: removing bracketed content, replacing separators (\texttt{/}, \texttt{;}, \texttt{|}) with commas, converting underscores and hyphens to spaces, stripping remaining punctuation, collapsing whitespace, and lowercasing. For evaluation metrics, manual inspection of the raw outputs revealed systematic sub-variant proliferation, so we applied additional hardcoded aggregation rules: all ROUGE variants (rouge-1, rouge-2, rouge-L) map to ``rouge'', all BLEU variants to ``bleu'', Pearson and Spearman correlation variants (including common misspellings) to their canonical forms, and all F1 score variants to ``f1''. Leading numeric prefixes (e.g., ``1accuracy'') were stripped.

\paragraph{Stage 2: Fuzzy clustering.} We clustered the deterministically normalised names using a greedy representative-matching procedure. Names are sorted by length (longest first) so that more specific names tend to become early cluster representatives. For each name, we compute the token-sort ratio against all existing cluster representatives. The token-sort ratio tokenises both strings, sorts the tokens alphabetically, and computes the normalised Levenshtein edit distance~\citep{levenshtein1966binary} on a $[0, 100]$ scale (implemented in \texttt{thefuzz}~\citep{thefuzz2023}). If any representative scores $\geq 90$, the name joins the best-scoring cluster; otherwise it seeds a new cluster. A prefix-safety check zeroes the score whenever one string is a strict prefix of the other (e.g., ``gpt4'' $\neq$ ``gpt4o''), preventing over-merging of distinct model variants. Manual inspection identified one additional hardcoded correction: the concatenated string ``gpt4o3mini'' was split into its constituent models. A sensitivity analysis over thresholds $T \in \{80, 85, 90, 95, 100\}$ confirms that the headline vocabulary-compression and coverage metrics are stable across this range (\Cref{sec:normalisation_sensitivity}); unless otherwise noted, main-text entity-name analyses use the default $T = 90$ clustering.

\subsection{Frequency analysis and comparison}
\label{sec:frequency}

For each entity type and each source (reference inventory, GPT-5.1, Gemini~3~Pro, DeepSeek-V3.2), we counted how often each entity appeared. The reference inventory contains significantly more unique models and evaluation metrics than the LLMs suggest, although the number of unique datasets is more comparable (\Cref{tab:vocab}). To obtain a fair comparison, we normalised the LLM suggestion counts into relative percentages against the reference totals. We report within-source percentages (the entity's count divided by the total count for that source) and normalised-to-reference percentages (the entity's count divided by the total reference count).

Vocabulary compression is quantified using two complementary measures. The Shannon entropy~\citep{shannon1948mathematical} of the frequency distribution is $H = -\sum_{i=1}^{S} p_i \ln p_i$ where $p_i$ is the relative frequency of entity $i$ among $S$ types (computed on mention counts, natural logarithm). The effective number of entities $\exp(H)$~\citep{jost2006entropy} represents the number of equally frequent entities producing the same entropy. The Gini coefficient~\citep{gini1912variabilita}
\begin{equation}
  G = \frac{2\sum_{i=1}^{n} i\, x_{(i)}}{n \sum_{i=1}^{n} x_{(i)}} - \frac{n+1}{n}
  \label{eq:gini}
\end{equation}
for frequency counts sorted in ascending order $x_{(1)} \le \cdots \le x_{(n)}$, measures concentration ($G = 0$ is perfect equality, $G = 1$ is maximal concentration).

To quantify rank-dependent amplification among shared entities, we fit $\log_{10}(r_i^{\mathrm{LLM}} / r_i^{\mathrm{ref}}) = \alpha + \beta \, \log_{10}(\mathrm{rank}_i^{\mathrm{ref}})$ by ordinary least squares, where $r_i$ denotes the relative frequency of entity $i$ and the superscript ``ref'' refers to the paper-derived reference inventory. A positive slope $\hat\beta$ indicates that LLMs amplify rarer shared entities proportionally more than frequent ones.

Inter-source agreement on top-ranked entities is measured by Jaccard@$K$~\citep{jaccard1912distribution}: $J@K = |T_K^{(a)} \cap T_K^{(b)}| \,/\, |T_K^{(a)} \cup T_K^{(b)}|$, where $T_K^{(s)}$ is the set of $K$ most frequent entities in source $s$ (ties broken alphabetically). We report $K = 20$ throughout.

\subsection{Taxonomy classification}
\label{sec:taxonomy_methods}

We prompted GPT-5.1 via the Batch API to classify all reference-inventory and LLM-suggested entities according to structured category dimensions:

\textbf{Datasets} were classified along the following dimensions: modality (text, image, audio, video, multimodal, \ldots), task type (classification, QA, generation, reasoning, \ldots), domain (general, scientific, healthcare, legal, \ldots), annotation type (supervised, semi-supervised, crowdsourced, \ldots), size (small: $<$10K samples, medium: 10K--100K, large: $>$100K), granularity (document, sentence, token, \ldots), linguistic scope (monolingual, multilingual, cross-lingual), cognitive/affective properties (reasoning, emotion, decision making, \ldots), and data quality (noisy, curated).

\textbf{Models} were classified by architecture (Transformer, CNN, RNN/LSTM, GNN, \ldots), training paradigm (supervised, self-supervised, few-shot, fine-tuning, RAG, \ldots), provider (OpenAI, Meta~AI, Google DeepMind, Anthropic, \ldots), openness (open or closed), and size (small: $<$1B, medium: 1--10B, large: 10--100B, extra-large: $>$100B parameters). The classifier itself uses this four-way model-size schema. In the downstream comparison helpers used for the summary size analyses, however, model size is normalised to three analysis bins by folding Extra-large into Large; the size percentages reported in \Cref{sec:taxonomy} should therefore be read as a combined top-end bucket rather than as the 10--100B band alone.

\textbf{Metrics} were classified by evaluation type (accuracy, ranking, regression, fairness, safety, efficiency, robustness, \ldots). The full classification prompt with all allowed values for each dimension is provided in \Cref{sec:prompt_taxonomy}. All classifier settings use this same schema. Unless otherwise noted, the main category results in \Cref{sec:taxonomy} use the entity-list-only setting, in which the classifier sees only entity lists. \Cref{sec:Ablations} quantifies how labels change when the generated experimental pipeline or web search are added, whereas \Cref{sec:StatisticalRobustness} recomputes the summary robustness tables when the classifier also sees pipeline context. The co-occurrence analyses use exports derived from that same richer setting.

\subsection{Co-occurrence analysis}
\label{sec:cooccurrence_methods}

We computed pairwise co-occurrence frequencies between taxonomy dimensions to assess whether LLMs reproduce the combinatorial structure of real research methodology. These co-occurrence tables come from the setting where the classifier also sees the generated pipeline, so they are not a direct reuse of the entity-list-only category distributions in \Cref{sec:taxonomy}. We analyse all 12 cross-entity co-occurrence pairs: 9 dataset--model pairs (task type, size, and cognitive/affective $\times$ architecture, provider, and openness) and 3 metric--model pairs (evaluation type $\times$ architecture, provider, and openness). For each pair of categories, we computed the percentage of papers (within each source) in which both categories co-occur. In ${\sim}$0.04\% of classifier outputs, a comma-delimited multi-label string (e.g., ``Retrieval, Reasoning'') was returned as a single field value rather than as separate items; these conjunction labels are carried through as atomic categories in the co-occurrence matrices and residual heatmaps, but their low frequency does not materially affect the results. The main text focuses on the two pairs with the clearest interpretive value, but the evaluation type $\times$ provider pair is the more strongly validated of the two because both axes are comparatively reliable in the blinded audit (though evaluation type's moderate $\kappa$ rests on the LLM stratum; the reference-inventory stratum has $\kappa = 0.061$ on $n = 17$ rows); task type $\times$ provider remains useful descriptive context. The displayed heatmaps are restricted to the top 7 row categories and top 7 provider columns selected by the largest row-wise and column-wise cell maxima after summing the aligned source matrices for each focal pair, whereas \Cref{fig:cooccurrence_jsd_appendix} and the residual analyses use the full aligned matrices subject to their stated support filters. For each co-occurrence pair, the row-wise JSD is computed as follows: for each row category $r$ shared between both the reference-inventory and LLM matrices, column counts are normalised to probability vectors and the JSD (\Cref{eq:jsd}, base-2) is computed; the reported mean is the average over all shared row categories. Residual analysis (\Cref{sec:residual_analysis}) complements this by separating structural co-occurrence shifts from simple overall frequency changes.

\clearpage
\section{Ablation studies}
\label{sec:Ablations}

This appendix is organized to separate checks about the labeling step from checks about whether the main conclusions survive alternative assumptions. \Cref{sec:pipeline_ablation} tests whether giving the taxonomy classifier the suggested pipeline changes the resulting overall category distributions, whereas \Cref{sec:websearch_ablation} tests whether web search mainly affects how GPT-5.1's suggestions are labeled. These analyses therefore tell us how sensitive the labeling step is; they do not mean the LLMs changed which entities they suggested.

\subsection{Pipeline context ablation}
\label{sec:pipeline_ablation}

In the suggestion step, each LLM produces not only entity lists but also a free-form experimental pipeline. We tested whether giving that extra context to the taxonomy classifier changes the resulting distributions by comparing two conditions: (1)~classifier sees the suggested entities and the experimental pipeline; and (2)~classifier sees only the entity lists. The entity-list-only condition is the baseline used for the main category figures in \Cref{sec:taxonomy}; this appendix quantifies how far the distributions move under the richer labeling setting.

\Cref{fig:ablation} shows the percentage-point change across dataset category dimensions for all three LLMs. The most consistent pattern across models is an increase in ``generation'' task-type classification when pipeline context is provided: $+3.6$ percentage points for GPT-5.1, $+3.1$ for Gemini~3~Pro, and $+3.5$ for DeepSeek-V3.2. Document-level granularity also increases consistently ($+3.2$ for GPT-5.1, $+2.6$ for Gemini~3~Pro, $+2.6$ for DeepSeek-V3.2; not shown in \Cref{fig:ablation}).

For model taxonomy, the largest shift is in architecture classification: with pipeline context, GPT-5.1's ``Generative'' category increases from 18.4\% to 23.8\% ($+5.3$ percentage points) while ``Transformer (all)'' decreases from 74.6\% to 69.6\% ($-5.0$ pp). This suggests that for GPT-5.1, pipeline context helps the classifier disambiguate between the broader Transformer family and more specific generative model types. Gemini~3~Pro shows smaller but directionally consistent shifts ($+$1.8~pp Generative, $-$1.7~pp Transformer), whereas DeepSeek-V3.2 shows small shifts in the opposite direction ($-$1.4~pp Generative, $+$1.6~pp Transformer), indicating that pipeline context does not uniformly disambiguate architecture labels across models. \Cref{tab:pipeline_ablation_model_metric} reports all model and metric subcategories where at least one LLM exceeds $|\Delta\text{pp}| \geq 2.0$; only the Generative/Transformer pair crosses this threshold, confirming that pipeline context has minimal effect on model and metric taxonomy distributions. For provider specifically, pipeline context reclassifies a small number of models from Other/Academic to specific commercial providers (GPT-5.1: 19/350, Gemini~3~Pro: 17/414, DeepSeek-V3.2: 8/268 models reclassified), but the overall impact on provider distributions is below 0.5~pp per category.

\Cref{tab:ablation_conditions} summarises the ablation conditions across models.

\subsection{Web search ablation}
\label{sec:websearch_ablation}

For GPT-5.1 only, we compared taxonomy classification with and without web search enabled. The most substantial effect is on model architecture classification: the ``Generative'' category increases from 18.4\% to 35.4\% with web search ($+17.0$ pp), while ``Transformer (all)'' drops from 74.6\% to 58.4\% ($-16.2$ pp). This shift indicates that architecture labels are sensitive to label operationalisation and retrieval context, in the sense that the classifier's access to web information changes how it categorises the same set of suggested models, rather than reflecting a change in the LLMs' underlying suggestion behaviour. Architecture-based findings throughout the paper should therefore be interpreted cautiously, as partly reflecting classifier sensitivity rather than stable properties of the suggestions themselves.

Training paradigm distributions also shift: few-shot and zero-shot learning both decrease (from ${\sim}14.5\%$ each to ${\sim}7.9\%$), while multi-task learning increases from 7.4\% to 11.8\%. Provider distributions remain largely stable ($<$2 percentage point changes), and model size distributions are minimally affected. These results suggest that web search primarily helps the classifier make finer-grained architecture distinctions rather than changing the fundamental distribution of suggested entities. \Cref{fig:websearch_ablation} visualises the effect across dataset, model, and metric dimensions.

\clearpage
\section{Statistical robustness}
\label{sec:StatisticalRobustness}

The preceding ablation studies (\Cref{sec:Ablations}) test the robustness of the taxonomy classification step. Here we report complementary robustness checks on the comparison methodology itself: effect sizes, popularity baselines, comparisons that equalize entity counts per paper, co-occurrence checks that separate structure from overall frequency shifts, paper-by-paper similarity tests, filtering of likely paper-specific entities (\Cref{sec:established_only}), broader provider groupings (\Cref{sec:provider_level1}), looser matching rules for model families (\Cref{sec:family_matching}), normalisation-threshold sensitivity (\Cref{sec:normalisation_sensitivity}), and a blinded cross-model audit of the extraction, classification, normalisation, and the step that asks whether an entity is pre-existing or paper-specific (\Cref{sec:annotation_validation}). Unless stated otherwise, the tables in this appendix are recomputed in the setting where the classifier also sees the generated pipeline and therefore function as robustness checks for the conclusions of \Cref{sec:taxonomy}, not as exact numeric copies of the main-text figures.

The section is organized by claim. \Cref{sec:effect_sizes,sec:popularity_baseline,sec:local_retrieval_baseline,sec:cardinality_matched} quantify the main taxonomy results, especially the strength and interpretation of provider divergence. \Cref{sec:residual_analysis,sec:content_sensitivity} address the structural and question-specificity claims that support the co-occurrence results. \Cref{sec:established_only,sec:provider_level1,sec:family_matching,sec:normalisation_sensitivity} test whether vocabulary compression and provider concentration survive stricter or looser matching assumptions and changes to the fuzzy-clustering threshold. Finally, \Cref{sec:annotation_validation} audits the extraction, classification, normalisation, and established-versus-paper-specific labeling steps that underlie the automated pipeline.

\subsection{Effect sizes and bootstrapped confidence intervals}
\label{sec:effect_sizes}

The Jensen--Shannon divergence~\citep{lin1991divergence} between discrete distributions $P$ and $Q$ is
\begin{equation}
  \mathrm{JSD}(P \| Q) = \tfrac{1}{2}\, D_{\mathrm{KL}}(P \| M) + \tfrac{1}{2}\, D_{\mathrm{KL}}(Q \| M), \quad M = \tfrac{1}{2}(P + Q)
  \label{eq:jsd}
\end{equation}
where $D_{\mathrm{KL}}$ is the Kullback--Leibler divergence computed with base-2 logarithms, yielding JSD in bits bounded by $[0, 1]$. Cram\'{e}r's $V$~\citep{cramer1946mathematical} quantifies the strength of association in a $r \times c$ contingency table:
\begin{equation}
  V = \sqrt{\frac{\chi^2}{n \cdot (\min(r, c) - 1)}}
  \label{eq:cramers_v}
\end{equation}
where $n$ is the sample size and $\chi^2$ the chi-square statistic. As a rough guide, $V < 0.10$ is conventionally considered small and $V > 0.30$ large~\citep{cohen1988statistical}, though these thresholds are approximate when $\min(r,c) > 2$.

\Cref{tab:effect_sizes} reports JSD with bootstrapped 95\% confidence intervals~\citep{efron1993introduction} (2,000 paper-level resamples, percentile method, seed\,$=$\,42) and Cram\'{e}r's $V$ for all 15 taxonomy dimensions $\times$ 3 LLMs in the same richer labeling setting where the classifier also sees the generated pipeline. Provider exhibits the largest effect size ($V = 0.33$--$0.35$, medium-to-large) and the largest JSD ($0.101$--$0.138$), with narrow confidence intervals confirming precise estimation. Most other dimensions have $V < 0.15$ (small effect). Dataset size ($V = 0.16$--$0.21$) and model architecture ($V = 0.13$--$0.17$) are, by Cram\'{e}r's $V$, the only other dimensions approaching medium effect sizes. Note that the ranking by $V$ differs slightly from the no-pipeline JSD ranking in \Cref{sec:taxonomy}, where dataset size and model size rank second and third; this reflects the sensitivity of each measure to sample size and number of categories. Because the underlying contingency tables tabulate label instances that are multi-label and nested within papers and entities, the $\chi^2$ p-values reported in \Cref{tab:effect_sizes} are descriptive rather than fully inferential; statistical inference rests on the JSD bootstrap CIs (paper-level resampling), Cram\'{e}r's $V$, and the robustness analyses in \Cref{sec:popularity_baseline,sec:local_retrieval_baseline,sec:cardinality_matched,sec:established_only,sec:provider_level1,sec:family_matching,sec:normalisation_sensitivity}.

\subsection{Popularity baseline comparison}
\label{sec:popularity_baseline}

Using the same richer labeling setting, we test whether LLM category distributions can be explained by simply recommending popular entities by comparing each LLM's divergence from the reference inventory against two popularity baselines (\Cref{tab:popularity_baseline}). For each paper, we construct a leave-one-out popularity distribution from all other reference papers (excluding the target paper to prevent corpus leakage) and ``suggest'' $k$ entities, where $k$ matches the number of entities that paper's LLM suggested. The deterministic top-$k$ baseline always assigns the $k$ most frequent entities; the stochastic sampled baseline draws $k$ entities without replacement with probability proportional to reference-corpus frequency (seed\,$=$\,42). \Cref{tab:popularity_baseline} reports the three dimensions emphasized in the main text: model provider, model size, and metric evaluation type. In all three cases, LLMs produce substantially lower JSD than the deterministic top-$k$ baseline, demonstrating that their suggestions are conditioned on the research question rather than defaulting to a fixed popularity ranking. The stochastic sampled baseline achieves near-zero JSD because drawing entities proportionally to reference-corpus frequency naturally reconstructs the aggregate taxonomy distribution; the LLMs' higher divergence relative to this floor indicates systematic distributional biases beyond what popularity sampling would produce.

\subsection{Local retrieval calibration baseline}
\label{sec:local_retrieval_baseline}

As a complementary calibration, we replace global popularity with a simple content-conditioned non-generative adviser. For each target paper, we build a leave-one-out BM25 neighborhood over the generated research questions and retrieve the top $N=25$ most similar reference papers within the same reference--LLM shared subset. We then aggregate dataset, model, or metric names across that local neighborhood and emit the deterministic top $k$ entities, where $k$ matches the number of entities suggested by the LLM for that paper. \Cref{tab:local_retrieval_baseline} reports the same three dimensions emphasized in the main text. This local retrieval baseline is far closer than global top-$k$ popularity on all three dimensions, showing that question-conditioned lexical retrieval already captures a large share of the signal available under the sparse research-question-only input. However, it remains more divergent than the LLMs on model provider (BM25 JSD\,$=$\,0.245--0.290 versus 0.101--0.138 for the LLMs), model size (0.071--0.089 versus 0.014--0.036), and metric evaluation type (0.059--0.084 versus 0.011--0.021). The calibration therefore sharpens, rather than replaces, the popularity result: the main findings are not reducible either to global popularity or to a simple question-conditioned lexical retriever.

\subsection{Equal-count comparisons}
\label{sec:cardinality_matched}

Again in the same richer labeling setting, LLMs and the reference inventory may differ in the number of entities per paper, which could inflate divergence estimates. These matched-count comparisons use the reference--LLM shared-paper subsets for the relevant entity type (datasets n$=904/891/910$, models n$=942/928/948$, metrics n$=937/924/943$ for GPT-5.1/Gemini~3~Pro/DeepSeek-V3.2). To control for this, we subsample the larger entity set to match the smaller per paper (1,000 iterations, seed\,$=$\,42) and recompute JSD on the matched distributions. Matched JSD values are slightly lower than full estimates but preserve the same divergence hierarchy: provider remains the dominant model dimension under matched counts (0.088--0.112 across LLMs), followed by model size (0.011--0.028).

\subsection{Co-occurrence structure after accounting for overall frequencies}
\label{sec:residual_analysis}

To disentangle structural co-occurrence changes from marginal frequency shifts, we compute Pearson residuals for each cell of the co-occurrence matrix. We first exclude rows and columns with marginal totals below~5 to ensure adequate expected cell counts. For the remaining cells, the expected count under independence is
\begin{equation}
  E_{ij} = R_i \, C_j \,/\, N
  \label{eq:expected}
\end{equation}
where $R_i$ and $C_j$ are the row and column marginal totals and $N$ is the grand total. The Pearson residual~\citep{pearson1900criterion} is
\begin{equation}
  e_{ij} = (O_{ij} - E_{ij}) \,/\, \sqrt{E_{ij}}
  \label{eq:residual}
\end{equation}
which is approximately standard-normal under the null hypothesis of independence. Cell-level $p$-values are obtained from the two-sided normal distribution and corrected for multiple comparisons using the Benjamini--Hochberg procedure~\citep{benjamini1995controlling} at $\alpha = 0.05$, applied independently within each co-occurrence matrix. Note that marginal filtering on totals $\geq 5$ does not guarantee $E_{ij} \geq 5$ in every cell, so the normal approximation remains approximate for sparse cells.

\Cref{tab:residual_summary} reports the Pearson correlation between reference-inventory and LLM residual matrices (on shared rows and columns), the number of significant cells in each source, and the number of sign flips among cells significant in at least one source. \Cref{fig:residual_tasktype,fig:residual_evtype} show the full residual heatmaps.

\subsection{Paper-by-paper specificity tests}
\label{sec:content_sensitivity}

To assess whether LLM suggestions track individual paper content rather than defaulting to generic recommendations, we conduct two complementary tests on pairwise reference--LLM shared-paper subsets.

\paragraph{Same-paper vs.\ shuffled-paper similarity.} For each shared paper, we build a taxonomy profile (a count vector over all taxonomy categories for that entity type) from both the reference inventory and the LLM, then compute cosine similarity $\cos(\mathbf{v}^{\mathrm{ref}}, \mathbf{v}^{\mathrm{LLM}}) = \mathbf{v}^{\mathrm{ref}} \cdot \mathbf{v}^{\mathrm{LLM}} \,/\, (\|\mathbf{v}^{\mathrm{ref}}\| \, \|\mathbf{v}^{\mathrm{LLM}}\|)$, set to $0$ when either vector has zero norm. We compare the mean same-paper similarity $S_{\mathrm{same}}$ against a shuffled baseline $S_{\mathrm{shuffled}}$ obtained by randomly permuting the mapping between reference-inventory and LLM paper indices (1,000 permutations, seed\,$=$\,42). The one-sided $p$-value is the fraction of permutations where $S_{\mathrm{shuffled}} \ge S_{\mathrm{same}}$. Across all nine entity type $\times$ LLM combinations, $S_{\mathrm{same}}$ significantly exceeds $S_{\mathrm{shuffled}}$ ($p < 0.001$ in all cases; \Cref{tab:content_sensitivity}), with deltas ranging from 0.075 (models) to 0.266 (metrics). The pairwise shared-paper counts are n$=904/891/910$ for datasets, n$=942/928/948$ for models, and n$=937/924/943$ for metrics in the reference-versus-GPT-5.1/Gemini~3~Pro/DeepSeek-V3.2 comparisons. This provides strong evidence that LLM suggestions are conditioned on individual research questions, not generated from a fixed generic template.

\paragraph{Inter-paper entity Jaccard decomposition.} We decompose the increase in inter-paper entity overlap into a vocabulary-compression component and an LLM-specific excess. For each source, we sample 10,000 unordered paper pairs and compute the mean pairwise Jaccard index~\citep{jaccard1912distribution} $J(a,b) = |E_a \cap E_b| \,/\, |E_a \cup E_b|$ on entity name sets $E_a, E_b$ (within-source, lowercased). A random-draw baseline controls for vocabulary compression: for each paper, we draw $k_i$ entities (matching the paper's actual cardinality) without replacement from the source's corpus-wide frequency distribution (100 draws per pair). We define $J_{\mathrm{excess}}(s) = J_{\mathrm{actual}}(s) - J_{\mathrm{random}}(s)$ and $\Delta_{\mathrm{excess}} = J_{\mathrm{excess}}(\mathrm{LLM}) - J_{\mathrm{excess}}(\mathrm{ref})$, with paired bootstrap 95\% confidence intervals~\citep{efron1993introduction} (2,000 resamples of paper pairs, percentile method). For datasets, $\Delta_{\mathrm{excess}}$ is near zero overall: it is indistinguishable from zero for Gemini~3~Pro and DeepSeek-V3.2 and small but positive for GPT-5.1, indicating that most of the increased inter-paper overlap is explained by vocabulary compression rather than a large additional homogenisation term. For models, $\Delta_{\mathrm{excess}} > 0$ (0.004--0.012), reflecting genuine excess homogenisation consistent with provider concentration. For metrics, the pattern is mixed: GPT-5.1 and Gemini~3~Pro show negative $\Delta_{\mathrm{excess}}$ (more content-specific than the reference inventory relative to their vocabulary), while DeepSeek-V3.2 shows modest positive excess.

\subsection{Long-tail sensitivity: robustness after removing low-frequency entities}
\label{sec:established_only}

Many reference-inventory entities appear in only one paper, some because they are genuinely paper-specific and others because they are established but niche. To test whether the main findings survive once low-frequency entities are removed from the reference set, we apply simple rule-based filters to the reference entity set while leaving LLM suggestions unfiltered. These are long-tail sensitivity analyses, not principled fairness corrections: the singleton filter has only 7.5\% precision as a proxy for paper-specificity (\Cref{tab:annotation_heuristic_calibration}), and the introducedness audit on which it is calibrated has weak inter-model agreement ($\kappa = 0.108$).

We define three rule-of-thumb proxies for paper-introduced entities: (1)~singleton filter: entities appearing in exactly one paper in the reference corpus; (2)~title-match filter: entities whose name appears in the focal paper's title; (3)~combined filter: entities satisfying both criteria. These are rarity proxies, not perfect tests of whether an entity is truly new to a paper, so they may remove genuinely established but niche entities. We evaluate all three filters, but the manuscript table focuses on singleton exclusion because the title-match and combined filters remove little and change the results negligibly. Because these robustness analyses compare the reference inventory against all three LLMs under common inclusion rules, they use the all-three-LLM shared-paper subsets: n$=878$ for datasets, n$=915$ for models, and n$=911$ for metrics. We calibrate these rules with a blinded cross-model audit over a stratified sample of 300 entity--paper pairs using Claude Opus~4.6 and GPT-5.4 (\Cref{sec:annotation_validation,tab:annotation_introducedness,tab:annotation_heuristic_calibration}).

\paragraph{Entity-name recall.} After excluding singletons (the most aggressive filter), recall improves substantially: dataset recall rises from 15.9\% to 42.7\% (GPT-5.1), 16.2\% to 43.2\% (Gemini~3~Pro), and 12.7\% to 35.9\% (DeepSeek-V3.2). Model recall improves from 6.8\% to 20.7\% (GPT-5.1), 7.9\% to 24.3\% (Gemini~3~Pro), and 5.1\% to 16.9\% (DeepSeek-V3.2). Metric recall improves from 14.1\% to 42.6\% (GPT-5.1), 13.1\% to 42.2\% (Gemini~3~Pro), and 12.2\% to 40.2\% (DeepSeek-V3.2). This indicates that much of the apparent coverage gap is driven by singleton or otherwise rare entities. However, singleton exclusion does not improve paper-level model coverage: model zero-coverage rises slightly from 65.1/65.2/83.5\% to 66.6/67.1/85.2\% for GPT-5.1, Gemini~3~Pro, and DeepSeek-V3.2, respectively. The singleton filter therefore mainly shrinks the target universe rather than making more papers recoverable. The title-match and combined filters have minimal impact (removing 0.3--3.3\% of entities depending on entity type), consistent with most paper-specific entities not appearing in titles.

\paragraph{Provider divergence persists.} Provider JSD changes only modestly after removing singletons: it decreases for GPT-5.1 ($0.111 \to 0.092$) and Gemini~3~Pro ($0.102 \to 0.090$) but edges up for DeepSeek-V3.2 ($0.137 \to 0.140$), confirming that provider concentration persists even under this aggressive filter for paper-specific items. The residual divergence remains substantial and follows the same pattern of Major commercial overrepresentation documented in \Cref{sec:taxonomy}. \Cref{tab:established_robustness} reports full results for all entity types and filter levels; \Cref{fig:robustness_dashboard}a provides a visual comparison.

\subsection{Broader provider grouping}
\label{sec:provider_level1}

The original provider taxonomy uses fine-grained categories for major commercial providers but aggregates all academic and independent models into a single ``Other/Academic'' category. This asymmetry may amplify the apparent concentration of LLM suggestions around commercial providers. To address this, we introduce a broader four-way grouping: (1)~Major commercial (OpenAI, Meta~AI, Google DeepMind, Anthropic, Alibaba/Qwen, DeepSeek, Mistral~AI); (2)~Other commercial (Cohere, Hugging Face, Stability~AI, Microsoft Research, NVIDIA/NeMo, Databricks/MosaicML); (3)~Academic/Community reused (Other/Academic entities appearing in 2+ papers); (4)~Other/Academic singleton (Other/Academic entities appearing in exactly one paper after the same deterministic normalisation and fuzzy clustering used in the main entity-level analysis). This frequency-based split therefore preserves the paper's canonical model-identity rule and requires no additional audit labels.

Under this broader grouping, the reference inventory is 54.9\% Major commercial, 27.7\% Other/Academic singleton, 14.5\% Academic/Community reused, and 2.9\% Other commercial. LLM suggestions overrepresent Major commercial ($+$17 to $+$19~pp excess) at the expense of Other/Academic singletons ($-$23.5 to $-$24.3~pp). Academic/Community reused is moderately overrepresented ($+$3.6 to $+$5.8~pp), suggesting that LLMs slightly favour established academic models over Other/Academic singletons. The JSD under this broader grouping (0.082--0.092) is lower than under the original provider taxonomy (0.102--0.137), indicating that some of the original divergence reflects the asymmetric granularity of commercial versus academic provider categories; however, the remaining divergence remains substantial. The cleaner interpretation is therefore commercial concentration combined with long-tail suppression, not a wholesale disappearance of reused academic/community models. Full distributions and excess shares are reported in \Cref{tab:provider_level1}; \Cref{fig:robustness_dashboard}b visualises the broader grouping.

\subsection{Credit for naming the right model family or provider}
\label{sec:family_matching}

Exact-name matching is the harshest possible evaluation: if the reference inventory uses ``Llama-3-8B'' and the LLM suggests ``Llama-3-70B'', exact matching treats this as a complete miss. We relax matching to three granularity levels: (1)~exact match: entity names must match after normalisation; (2)~family match: model names are mapped to families via regex patterns (e.g., all Llama variants $\to$ ``llama'', all GPT-4 variants $\to$ ``gpt4''); (3)~provider match: only the provider label must match.

Mean per-paper model recall improves substantially with relaxed matching. At the exact level, recall is 6.2\% (GPT-5.1), 6.5\% (Gemini~3~Pro), and 3.7\% (DeepSeek-V3.2). At the family level, recall rises to 28.2\%, 27.9\%, and 20.0\%. At the provider level (deduplicated per paper), recall reaches 53.1\%, 53.3\%, and 44.9\%. This demonstrates that LLMs often suggest the correct model family or provider even when they miss the exact version, size, or fine-tuned variant. The gap between family-level and provider-level recall (${\sim}25$~pp) quantifies the extent to which LLMs suggest models from the correct provider but from the wrong family within that provider's offerings. Note that these per-paper recall figures are not directly comparable to the corpus-level recall in \Cref{sec:established_only}, which counts unique entities across the entire corpus rather than averaging per-paper overlap. \Cref{tab:family_matching} reports mean and median recall at each level; \Cref{fig:robustness_dashboard}c summarises the pattern.

\subsection{Normalisation-threshold sensitivity}
\label{sec:normalisation_sensitivity}

The fuzzy-clustering threshold (token-sort ratio $\geq T$) governs how aggressively near-duplicate entity names are merged (\Cref{sec:normalisation}). At the default $T = 90$, the blinded audit finds 76\% merge precision (\Cref{tab:annotation_normalization}). To verify that headline metrics are not an artefact of this particular threshold, \Cref{tab:normalisation_sensitivity} reports vocabulary size, effective number, Gini coefficient, and zero-LLM-coverage for model entities across $T \in \{80, 85, 90, 95, 100\}$. As $T$ increases (stricter merging), vocabulary sizes grow monotonically because fewer names are merged, but the relative compression ratio (reference vocabulary to LLM vocabulary) and the ordering of all information-theoretic summaries remain stable across the full range. Dataset and metric results follow the same pattern (full data in the supplementary CSV).

\begin{table}[h]
\centering
\caption{\textbf{The vocabulary-compression hierarchy is stable across fuzzy-clustering thresholds, ruling out normalisation sensitivity as an artefact.} Vocabulary size, effective number $\exp(H)$, Gini coefficient, and zero-LLM-coverage for model entities across fuzzy-clustering thresholds $T \in \{80, 85, 90, 95, 100\}$. LLM columns show the range across GPT-5.1, Gemini~3~Pro, and DeepSeek-V3.2. Bold row: default threshold used throughout the manuscript. Dataset and metric results follow the same pattern (full data in supplementary CSV).}
\label{tab:normalisation_sensitivity}
\begin{adjustbox}{max width=\textwidth}
\begin{tabular}{r rr rr rr r}
\toprule
& \multicolumn{2}{c}{Vocabulary size} & \multicolumn{2}{c}{Effective $N$} & \multicolumn{2}{c}{Gini} & \\
\cmidrule(lr){2-3} \cmidrule(lr){4-5} \cmidrule(lr){6-7}
$T$ & Ref. & LLMs & Ref. & LLMs & Ref. & LLMs & Zero-cov (\%) \\
\midrule
80 & 2,931 & 347--528 & 839 & 47--81 & 0.60 & 0.77--0.80 & 92.3 \\
85 & 3,260 & 373--568 & 1039 & 56--92 & 0.57 & 0.76--0.79 & 90.9 \\
\textbf{90} & \textbf{3,546} & \textbf{387--594} & \textbf{1232} & \textbf{59--96} & \textbf{0.54} & \textbf{0.75--0.79} & \textbf{90.2} \\
95 & 3,797 & 403--617 & 1454 & 60--103 & 0.51 & 0.75--0.78 & 89.6 \\
100 & 3,874 & 404--622 & 1524 & 61--104 & 0.50 & 0.74--0.78 & 89.5 \\
\bottomrule
\end{tabular}
\end{adjustbox}
\end{table}

\subsection{Blinded two-model audit of the pipeline}
\label{sec:annotation_validation}

The pipeline uses GPT-5.1 for two automated steps, entity extraction and taxonomy classification, together with deterministic rules and fuzzy string clustering (threshold~90) for entity normalisation. We audit these steps with a blinded two-model protocol using Claude Opus~4.6 (\texttt{high}) in Claude Code and GPT-5.4 (\texttt{xhigh}) in OpenAI Codex as independent model raters, under the API data-handling procedure described in \Cref{sec:api_data_handling}. Both systems annotate stratified samples across four tasks: (1)~extraction validation (90~rows: 30~papers $\times$ 3~entity types), verifying whether pipeline-extracted entities genuinely appear in each paper's experiments; (2)~classification validation (180~sampled entities, stratified by source: reference inventory and LLM outputs; each entity is audited on all taxonomy dimensions applicable to its entity type, so the audit unit is the entity~$\times$~applicable-dimension judgment, and the per-dimension $n_{\mathrm{cons.}}$ and $n_{\mathrm{pair.}}$ counts in \Cref{tab:annotation_classification} sum across these entity--dimension pairs rather than across the 180 sampled entities), verifying taxonomy label assignments; (3)~normalisation validation (100~fuzzy-merge decisions, balanced between merged pairs and near-miss non-merged pairs with similarity scores 80--89), verifying merge correctness; and (4)~introducedness validation (300~entity--paper pairs: 120~model, 100~dataset, 80~metric, quota-sampled across frequency bands), classifying whether entities are pre-existing, newly introduced by the paper, or paper-specific derivatives of existing items. Inter-model reliability is measured with Cohen's~$\kappa$~\citep{cohen1960coefficient} (classification, normalisation, introducedness) and intraclass correlation~\citep{shrout1979intraclass} (extraction counts). We do not adjudicate disagreements; instead, final pipeline-result metrics are computed only on rows where both systems agree after multi-label answers are put into a consistent order, while agreement metrics use all rows where both systems provided usable labels.

\paragraph{Pipeline validation.} Extraction validation reports mean per-paper precision (correct / extracted), recall (correct / (correct $+$ missed)), F1, and hallucination rate on rows where both systems agree (\Cref{tab:annotation_extraction}). The audited taxonomy labels are drawn from the with-pipeline [WP] classifier outputs used throughout \Cref{sec:StatisticalRobustness}; for dimensions whose labels are sensitive to pipeline context (notably architecture and task type, \Cref{sec:websearch_ablation}), the reliability tiers below pertain to the [WP] branch, whereas provider labels barely move across branches (\Cref{sec:pipeline_ablation}), so provider tier assignments apply to both [EL] and [WP]. Classification validation reports exact-match accuracy against the original pipeline labels on those agreed rows, with $\kappa$ values by source stratum and dimension computed on all rows where both systems gave usable labels (\Cref{tab:annotation_classification}). Normalisation validation reports merge precision on merged pairs and error rate on the near-miss band on agreed rows, with overall $\kappa$ on the full sample (\Cref{tab:annotation_normalization}). Agreement is highest for normalisation ($\kappa = 0.898$), moderate across classification dimensions (median $\kappa = 0.660$), lower for extraction counts (overall ICC$(2,1) = 0.469$), and weakest for introducedness (overall $\kappa = 0.108$). \Cref{tab:annotation_summary} provides a compact reliability overview; conditional-on-consensus diagnostics (precision, recall, accuracy) are reported in the task-specific tables. These diagnostics describe only the subset of rows on which both auditors converged and serve as best-case cleanliness checks for agreed cases, not as corpus-level accuracy estimates. Consensus rates vary substantially: normalisation reaches 95\%, classification 77\%, and introducedness 79\%, but extraction consensus is only 59\% (53 of 90~rows), so the perfect precision/recall/F1 reported there applies to a consensus subset, not the full audit.

\paragraph{Four-tier validation of taxonomy dimensions.} The classification audit supports a four-tier interpretation of the 15~taxonomy dimensions based on consensus-row accuracy, inter-model $\kappa$, and sample size. Strong dimensions achieve the highest agreement: provider ($\kappa = 0.923$--$1.000$, accuracy 89.7--94.1\%) and openness ($\kappa = 1.000$, accuracy 94.1--96.8\%). Moderate dimensions meet the $\kappa \geq 0.5$ and accuracy $\geq$75\% thresholds in at least one source stratum but with narrower margins: modality (Reference: $\kappa = 0.756$, accuracy 85.7\%, $n_{\mathrm{pair.}} = 17$; LLM: $\kappa = 0.654$, accuracy 85.7\%, $n_{\mathrm{pair.}} = 44$), evaluation type (LLM: $\kappa = 0.588$, accuracy 89.7\%; Reference: $\kappa = 0.061$ on only $n = 17$ pairwise rows, a small-sample artifact), and linguistic scope (LLM: $\kappa = 0.660$, accuracy 85.0\%, $n_{\mathrm{pair.}} = 23$). Size is tentative: it meets accuracy and $\kappa$ thresholds in the reference strata (dataset Reference: 90.0\%, $\kappa = 0.642$; model Reference: 87.5\%, $\kappa = 1.000$) but the dataset LLM stratum drops to $\kappa = 0.406$, individual strata narrowly miss the sample-size floor ($n_{\mathrm{pair.}} = 8$--$13$ in the stronger strata), and model-side LLM size accuracy drops to 47.1\%; size-dependent claims therefore carry more uncertainty than provider or evaluation-type claims. The remaining dimensions are exploratory for distinct reasons: task type, domain, annotation, architecture, and training paradigm have consensus-row accuracy below 50\% in at least one stratum; granularity and data quality achieve high accuracy but fall below the $\kappa$ threshold ($\kappa = 0.25$--$0.37$ and $\kappa = 0.00$--$0.50$, respectively); and cognitive/affective has insufficient sample size ($n_{\mathrm{pair.}} \leq 6$). Analyses resting on exploratory dimensions (including task type $\times$ provider co-occurrence) provide descriptive context rather than definitive structural claims. This four-tier distinction is flagged explicitly in figure annotations and throughout the results narrative.

\paragraph{How well the singleton and title rules capture paper-specific entities.} The singleton and title-match rules from \Cref{sec:established_only} are proxies for paper-introduced entities. To calibrate them, each of the 300~entity--paper pairs receives an audit-confirmed label saying whether the entity is pre-existing reusable, paper-introduced, paper-specific derivative, or unclear. \Cref{tab:annotation_introducedness} reports the agreed label distribution; \Cref{tab:annotation_heuristic_calibration} reports the precision, recall, and specificity of those rules against the agreed labels. No agreed row remained labelled ``unclear,'' so the conservative (unclear $\to$ pre-existing) and liberal (unclear $\to$ paper-introduced) analyses coincide numerically. The sample is quota-based across frequency bands and entity types; raw percentages are sample-level estimates, not corpus-weighted distributions.

\paragraph{Mismatch type characterisation.} For each entity--paper pair, both systems also annotate the mismatch type that best characterises why the entity might be missed by LLM suggestions: alias/variant, same family, paper-specific, or established but absent. We treat this field as auxiliary context rather than a quantitative calibration target, because overall inter-model agreement is low ($\kappa = 0.040$).

\paragraph{Caveats.} This is a blinded model-assisted audit, not human manual validation. Claude Opus~4.6 relied more heavily on \texttt{domain-knowledge} tags in dataset and model classification, whereas GPT-5.4 more often marked fields \texttt{unresolved-after-review} or \texttt{paper-reviewed}; to avoid adjudicating across these evidence-use profiles, final pipeline metrics are reported only on consensus rows. The validation samples are designed to test pipeline decisions under stricter conditions, not to estimate corpus-level prevalence. Classification accuracy should be interpreted per source stratum. Extraction metrics are computed per paper and averaged over consensus rows. The introducedness sample is quota-based, so extrapolation to corpus-level distributions requires caution.

\paragraph{Model-swap robustness.}
We further test whether the headline provider-concentration finding is an artefact of using GPT-5.1 for extraction or classification through three supplementary robustness checks that are orthogonal to the blinded audit above.
First, a deterministic regex-based mapping (60+ hand-crafted rules, requiring no LLM) agrees with 92--98\% of GPT-5.1 provider labels among regex-classifiable model entities (56--78\% coverage per source) across all four corpus sources ($n = 1{,}000$ papers each for the reference inventory and GPT-5.1; $n = 998$ for DeepSeek-V3.2; $n = 981$ for Gemini~3~Pro).
Second, we re-extract entities from 94 papers using Claude Opus~4.6 with reconstructed full text, under the same non-public scientific TDM and API data-handling procedure described in \Cref{sec:copyright_tdm,sec:api_data_handling}; the resulting provider distribution is near-identical to the GPT-5.1 extraction (Spearman $\rho = 0.991$, $p < 0.0001$; mean Jaccard similarity on model names $= 0.62$).
Third, we re-classify the GPT-5.1-extracted entities from 200 papers with Claude Opus~4.6 using the identical taxonomy prompt. Cross-classifier agreement is almost perfect for provider ($\kappa = 0.961$, $n = 1{,}498$ model entities) and openness ($\kappa = 0.912$), substantial for evaluation type ($\kappa = 0.711$, $n = 996$ metric entities) and modality ($\kappa = 0.602$, $n = 659$ dataset entities), fair for architecture ($\kappa = 0.339$) and the tentative size dimension ($\kappa = 0.280$), and slight for training paradigm ($\kappa = 0.200$).
This gradient is broadly consistent with the reliability hierarchy established by the blinded audit: the strong dimensions remain strong under classifier substitution, while the exploratory and tentative dimensions show the same instability regardless of which model performs the classification.

\begin{table}[!htbp]
\centering
\caption{\textbf{Blinded cross-model audit: inter-model reliability.} Claude Opus 4.6 (\texttt{high}) in Claude Code and GPT-5.4 (\texttt{xhigh}) in OpenAI Codex independently audited stratified validation samples for the four pipeline stages. Inter-model agreement is computed on all pairwise-complete rows. The consensus rate shows the fraction of rows on which both auditors converged after label canonicalisation; conditional-on-consensus diagnostics (precision, recall, accuracy) are reported in the task-specific tables (\Cref{tab:annotation_extraction,tab:annotation_classification,tab:annotation_normalization,tab:annotation_introducedness}). For classification, $n = 180$ refers to sampled entities; each entity contributes one judgment per applicable taxonomy dimension, so the per-dimension counts in \Cref{tab:annotation_classification} sum across entity~$\times$~dimension pairs.}
\label{tab:annotation_summary}
\begin{adjustbox}{max width=\textwidth}
\footnotesize
\begin{tabular}{l l r l l r}
\toprule
Task & Scope & $n$ & Agreement metric & Inter-model agreement & Consensus rate (\%) \\
\midrule
Extraction & 30 papers $\times$ 3 entity types & 90 & ICC(2,1) & 0.469 & 58.9 \\
Classification & 180 sampled entities spanning 15 taxonomy dimensions & 180 & Median $\kappa$ & 0.660 & 77.0 \\
Normalisation & 100 entity pairs (merged + near-miss) & 100 & $\kappa$ & 0.898 & 95.0 \\
Introducedness & 300 entity--paper pairs & 300 & $\kappa$ & 0.108 & 79.3 \\
\bottomrule
\end{tabular}
\end{adjustbox}
\end{table}

\setcounter{figure}{0}
\renewcommand{\thefigure}{D\arabic{figure}}
\setcounter{table}{0}
\renewcommand{\thetable}{E\arabic{table}}

\clearpage
\section{Supplementary figures}
\label{sec:supplementary_figures}

\noindent\mbox{}

\begin{figure}[p]
\centering
\includegraphics[width=0.78\textwidth]{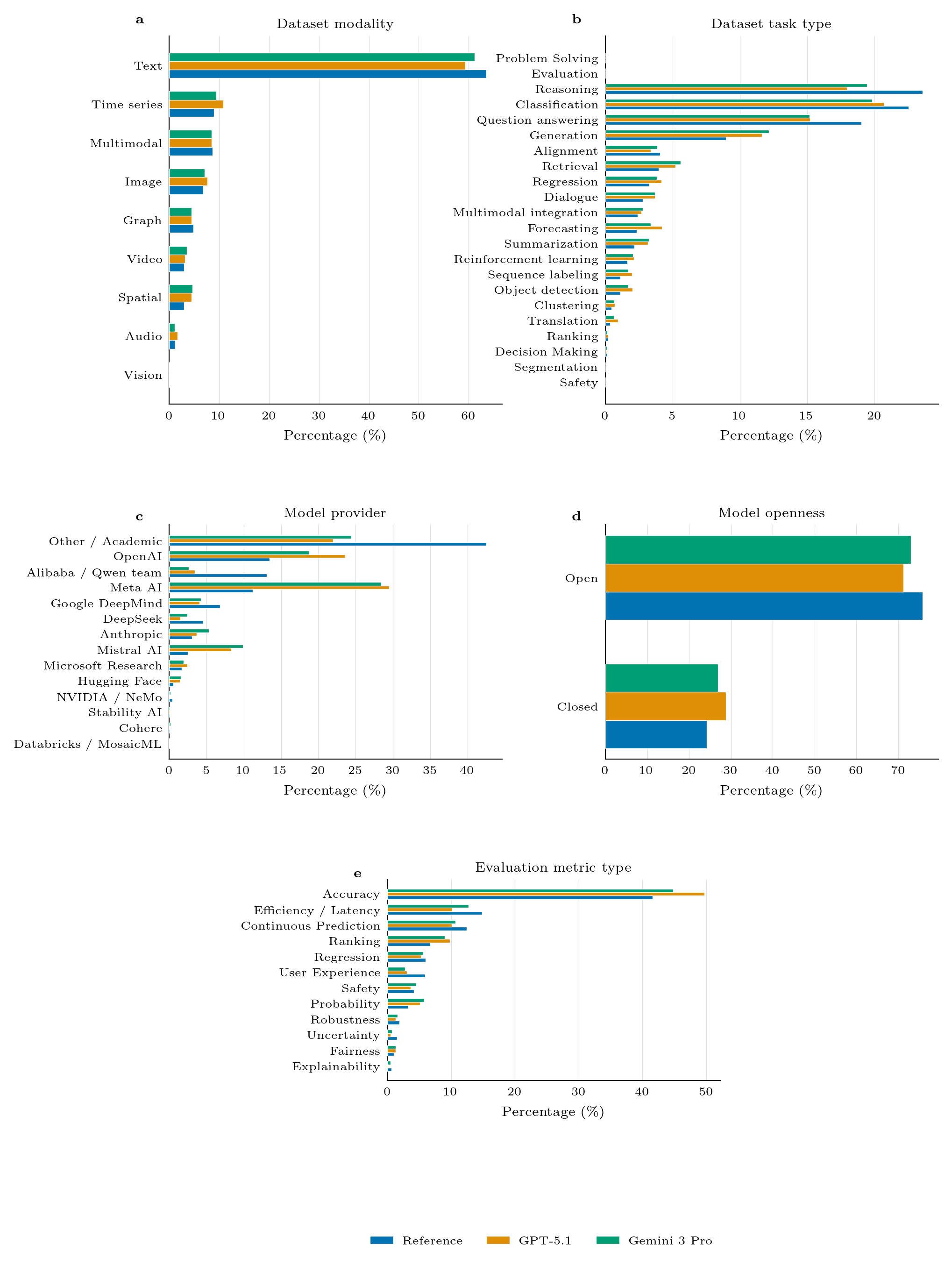}
\caption{\textbf{Across the five most-referenced taxonomy dimensions, the most pronounced LLM-versus-reference divergence is concentrated in model provider distributions~[EL].} Computed in the baseline setting where the classifier sees only entity lists, not the generated pipeline, using classified outputs from n$=1{,}000$ reference-inventory papers, 1{,}000 GPT-5.1 papers, 981 Gemini~3~Pro papers, and 998 DeepSeek-V3.2 papers. Grouped horizontal bar charts for five taxonomy dimensions: dataset modality (\textbf{a}), dataset task type (\textbf{b}), model provider (\textbf{c}), model openness (\textbf{d}), and evaluation metric type (\textbf{e}). For visual clarity, only GPT-5.1 and Gemini~3~Pro are shown; DeepSeek-V3.2 exhibits similar patterns. In this baseline, LLMs consistently overweight the top combined model-size bucket (GPT-5.1 59.1\%, Gemini~3~Pro 58.3\%, DeepSeek-V3.2 63.2\%, vs.\ 43.4\% in the reference inventory), and metric evaluation type shifts toward accuracy-like metrics (reference 41.6\%; LLMs 44.8--55.9\%) while user-experience and efficiency metrics decline. Within-dataset category shifts are modest: self-supervised annotation rises from 4.7\% to 7.6--8.1\%, multilingual datasets from 4.9\% to 5.6--8.2\%, and the education domain falls from 11.2\% to $\sim$8.3\%. The take-away is that LLMs overweight major commercial providers relative to the paper corpus, with a Jensen--Shannon divergence about 3--5$\times$ larger than the next-largest taxonomy dimension (\Cref{tab:effect_sizes}).}
\label{fig:taxonomy_appendix}
\end{figure}

\FloatBarrier

\begin{figure}[p]
\centering
\includegraphics[width=0.85\textwidth]{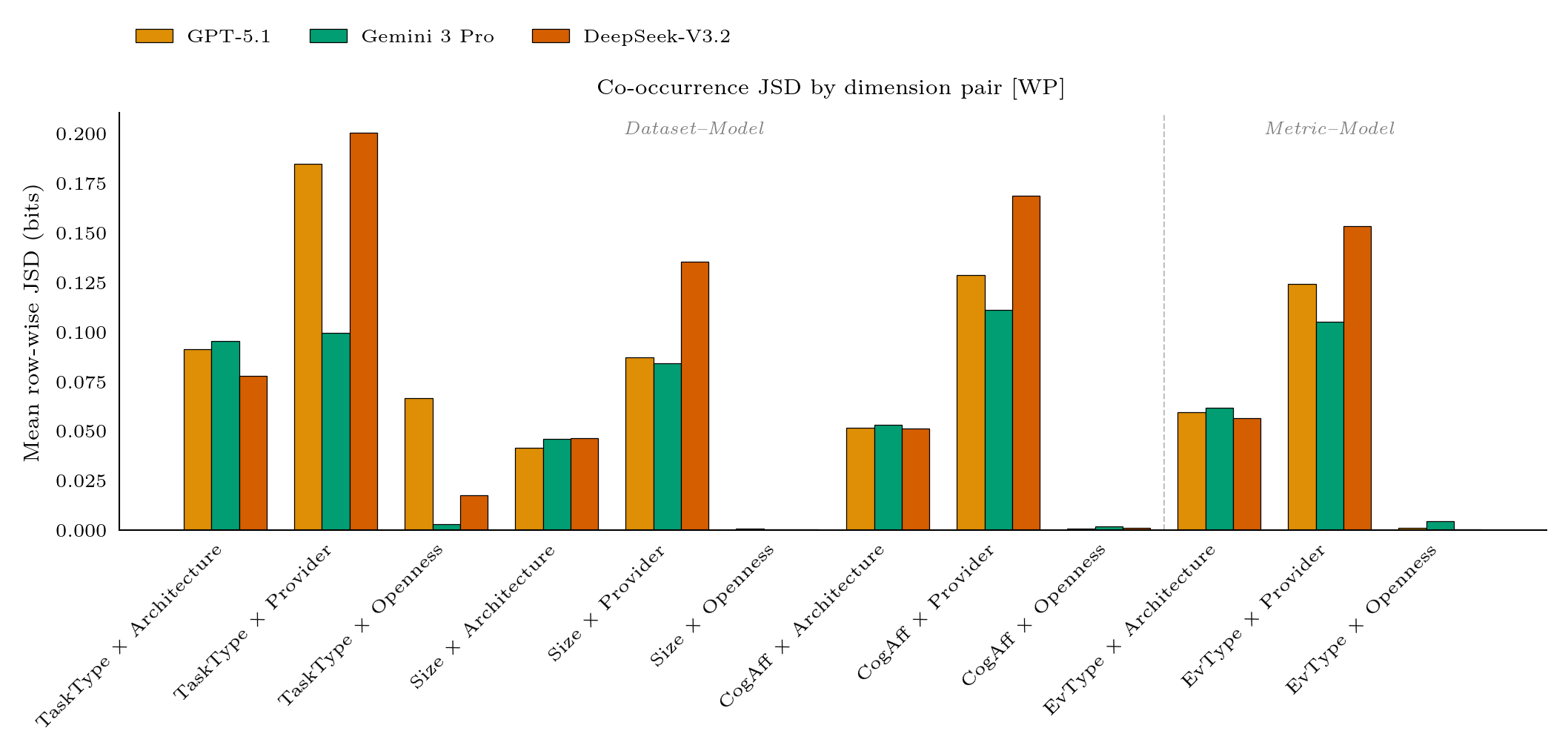}
\caption{\textbf{Provider-based co-occurrence pairs show the largest structural divergence between LLMs and the reference inventory, while openness-based pairs are the most faithfully preserved~[WP].} Computed from the with-pipeline classification branch. Source files contain n$=1{,}000$ reference-inventory papers, 1{,}000 GPT-5.1 papers, 981 Gemini~3~Pro papers, and 998 DeepSeek-V3.2 papers, but each co-occurrence pair uses the subset of papers with non-empty labels for both participating entity types, so support varies by pair and source. Mean row-wise Jensen--Shannon divergence between the reference-inventory and LLM co-occurrence matrices is shown for all 12 cross-entity taxonomy combinations: 9 dataset--model pairs (task type, size, and cognitive/affective $\times$ architecture, provider, and openness) and 3 metric--model pairs (evaluation type $\times$ architecture, provider, and openness). For the two focal provider pairs, evaluation type $\times$ provider has mean row-wise JSD of 0.124/0.105/0.153 (GPT-5.1/Gemini~3~Pro/DeepSeek-V3.2) and task type $\times$ provider of 0.185/0.100/0.201; other provider-based pairs show comparable divergence. Architecture- and task-type-based pairs should be read descriptively because those dimensions are label-sensitive or exploratory. Averaged across all 12 pairs, Gemini~3~Pro sits closest to the reference-inventory co-occurrence structure and DeepSeek-V3.2 farthest, but the ordering varies by pair.}
\label{fig:cooccurrence_jsd_appendix}
\end{figure}

\FloatBarrier

\begin{figure}[H]
\centering
\includegraphics[width=0.5\textwidth]{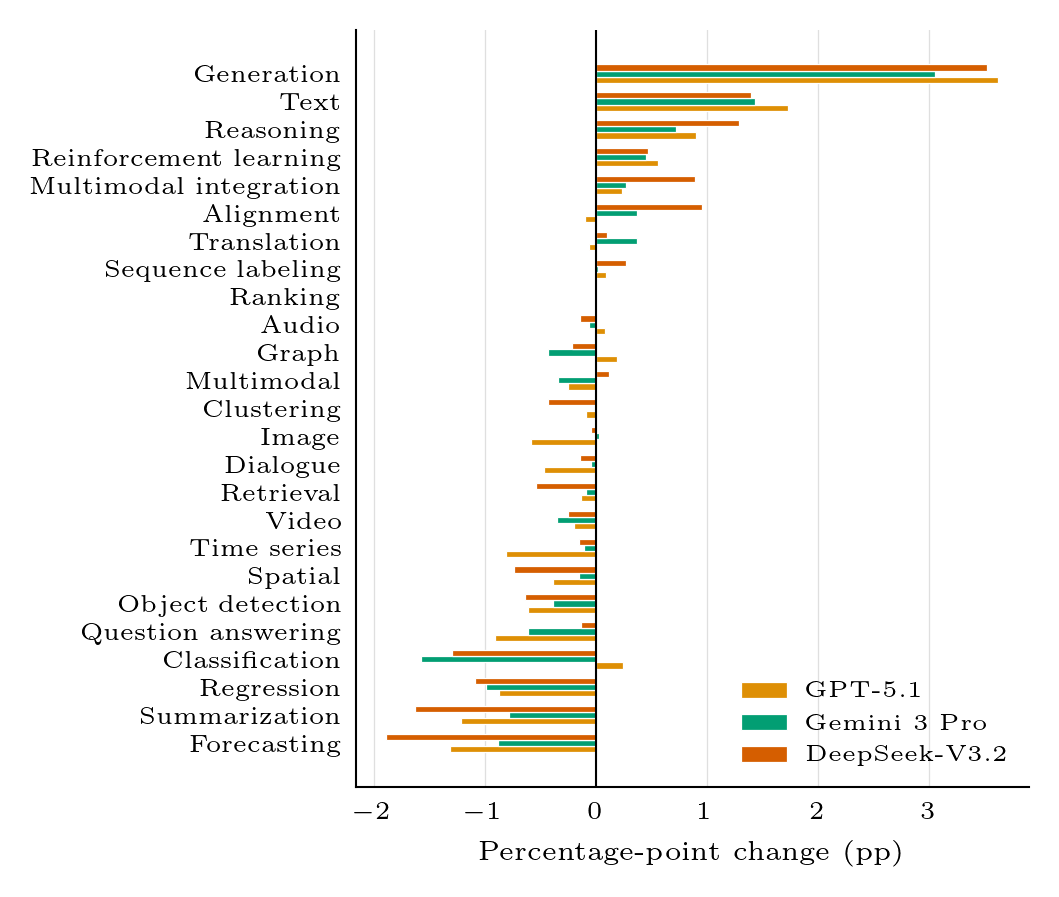}
\caption{\textbf{Pipeline context mainly relabels generation and document-level granularity categories, leaving most dataset dimensions nearly unchanged.} Computed on the classified suggestion outputs available for each model (n$=1{,}000$ GPT-5.1 papers, 981 Gemini~3~Pro papers, 998 DeepSeek-V3.2 papers). Percentage-point change in dataset category labels when the classifier sees the generated pipeline in addition to the entity lists, relative to the baseline where it sees only the entity lists, across three LLMs (GPT-5.1, Gemini~3~Pro, DeepSeek-V3.2); positive values indicate increased classification frequency with pipeline context. The shift is small in aggregate, indicating that the entity-list-only baseline in \Cref{sec:taxonomy} is not artificially dominated by labelling sensitivity.}
\label{fig:ablation}
\end{figure}

\begin{figure}[H]
\centering
\includegraphics[width=0.7\textwidth]{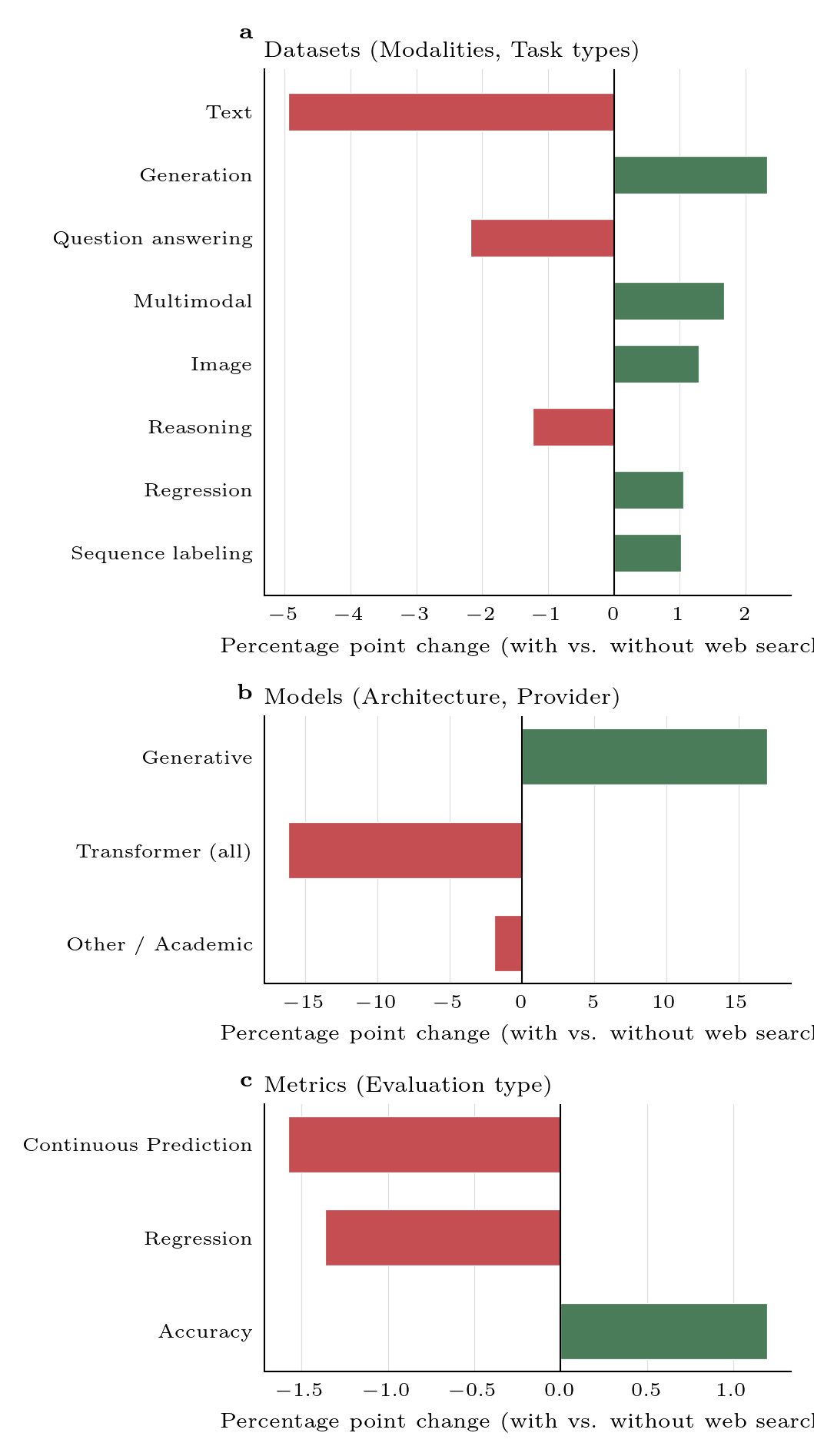}
\caption{\textbf{Enabling web search at classification time mainly relabels architecture categories without shifting provider, size, or evaluation-type distributions.} Computed on the 1,000 GPT-5.1 suggestion rows classified with and without web search. Percentage-point change in category labels when web search is enabled versus disabled, for datasets (\textbf{a}), models (\textbf{b}), and metrics (\textbf{c}); only subcategories with $|\Delta| \geq 1$~pp are shown. This confirms that architecture findings are label-sensitive and exploratory, whereas the provider-level claims in \Cref{sec:taxonomy} are stable under this ablation.}
\label{fig:websearch_ablation}
\end{figure}

\begin{figure}[H]
\centering
\includegraphics[width=\textwidth]{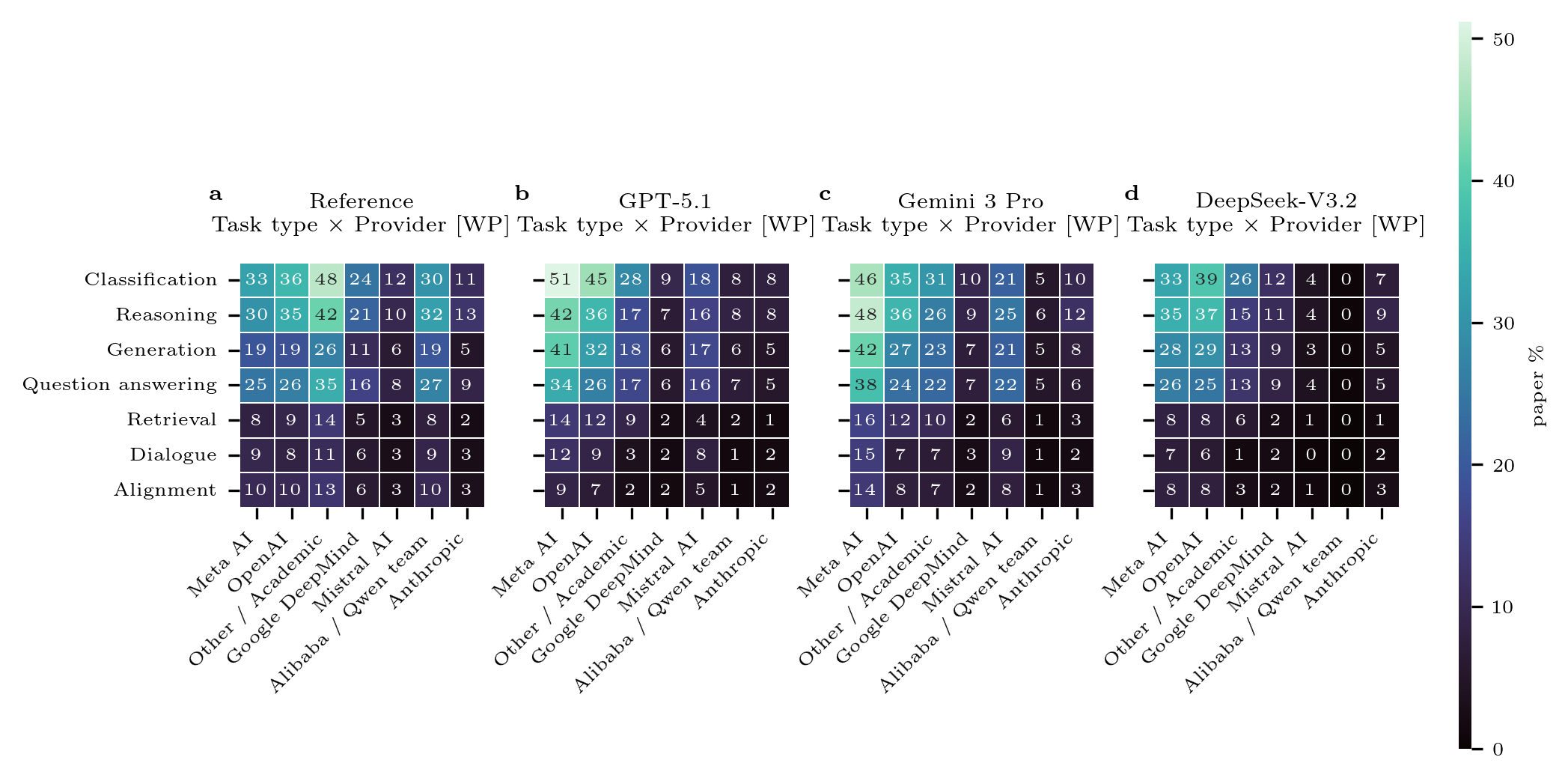}
\caption{\textbf{Across major task types, Meta~AI and OpenAI dominate in LLM suggestions whereas Other/Academic models dominate in the paper-derived reference inventory~[WP, exploratory].} Task-type labels are exploratory (10--32\% audit accuracy in the primary strata), so these panels provide descriptive context rather than definitive structural claims. Layout and conventions as in \Cref{fig:cooccurrence} but for task type $\times$ provider. In the reference inventory~(\textbf{a}), Other/Academic models dominate classification, reasoning, and question answering; in LLM suggestions~(\textbf{b}--\textbf{d}), this is inverted, with DeepSeek-V3.2 showing the strongest Other/Academic suppression. The fine-grained Other/Academic category aggregates reused academic/community models with the singleton-defined long tail (\Cref{sec:provider_level1}), so the co-occurrence shift partly reflects long-tail suppression rather than displacement of established academic models.}
\label{fig:cooccurrence_tasktype_appendix}
\end{figure}

\begin{figure}[H]
\centering
\includegraphics[width=\textwidth]{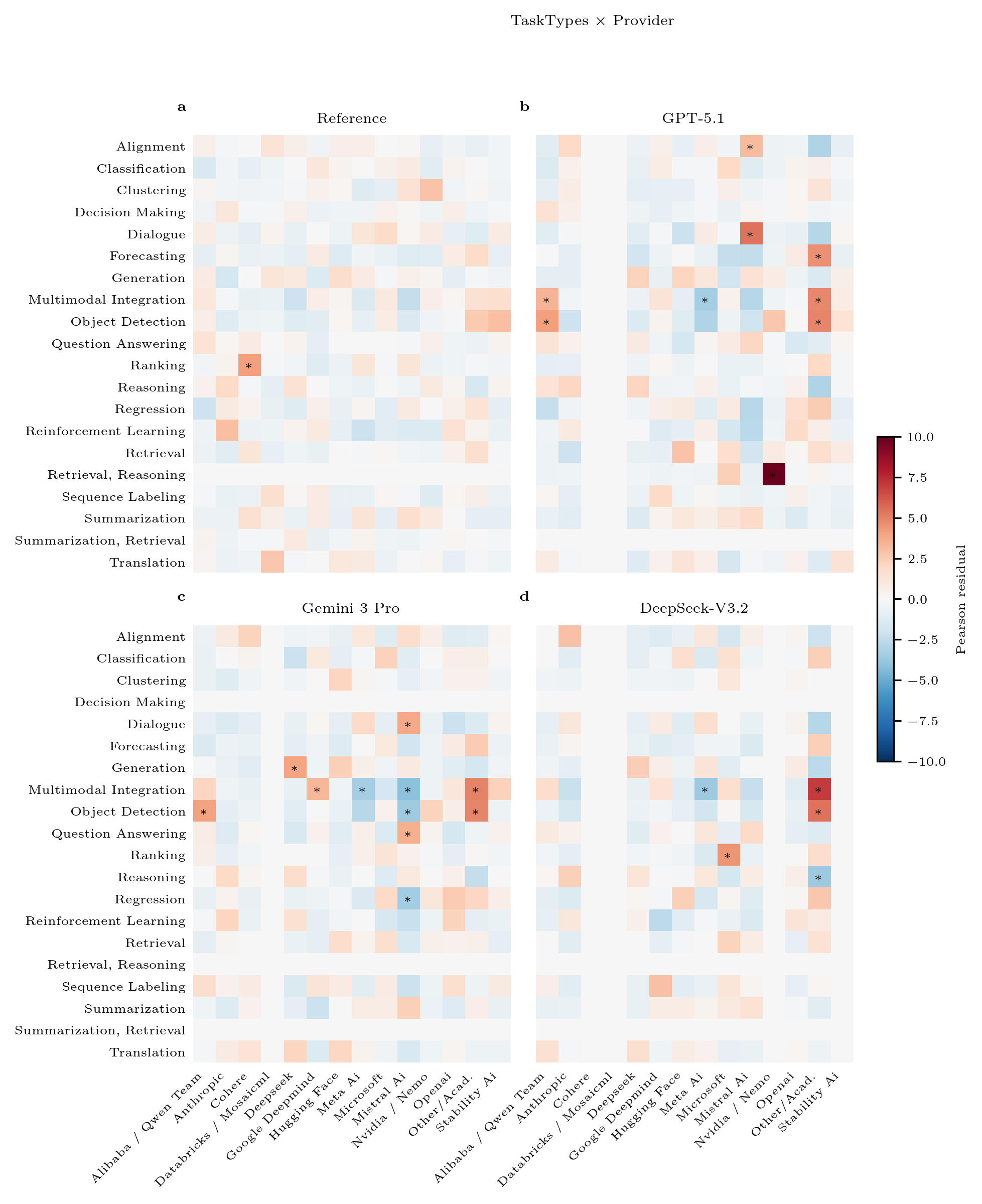}
\caption{\textbf{LLMs produce sharper task-type $\times$ provider residual patterns than the paper-derived reference inventory, indicating a narrower combinatorial landscape~[WP].} Residual matrices are built from pair-specific with-pipeline source subsets in the classified files (n$=1{,}000$ reference-inventory papers, 1{,}000 GPT-5.1 papers, 981 Gemini~3~Pro papers, 998 DeepSeek-V3.2 papers); residual correlations and significance counts after marginal filtering are reported in \Cref{tab:residual_summary}. Cells show how much each pairing appears above or below what would be expected from the overall row and column frequencies, for the reference inventory (\textbf{a}), GPT-5.1 (\textbf{b}), Gemini~3~Pro (\textbf{c}), and DeepSeek-V3.2 (\textbf{d}); asterisks mark FDR-significant cells ($q < 0.05$), positive (red) residuals indicate co-occurrence above expectation, and negative (blue) below.}
\label{fig:residual_tasktype}
\end{figure}

\begin{figure}[H]
\centering
\includegraphics[width=\textwidth]{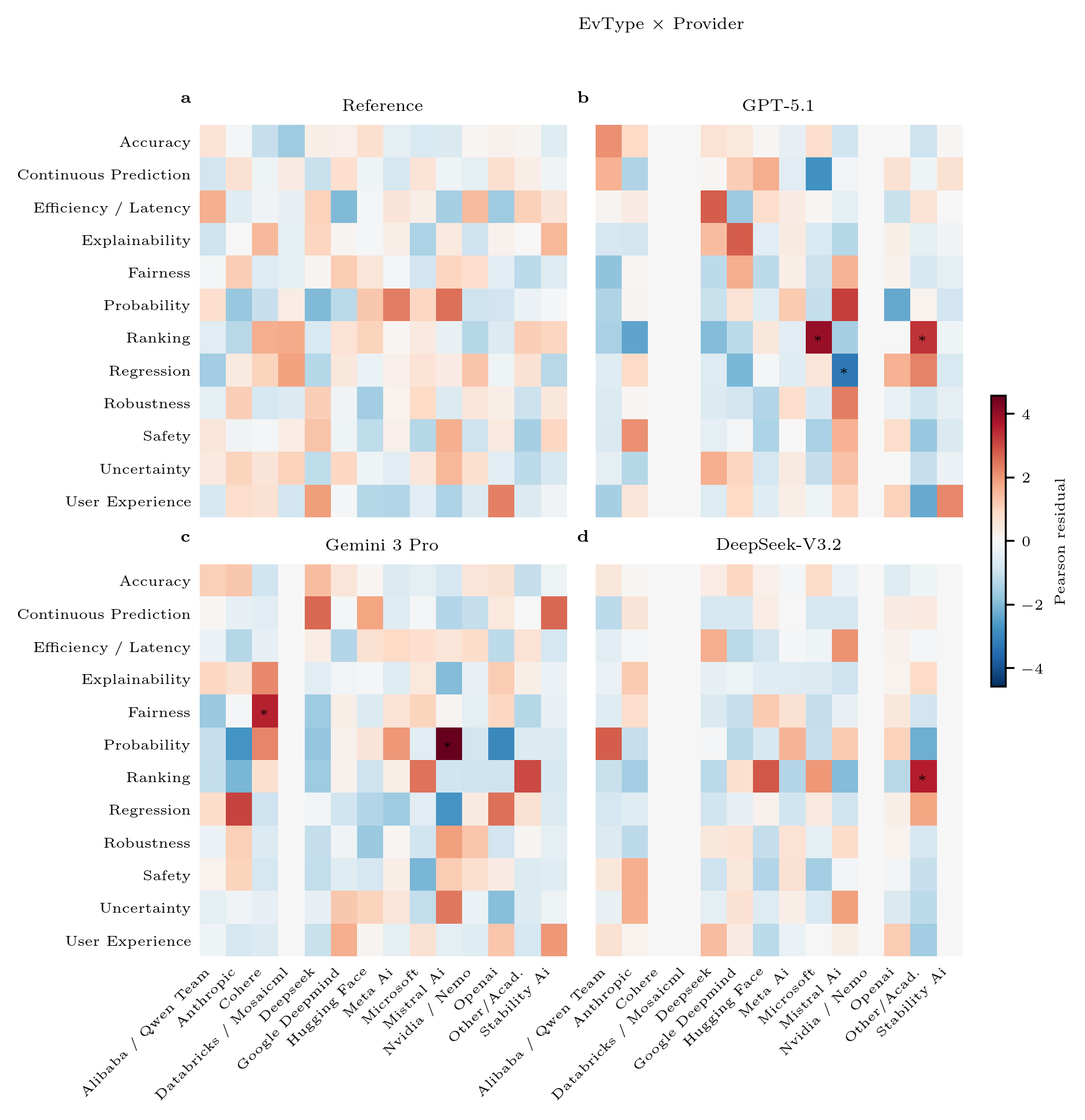}
\caption{\textbf{Evaluation-type $\times$ provider residuals are sparser and more concentrated in LLM suggestions than in the paper-derived reference inventory~[WP].} Layout as in \Cref{fig:residual_tasktype}, with the same source-count and residual-analysis conventions. The pattern of more significant residual cells in LLMs than in the reference inventory reinforces that LLMs narrow the combinatorial landscape around provider-centred pairings.}
\label{fig:residual_evtype}
\end{figure}

\begin{figure}[H]
\centering
\includegraphics[width=\textwidth]{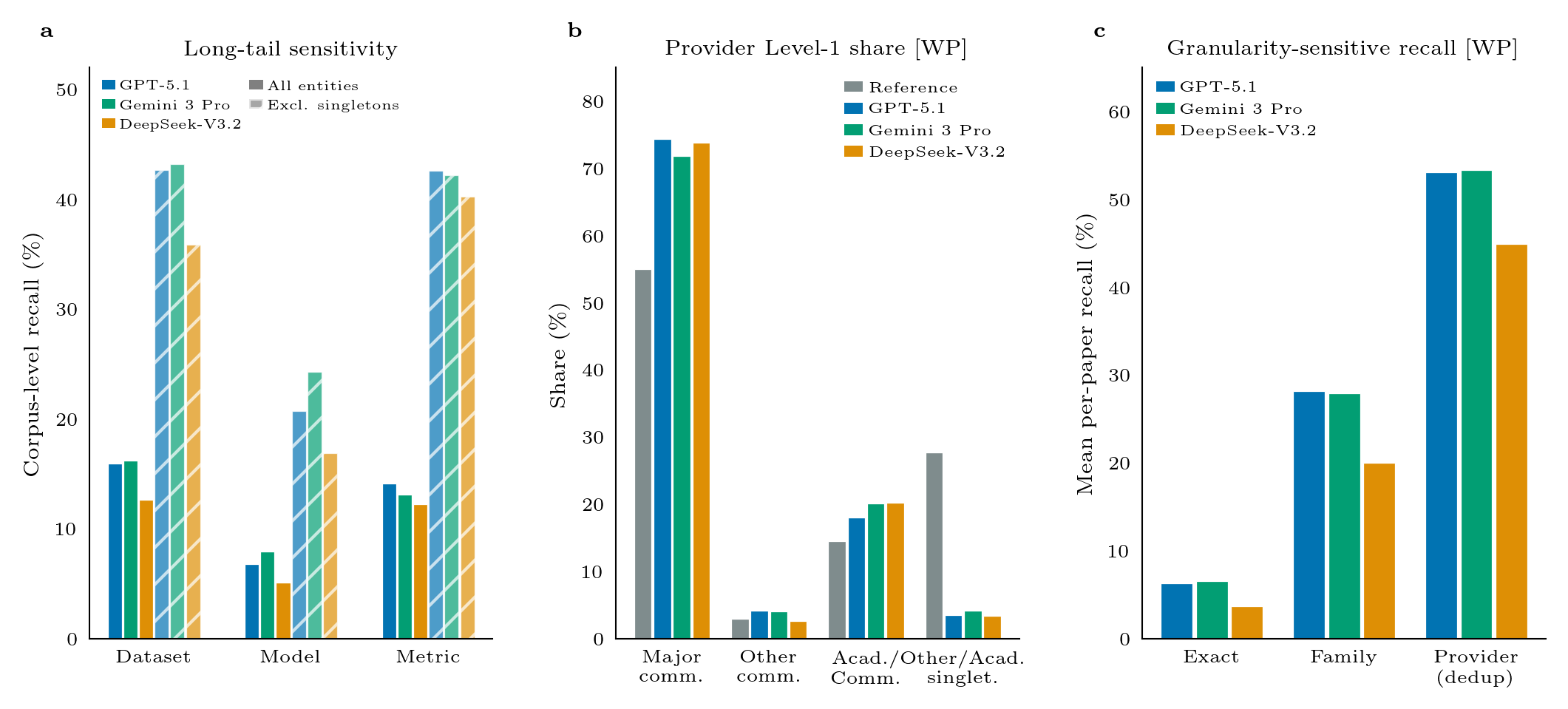}
\caption{\textbf{Provider concentration and long-tail suppression survive singleton exclusion, broader provider regrouping, and relaxed family/provider matching.} Panel \textbf{a} uses the all-three-LLM shared-paper subsets (datasets n$=878$, models n$=915$, metrics n$=911$); it compares entity names only and is therefore branch-independent. Panels \textbf{b} and \textbf{c} use the 915 shared model papers with taxonomy labels from the with-pipeline setting~[WP]. (\textbf{a})~Corpus-level recall before and after excluding singleton entities from the reference inventory; hatched bars show recall after singleton exclusion. (\textbf{b})~Provider distribution under the broader four-way provider grouping~[WP]; LLMs overrepresent major commercial providers and underrepresent Other/Academic singletons. (\textbf{c})~Mean per-paper model recall under exact-name, family-level, and provider-level matching (provider recall deduplicated per paper)~[WP]. The sharp jump from exact to family/provider matching shows that many apparent misses are granularity errors rather than complete neighbourhood failures.}
\label{fig:robustness_dashboard}
\end{figure}

\clearpage
\section{Supplementary tables}
\label{sec:supplementary_tables}

The tables below provide the detailed numerical counterparts to the appendix claims summarized above: first the label-sensitivity checks, then the effect-size tables, paper-by-paper similarity estimates, robustness filters, looser matching rules, normalisation-threshold sensitivity, and audit outputs.

\begin{table}[h!]
\centering
\caption{\textbf{Largest shifts in model and metric labels when the classifier sees the generated pipeline.} Computed on the classified suggestion outputs available for each model (n$=1{,}000$ GPT-5.1 papers, 981 Gemini~3~Pro papers, 998 DeepSeek-V3.2 papers). Each value represents the change in the share of a subcategory when the classifier sees the generated pipeline as well as the entity lists. Only subcategories where at least one LLM exhibits $|\Delta\text{pp}| \geq 2.0$ are shown. Bold values indicate $|\Delta\text{pp}| \geq 2.0$.}
\label{tab:pipeline_ablation_model_metric}
\begin{adjustbox}{max width=\textwidth}
\begin{tabular}{ll rrr}
\toprule
Dimension & Subcategory & GPT-5.1 $\Delta$pp & Gemini~3~Pro $\Delta$pp & DeepSeek-V3.2 $\Delta$pp \\
\midrule
Architecture & Generative & \textbf{+5.3} & +1.8 & -1.4 \\
Architecture & Transformer (all) & \textbf{-5.0} & -1.7 & +1.6 \\
\bottomrule
\end{tabular}
\end{adjustbox}
\end{table}

\begin{table}[H]
\centering
\caption{\textbf{Ablation conditions across models.} All three LLMs are evaluated with and without pipeline context. Web search is tested only for GPT-5.1.}
\label{tab:ablation_conditions}
\begin{tabular}{lccc}
\toprule
\textbf{Ablation} & \textbf{GPT-5.1} & \textbf{Gemini~3~Pro} & \textbf{DeepSeek-V3.2} \\
\midrule
Classification with pipeline context & Yes & Yes & Yes \\
Classification without pipeline context & Yes & Yes & Yes \\
Classification with web search & Yes & No & No \\
Classification without web search & Yes & No & No \\
\bottomrule
\end{tabular}
\end{table}

\begin{table}[H]
\centering
\caption{\textbf{Provider exhibits the largest Jensen--Shannon divergence and the largest Cram\'{e}r's $V$ among the 15 taxonomy dimensions. The provider JSD is about 3--5$\times$ the next-largest JSD; the Cram\'{e}r's $V$ margin is smaller.} Computed in the setting where the classifier also sees the generated pipeline. Pairwise reference--LLM shared-paper counts are n$=904/891/910$ for datasets, n$=942/928/948$ for models, and n$=937/924/943$ for metrics (GPT-5.1/Gemini~3~Pro/DeepSeek-V3.2). Jensen--Shannon divergence (JSD, base-2) with bootstrapped 95\% confidence intervals, Cram\'{e}r's $V$ effect sizes, and $\chi^2$ test statistics are reported for all category dimensions. $n$ denotes the total number of label instances in the contingency table. $\chi^2$ p-values are descriptive given the multi-label, paper-nested structure of the contingency tables; inferential weight rests on JSD with paper-level bootstrap CIs and the robustness analyses in this appendix.}
\label{tab:effect_sizes}
\begin{adjustbox}{max width=\textwidth}
\footnotesize
\begin{tabular}{llllrrrr}
\toprule
Entity & Dimension & LLM & JSD [95\% CI] & Cram\'{e}r's $V$ & $\chi^2$ & $p$ & $n$ \\
\midrule
Dataset & Modalities & GPT-5.1 & 0.003 [0.002, 0.007] & 0.058 & 29.2 & $2.13e-03$ & 8663 \\
 & TaskTypes & GPT-5.1 & 0.015 [0.012, 0.021] & 0.141 & 289.6 & $8.44e-46$ & 14565 \\
 & Domains & GPT-5.1 & 0.006 [0.005, 0.012] & 0.092 & 94.0 & $1.14e-12$ & 11028 \\
 & Annotation & GPT-5.1 & 0.002 [0.001, 0.005] & 0.053 & 31.4 & $5.16e-05$ & 11261 \\
 & Size & GPT-5.1 & 0.032 [0.023, 0.042] & 0.208 & 296.9 & $3.34e-65$ & 6838 \\
 & Granularity & GPT-5.1 & 0.006 [0.003, 0.014] & 0.091 & 57.9 & $3.36e-11$ & 7049 \\
 & Linguistic & GPT-5.1 & 0.002 [0.001, 0.005] & 0.055 & 20.6 & $3.45e-05$ & 6838 \\
 & CognitiveAffective & GPT-5.1 & 0.005 [0.003, 0.009] & 0.081 & 94.2 & $6.13e-14$ & 14457 \\
 & DataQuality & GPT-5.1 & 0.000 [0.000, 0.001] & 0.011 & 0.9 & $3.41e-01$ & 6843 \\
 & Modalities & Gemini~3~Pro & 0.002 [0.001, 0.007] & 0.048 & 19.3 & $2.29e-02$ & 8347 \\
 & TaskTypes & Gemini~3~Pro & 0.015 [0.013, 0.022] & 0.145 & 297.6 & $1.83e-48$ & 14235 \\
 & Domains & Gemini~3~Pro & 0.007 [0.005, 0.013] & 0.095 & 95.7 & $5.62e-13$ & 10719 \\
 & Annotation & Gemini~3~Pro & 0.004 [0.003, 0.008] & 0.076 & 63.0 & $3.76e-11$ & 10792 \\
 & Size & Gemini~3~Pro & 0.018 [0.012, 0.026] & 0.157 & 162.3 & $5.64e-36$ & 6581 \\
 & Granularity & Gemini~3~Pro & 0.006 [0.003, 0.014] & 0.089 & 53.9 & $2.19e-10$ & 6776 \\
 & Linguistic & Gemini~3~Pro & 0.002 [0.000, 0.004] & 0.046 & 14.2 & $8.25e-04$ & 6581 \\
 & CognitiveAffective & Gemini~3~Pro & 0.004 [0.003, 0.007] & 0.072 & 73.0 & $5.44e-10$ & 14143 \\
 & DataQuality & Gemini~3~Pro & 0.001 [0.000, 0.003] & 0.029 & 5.5 & $1.86e-02$ & 6583 \\
 & Modalities & DeepSeek-V3.2 & 0.002 [0.002, 0.008] & 0.054 & 21.5 & $1.77e-02$ & 7302 \\
 & TaskTypes & DeepSeek-V3.2 & 0.014 [0.012, 0.021] & 0.134 & 222.5 & $4.05e-33$ & 12311 \\
 & Domains & DeepSeek-V3.2 & 0.006 [0.005, 0.012] & 0.087 & 71.5 & $1.20e-08$ & 9375 \\
 & Annotation & DeepSeek-V3.2 & 0.002 [0.001, 0.004] & 0.046 & 20.2 & $5.23e-03$ & 9688 \\
 & Size & DeepSeek-V3.2 & 0.031 [0.022, 0.042] & 0.198 & 230.8 & $7.52e-51$ & 5891 \\
 & Granularity & DeepSeek-V3.2 & 0.007 [0.004, 0.015] & 0.095 & 54.3 & $1.82e-10$ & 6033 \\
 & Linguistic & DeepSeek-V3.2 & 0.001 [0.000, 0.002] & 0.026 & 4.0 & $1.35e-01$ & 5891 \\
 & CognitiveAffective & DeepSeek-V3.2 & 0.006 [0.004, 0.011] & 0.089 & 97.8 & $1.23e-14$ & 12223 \\
 & DataQuality & DeepSeek-V3.2 & 0.000 [0.000, 0.001] & 0.001 & 0.0 & $9.23e-01$ & 5893 \\
Model & Architecture & GPT-5.1 & 0.021 [0.015, 0.029] & 0.152 & 313.8 & $1.92e-61$ & 13591 \\
 & TrainingParadigm & GPT-5.1 & 0.007 [0.005, 0.010] & 0.093 & 369.6 & $5.84e-71$ & 42714 \\
 & Provider & GPT-5.1 & 0.112 [0.101, 0.124] & 0.353 & 1,405.4 & $1.05e-292$ & 11290 \\
 & Openness & GPT-5.1 & 0.002 [0.001, 0.004] & 0.045 & 22.4 & $2.16e-06$ & 11285 \\
 & Size & GPT-5.1 & 0.015 [0.010, 0.021] & 0.129 & 188.5 & $1.15e-41$ & 11269 \\
 & Architecture & Gemini~3~Pro & 0.024 [0.019, 0.033] & 0.166 & 368.5 & $2.89e-72$ & 13426 \\
 & TrainingParadigm & Gemini~3~Pro & 0.006 [0.004, 0.008] & 0.082 & 278.3 & $1.66e-52$ & 41707 \\
 & Provider & Gemini~3~Pro & 0.101 [0.091, 0.114] & 0.337 & 1,266.7 & $7.75e-263$ & 11136 \\
 & Openness & Gemini~3~Pro & 0.001 [0.000, 0.002] & 0.024 & 6.5 & $1.08e-02$ & 11130 \\
 & Size & Gemini~3~Pro & 0.014 [0.010, 0.020] & 0.125 & 173.5 & $2.13e-38$ & 11108 \\
 & Architecture & DeepSeek-V3.2 & 0.017 [0.013, 0.023] & 0.129 & 207.7 & $3.97e-39$ & 12448 \\
 & TrainingParadigm & DeepSeek-V3.2 & 0.006 [0.004, 0.008] & 0.076 & 224.1 & $7.37e-42$ & 39105 \\
 & Provider & DeepSeek-V3.2 & 0.138 [0.126, 0.153] & 0.334 & 1,177.6 & $1.17e-243$ & 10582 \\
 & Openness & DeepSeek-V3.2 & 0.009 [0.006, 0.013] & 0.096 & 97.7 & $4.87e-23$ & 10577 \\
 & Size & DeepSeek-V3.2 & 0.036 [0.028, 0.046] & 0.179 & 337.7 & $4.72e-74$ & 10555 \\
Metric & EvType & GPT-5.1 & 0.018 [0.014, 0.025] & 0.153 & 212.2 & $2.12e-39$ & 9048 \\
 & EvType & Gemini~3~Pro & 0.011 [0.008, 0.017] & 0.118 & 114.9 & $1.90e-19$ & 8243 \\
 & EvType & DeepSeek-V3.2 & 0.021 [0.017, 0.029] & 0.159 & 198.6 & $1.44e-36$ & 7868 \\
\bottomrule
\end{tabular}
\end{adjustbox}
\end{table}

\begin{table}[H]
\centering
\caption{\textbf{LLM suggestions fall between a popularity-proportional sampler and a deterministic top-$k$ ranker, confirming that they respond to the research question while still exhibiting distributional biases.} Computed on the with-pipeline classification branch using pairwise reference--LLM shared-paper subsets. Model-provider and model-size comparisons use n$=942/928/948$ shared papers for GPT-5.1/Gemini~3~Pro/DeepSeek-V3.2; metric evaluation type uses n$=937/924/943$. Jensen--Shannon divergence between the paper-derived reference inventory and each LLM (JSD$_{\mathrm{LLM}}$), a deterministic top-$k$ baseline (JSD$_{\mathrm{top\text{-}k}}$), and a stochastic popularity-sampled baseline (JSD$_{\mathrm{sampled}}$) is reported. Lower values indicate closer alignment with the reference inventory.}
\label{tab:popularity_baseline}
\begin{adjustbox}{max width=\textwidth}
\footnotesize
\begin{tabular}{llrrr}
\toprule
Dimension & LLM & JSD$_{\mathrm{LLM}}$ & JSD$_{\mathrm{top\text{-}k}}$ & JSD$_{\mathrm{sampled}}$ \\
\midrule
Model provider & GPT-5.1 & 0.1117 & 0.6423 & 0.0010 \\
 & Gemini~3~Pro & 0.1012 & 0.6535 & 0.0016 \\
 & DeepSeek-V3.2 & 0.1384 & 0.6915 & 0.0016 \\
Model size & GPT-5.1 & 0.0151 & 0.1571 & 0.0001 \\
 & Gemini~3~Pro & 0.0140 & 0.1562 & 0.0004 \\
 & DeepSeek-V3.2 & 0.0358 & 0.1819 & 0.0005 \\
Metric evaluation type & GPT-5.1 & 0.0178 & 0.2642 & 0.0031 \\
 & Gemini~3~Pro & 0.0114 & 0.3249 & 0.0042 \\
 & DeepSeek-V3.2 & 0.0211 & 0.3282 & 0.0049 \\
\bottomrule
\end{tabular}
\end{adjustbox}
\end{table}

\begin{table}[H]
\centering
\caption{\textbf{Local BM25 retrieval sharpens, rather than replaces, the popularity calibration: LLMs remain closer to the reference inventory than a content-conditioned retriever.} Computed on the with-pipeline classification branch using pairwise reference--LLM shared-paper subsets and leave-one-out BM25 retrieval over generated research questions with top-$N$ neighborhoods (N=25). Model-provider and model-size comparisons use n$=942/928/948$ shared papers for GPT-5.1/Gemini~3~Pro/DeepSeek-V3.2; metric evaluation type uses n$=937/924/943$. Jensen--Shannon divergence between the paper-derived reference inventory and each LLM (JSD$_{\mathrm{LLM}}$) and the deterministic local BM25 baseline (JSD$_{\mathrm{local\text{-}BM25}}$) is reported. Lower values indicate closer alignment with the reference inventory.}
\label{tab:local_retrieval_baseline}
\begin{adjustbox}{max width=\textwidth}
\footnotesize
\begin{tabular}{llrr}
\toprule
Dimension & LLM & JSD$_{\mathrm{LLM}}$ & JSD$_{\mathrm{local\text{-}BM25}}$ \\
\midrule
Model provider & GPT-5.1 & 0.1117 & 0.2455 \\
 & Gemini~3~Pro & 0.1012 & 0.2402 \\
 & DeepSeek-V3.2 & 0.1384 & 0.2902 \\
Model size & GPT-5.1 & 0.0151 & 0.0721 \\
 & Gemini~3~Pro & 0.0140 & 0.0707 \\
 & DeepSeek-V3.2 & 0.0358 & 0.0894 \\
Metric evaluation type & GPT-5.1 & 0.0178 & 0.0591 \\
 & Gemini~3~Pro & 0.0114 & 0.0728 \\
 & DeepSeek-V3.2 & 0.0211 & 0.0843 \\
\bottomrule
\end{tabular}
\end{adjustbox}
\end{table}

\begin{table}[H]
\centering
\caption{\textbf{LLMs preserve the sign of co-occurrence associations but sharpen them into a narrower pattern.} Residual correlations are computed on shared rows and columns after marginal filtering. The underlying co-occurrence matrices come from pair-specific with-pipeline source subsets in the classified files (n$=1{,}000$ reference-inventory papers, 1{,}000 GPT-5.1 papers, 981 Gemini~3~Pro papers, 998 DeepSeek-V3.2 papers). Pearson correlation of residual matrices between the paper-derived reference inventory and each LLM, number of FDR-significant cells, and sign flips among significant cells.}
\label{tab:residual_summary}
\begin{adjustbox}{max width=\textwidth}
\begin{tabular}{llrrrr}
\toprule
Co-occurrence & LLM & Residual $r$ & $n_{\mathrm{sig}}^{\mathrm{ref}}$ & $n_{\mathrm{sig}}^{\mathrm{LLM}}$ & Sign flips \\
\midrule
TaskTypes × Architecture & GPT-5.1 & 0.725 & 7 & 10 & 0 \\
 & Gemini~3~Pro & 0.746 & 7 & 9 & 0 \\
 & DeepSeek-V3.2 & 0.718 & 7 & 8 & 0 \\
TaskTypes × Provider & GPT-5.1 & 0.492 & 1 & 9 & 0 \\
 & Gemini~3~Pro & 0.409 & 1 & 11 & 3 \\
 & DeepSeek-V3.2 & 0.435 & 1 & 5 & 1 \\
TaskTypes × Openness & GPT-5.1 & 0.498 & 0 & 0 & 0 \\
 & Gemini~3~Pro & 0.372 & 0 & 0 & 0 \\
 & DeepSeek-V3.2 & 0.591 & 0 & 0 & 0 \\
Size × Architecture & GPT-5.1 & 0.786 & 0 & 0 & 0 \\
 & Gemini~3~Pro & 0.755 & 0 & 0 & 0 \\
 & DeepSeek-V3.2 & 0.454 & 0 & 0 & 0 \\
Size × Provider & GPT-5.1 & 0.193 & 0 & 0 & 0 \\
 & Gemini~3~Pro & 0.497 & 0 & 0 & 0 \\
 & DeepSeek-V3.2 & 0.375 & 0 & 0 & 0 \\
Size × Openness & GPT-5.1 & -0.091 & 0 & 0 & 0 \\
 & Gemini~3~Pro & 0.825 & 0 & 0 & 0 \\
 & DeepSeek-V3.2 & 0.732 & 0 & 0 & 0 \\
CogAff × Architecture & GPT-5.1 & 0.724 & 1 & 2 & 0 \\
 & Gemini~3~Pro & 0.680 & 1 & 2 & 0 \\
 & DeepSeek-V3.2 & 0.673 & 1 & 2 & 0 \\
CogAff × Provider & GPT-5.1 & 0.493 & 0 & 15 & 2 \\
 & Gemini~3~Pro & 0.498 & 0 & 7 & 1 \\
 & DeepSeek-V3.2 & 0.382 & 0 & 6 & 0 \\
CogAff × Openness & GPT-5.1 & 0.776 & 0 & 0 & 0 \\
 & Gemini~3~Pro & 0.717 & 0 & 0 & 0 \\
 & DeepSeek-V3.2 & 0.646 & 0 & 0 & 0 \\
EvType × Architecture & GPT-5.1 & 0.612 & 1 & 1 & 0 \\
 & Gemini~3~Pro & 0.515 & 1 & 1 & 0 \\
 & DeepSeek-V3.2 & 0.550 & 1 & 3 & 0 \\
EvType × Provider & GPT-5.1 & 0.322 & 0 & 3 & 1 \\
 & Gemini~3~Pro & 0.289 & 0 & 2 & 1 \\
 & DeepSeek-V3.2 & 0.414 & 0 & 1 & 0 \\
EvType × Openness & GPT-5.1 & 0.716 & 0 & 0 & 0 \\
 & Gemini~3~Pro & 0.726 & 0 & 0 & 0 \\
 & DeepSeek-V3.2 & 0.851 & 0 & 0 & 0 \\
\bottomrule
\end{tabular}
\end{adjustbox}
\end{table}

\begin{table}[H]
\centering
\caption{\textbf{LLM suggestions track individual paper content, while excess homogenisation beyond vocabulary compression concentrates in model suggestions.} Pairwise shared-paper counts are n$=904/891/910$ for datasets, n$=942/928/948$ for models, and n$=937/924/943$ for metrics in the reference-versus-GPT-5.1/Gemini~3~Pro/DeepSeek-V3.2 comparisons. \textit{Top:} Same-paper vs.\ shuffled taxonomy cosine similarity; $\Delta = S_{\mathrm{same}} - S_{\mathrm{shuffled}}$. \textit{Bottom:} Inter-paper entity Jaccard decomposition; $\Delta_{\mathrm{excess}}$ isolates homogenisation beyond vocabulary compression, with bootstrapped 95\% CIs. Negative $\Delta_{\mathrm{excess}}$ indicates LLMs are more content-specific than the paper-derived reference inventory relative to their vocabulary size.}
\label{tab:content_sensitivity}
\begin{adjustbox}{max width=\textwidth}
\footnotesize
\begin{tabular}{llrrrr}
\toprule
\multicolumn{6}{l}{\textit{Test 1: Same-paper vs.\ shuffled taxonomy similarity}} \\
\midrule
Entity type & LLM & $S_{\mathrm{same}}$ & $S_{\mathrm{shuffled}}$ & $\Delta$ & $p$ \\
\midrule
Datasets & GPT-5.1 & 0.721 & 0.496 & 0.225 & $<$0.001 \\
 & Gemini~3~Pro & 0.727 & 0.508 & 0.219 & $<$0.001 \\
 & DeepSeek-V3.2 & 0.716 & 0.493 & 0.223 & $<$0.001 \\
Models & GPT-5.1 & 0.790 & 0.710 & 0.081 & $<$0.001 \\
 & Gemini~3~Pro & 0.790 & 0.704 & 0.086 & $<$0.001 \\
 & DeepSeek-V3.2 & 0.764 & 0.689 & 0.075 & $<$0.001 \\
Metrics & GPT-5.1 & 0.711 & 0.451 & 0.261 & $<$0.001 \\
 & Gemini~3~Pro & 0.693 & 0.427 & 0.266 & $<$0.001 \\
 & DeepSeek-V3.2 & 0.699 & 0.464 & 0.234 & $<$0.001 \\
\midrule
\addlinespace
\multicolumn{6}{l}{\textit{Test 2: Inter-paper entity Jaccard decomposition}} \\
\midrule
Entity type & Source & $J_{\mathrm{actual}}$ & $J_{\mathrm{random}}$ & \multicolumn{2}{c}{$J_{\mathrm{excess}}$ [95\% CI]} \\
\midrule
Datasets & Reference & 0.0014 & 0.0022 & \multicolumn{2}{c}{-0.0008} \\
 & GPT-5.1 & 0.0061 & 0.0059 & \multicolumn{2}{c}{0.0002} \\
 & $\Delta_{\mathrm{excess}}$ (GPT-5.1) & -- & -- & \multicolumn{2}{c}{0.0010 [0.0001, 0.0019]} \\
\addlinespace
 & Reference & 0.0016 & 0.0022 & \multicolumn{2}{c}{-0.0006} \\
 & Gemini~3~Pro & 0.0054 & 0.0061 & \multicolumn{2}{c}{-0.0007} \\
 & $\Delta_{\mathrm{excess}}$ (Gemini~3~Pro) & -- & -- & \multicolumn{2}{c}{-0.0000 [-0.0008, 0.0008]} \\
\addlinespace
 & Reference & 0.0014 & 0.0022 & \multicolumn{2}{c}{-0.0008} \\
 & DeepSeek-V3.2 & 0.0049 & 0.0056 & \multicolumn{2}{c}{-0.0007} \\
 & $\Delta_{\mathrm{excess}}$ (DeepSeek-V3.2) & -- & -- & \multicolumn{2}{c}{0.0001 [-0.0007, 0.0010]} \\
\addlinespace
Models & Reference & 0.0079 & 0.0059 & \multicolumn{2}{c}{0.0019} \\
 & GPT-5.1 & 0.0454 & 0.0399 & \multicolumn{2}{c}{0.0055} \\
 & $\Delta_{\mathrm{excess}}$ (GPT-5.1) & -- & -- & \multicolumn{2}{c}{0.0036 [0.0018, 0.0055]} \\
\addlinespace
 & Reference & 0.0080 & 0.0060 & \multicolumn{2}{c}{0.0020} \\
 & Gemini~3~Pro & 0.0686 & 0.0542 & \multicolumn{2}{c}{0.0144} \\
 & $\Delta_{\mathrm{excess}}$ (Gemini~3~Pro) & -- & -- & \multicolumn{2}{c}{0.0124 [0.0097, 0.0150]} \\
\addlinespace
 & Reference & 0.0080 & 0.0060 & \multicolumn{2}{c}{0.0020} \\
 & DeepSeek-V3.2 & 0.0783 & 0.0641 & \multicolumn{2}{c}{0.0142} \\
 & $\Delta_{\mathrm{excess}}$ (DeepSeek-V3.2) & -- & -- & \multicolumn{2}{c}{0.0122 [0.0092, 0.0152]} \\
\addlinespace
Metrics & Reference & 0.0293 & 0.0163 & \multicolumn{2}{c}{0.0130} \\
 & GPT-5.1 & 0.0367 & 0.0324 & \multicolumn{2}{c}{0.0043} \\
 & $\Delta_{\mathrm{excess}}$ (GPT-5.1) & -- & -- & \multicolumn{2}{c}{-0.0087 [-0.0107, -0.0067]} \\
\addlinespace
 & Reference & 0.0274 & 0.0160 & \multicolumn{2}{c}{0.0114} \\
 & Gemini~3~Pro & 0.0196 & 0.0188 & \multicolumn{2}{c}{0.0008} \\
 & $\Delta_{\mathrm{excess}}$ (Gemini~3~Pro) & -- & -- & \multicolumn{2}{c}{-0.0106 [-0.0123, -0.0089]} \\
\addlinespace
 & Reference & 0.0279 & 0.0159 & \multicolumn{2}{c}{0.0120} \\
 & DeepSeek-V3.2 & 0.0725 & 0.0520 & \multicolumn{2}{c}{0.0205} \\
 & $\Delta_{\mathrm{excess}}$ (DeepSeek-V3.2) & -- & -- & \multicolumn{2}{c}{0.0085 [0.0058, 0.0113]} \\
\addlinespace
\bottomrule
\end{tabular}
\end{adjustbox}
\end{table}

\begin{table}[H]
\centering
\caption{\textbf{Provider concentration persists after removing singleton entities, confirming it is not an artefact of rare paper-specific references.} These analyses use the all-three-LLM shared-paper subsets: datasets n$=878$, models n$=915$, metrics n$=911$. ``All'' retains the full reference-inventory entity set; ``Excl.\ singletons'' removes entities appearing in exactly one paper. Provider JSD is reported only for models (the only entity type with a provider taxonomy dimension). Zero-coverage rate denominator = papers with $\geq$1 reference-inventory entity after filtering. Title-match and combined filters are omitted as they remove 0.3--3.3\% of entities and produce negligible changes. $n_\text{ref}$ denotes the number of unique normalised reference-inventory entity names in the shared-paper subset retained under each filter.}
\label{tab:established_robustness}
\begin{adjustbox}{max width=\textwidth}
\begin{tabular}{ll r rrr rrr rrr}
\toprule
& & & \multicolumn{3}{c}{Corpus-level recall (\%)} & \multicolumn{3}{c}{Zero-coverage rate (\%)} & \multicolumn{3}{c}{Provider JSD} \\
\cmidrule(lr){4-6} \cmidrule(lr){7-9} \cmidrule(lr){10-12}
Entity & Filter & $n_\text{ref}$ & GPT-5.1 & Gemini~3~Pro & DeepSeek-V3.2 & GPT-5.1 & Gemini~3~Pro & DeepSeek-V3.2 & GPT-5.1 & Gemini~3~Pro & DeepSeek-V3.2 \\
\midrule
Dataset  & All & 2{,}504 & 15.9 & 16.2 & 12.7 & 64.9 & 63.1 & 71.1 & \multicolumn{3}{c}{---} \\
         & Excl.\ singletons & 354 & 42.7 & 43.2 & 35.9 & 54.1 & 51.6 & 63.7 & \multicolumn{3}{c}{---} \\
\addlinespace
Model    & All & 3{,}323 & 6.8 & 7.9 & 5.1 & 65.1 & 65.2 & 83.5 & 0.111 & 0.102 & 0.137 \\
         & Excl.\ singletons & 700 & 20.7 & 24.3 & 16.9 & 66.6 & 67.1 & 85.2 & 0.092 & 0.090 & 0.140 \\
\addlinespace
Metric   & All & 2{,}392 & 14.1 & 13.1 & 12.2 & 44.2 & 51.7 & 42.7 & \multicolumn{3}{c}{---} \\
         & Excl.\ singletons & 465 & 42.6 & 42.2 & 40.2 & 43.3 & 50.5 & 40.8 & \multicolumn{3}{c}{---} \\
\bottomrule
\end{tabular}
\end{adjustbox}
\end{table}

\begin{table}[H]
\centering
\caption{\textbf{Under a broader provider regrouping, the deficit concentrates in singleton-defined long-tail models while reused academic/community models are modestly overrepresented.} Computed on the all-three-LLM shared model-paper intersection (n$=915$). Share (\%) shows the proportion of model mentions falling in each broad provider category; excess is the percentage-point difference (LLM $-$ Reference). JSD (original) uses the fine-grained provider taxonomy; JSD (broader grouping) uses the four-category taxonomy.}
\label{tab:provider_level1}
\begin{adjustbox}{max width=\textwidth}
\begin{tabular}{l rr rr rr rr rr}
\toprule
& \multicolumn{2}{c}{Major commercial} & \multicolumn{2}{c}{Other commercial} & \multicolumn{2}{c}{Academic/Community} & \multicolumn{2}{c}{Other/Acad.\ singleton} & \multicolumn{2}{c}{JSD} \\
\cmidrule(lr){2-3} \cmidrule(lr){4-5} \cmidrule(lr){6-7} \cmidrule(lr){8-9} \cmidrule(lr){10-11}
Source & Share & Excess & Share & Excess & Share & Excess & Share & Excess & Original & Broad \\
\midrule
Reference & 54.9 & --- & 2.9 & --- & 14.5 & --- & 27.7 & --- & --- & --- \\
\addlinespace
GPT-5.1 & 74.3 & $+$19.4 & 4.1 & $+$1.2 & 18.0 & $+$3.6 & 3.5 & $-$24.1 & 0.111 & 0.089 \\
Gemini~3~Pro & 71.8 & $+$16.9 & 4.0 & $+$1.1 & 20.1 & $+$5.6 & 4.1 & $-$23.5 & 0.102 & 0.082 \\
DeepSeek-V3.2 & 73.8 & $+$18.9 & 2.5 & $-$0.4 & 20.2 & $+$5.8 & 3.4 & $-$24.3 & 0.137 & 0.092 \\
\bottomrule
\end{tabular}
\end{adjustbox}
\end{table}

\begin{table}[H]
\centering
\caption{\textbf{Exact-name model recall is low, but family- and provider-level matching recovers most of the apparent miss.} Mean per-paper recall at three matching granularities. Exact: normalised entity names must match. Family: model names mapped to families via regex (e.g., all Llama variants $\to$ ``llama''). Provider (dedup): only the provider label must match, deduplicated per paper. These per-paper figures are not comparable to the corpus-level recall in \Cref{tab:established_robustness}, which counts unique entities across the entire corpus. $n = 915$ shared papers (papers where the paper-derived reference inventory and all three LLMs contain model entities after normalisation).}
\label{tab:family_matching}
\begin{adjustbox}{max width=\textwidth}
\begin{tabular}{l rr rr rr}
\toprule
& \multicolumn{2}{c}{Exact} & \multicolumn{2}{c}{Family} & \multicolumn{2}{c}{Provider (dedup)} \\
\cmidrule(lr){2-3} \cmidrule(lr){4-5} \cmidrule(lr){6-7}
LLM & Mean (\%) & Median (\%) & Mean (\%) & Median (\%) & Mean (\%) & Median (\%) \\
\midrule
GPT-5.1 & 6.2 & 0.0 & 28.2 & 23.1 & 53.1 & 50.0 \\
Gemini~3~Pro & 6.5 & 0.0 & 27.9 & 23.7 & 53.3 & 50.0 \\
DeepSeek-V3.2 & 3.7 & 0.0 & 20.0 & 12.5 & 44.9 & 40.0 \\
\bottomrule
\end{tabular}
\end{adjustbox}
\end{table}

\begin{table}[H]
\centering
\caption{\textbf{Conditional-on-consensus extraction diagnostics; not corpus-level or full-audit accuracy.} Precision = correctly extracted / total extracted; recall = correctly extracted / (correctly extracted $+$ missed); hallucination rate = hallucinated / total extracted. Precision, recall, F1, and hallucination rate are mean per-paper values over consensus rows only (53 of 90 audited rows; overall extraction consensus 58.9\%). The 100\% precision/recall/F1 therefore describe the majority-selected subsample, not the full audit. ICC(2,1) is computed on the full pairwise-complete count series between Claude Opus 4.6 and GPT-5.4.}
\label{tab:annotation_extraction}
\begin{adjustbox}{max width=\textwidth}
\begin{tabular}{l r rrrr r}
\toprule
Entity type & $n_{\mathrm{cons.}}$ & Precision (\%) & Recall (\%) & F1 (\%) & Halluc.\ rate (\%) & ICC(2,1) \\
\midrule
Datasets & 23 & 100.0 & 100.0 & 100.0 & 0.0 & 0.746 \\
Models & 13 & 100.0 & 100.0 & 100.0 & 0.0 & 0.304 \\
Metrics & 17 & 100.0 & 100.0 & 100.0 & 0.0 & 0.661 \\
\textbf{Overall} & 53 & 100.0 & 100.0 & 100.0 & 0.0 & 0.469 \\
\bottomrule
\end{tabular}
\end{adjustbox}
\end{table}

\begingroup
\footnotesize
\setlength{\LTcapwidth}{\textwidth}

\begin{longtable}{l l l r r r r}
\caption{\textbf{Blinded cross-model audit: classification validation.} Exact-match accuracy requires set equality for multi-label dimensions after alphabetical canonicalisation. Accuracy is computed on consensus rows only; $\kappa$ is computed on all pairwise-complete rows. Source separates paper-derived reference-inventory entities (Reference) from LLM-suggested entities pooled across GPT-5.1, Gemini 3 Pro, and DeepSeek-V3.2 (LLM). The four-tier validation assignments (\Cref{sec:annotation_validation}) are based on the Reference and pooled-LLM (LLM) strata, which provide the largest sample sizes and the cleanest source separation; the per-model strata (DeepSeek-V3.2, GPT-5.1, Gemini~3~Pro) are supplementary descriptive checks with smaller samples and should not be used to override tier assignments.}
\label{tab:annotation_classification} \\
\toprule
Entity type & Source & Dimension & $n_{\mathrm{cons.}}$ & Accuracy (\%) & $\kappa$ & $n_{\mathrm{pair.}}$ \\
\midrule
\endfirsthead

\caption[]{\textbf{Blinded cross-model audit: classification validation.} (continued)} \\
\toprule
Entity type & Source & Dimension & $n_{\mathrm{cons.}}$ & Accuracy (\%) & $\kappa$ & $n_{\mathrm{pair.}}$ \\
\midrule
\endhead

\midrule
\multicolumn{7}{r}{\emph{Continued on next page}} \\
\endfoot

\bottomrule
\endlastfoot

Dataset & Reference & Modality & 14 & 85.7 & 0.756 & 17 \\
 &  & Task type & 10 & 10.0 & 0.518 & 18 \\
 &  & Domain & 7 & 14.3 & 0.329 & 18 \\
 &  & Annotation & 8 & 0.0 & 0.710 & 10 \\
 &  & Size & 10 & 90.0 & 0.642 & 13 \\
 &  & Granularity & 9 & 77.8 & 0.367 & 16 \\
 &  & Linguistic scope & 9 & 88.9 & 1.000 & 9 \\
 &  & Cognitive/affective & 4 & 50.0 & 0.600 & 6 \\
 &  & Data quality & 16 & 100.0 & 0.000 & 18 \\
 & LLM & Modality & 35 & 85.7 & 0.654 & 44 \\
 &  & Task type & 25 & 32.0 & 0.520 & 44 \\
 &  & Domain & 29 & 51.7 & 0.592 & 45 \\
 &  & Annotation & 18 & 16.7 & 0.747 & 21 \\
 &  & Size & 16 & 87.5 & 0.406 & 24 \\
 &  & Granularity & 24 & 75.0 & 0.254 & 37 \\
 &  & Linguistic scope & 20 & 85.0 & 0.660 & 23 \\
 &  & Cognitive/affective & 1 & 100.0 & 0.250 & 3 \\
 &  & Data quality & 39 & 97.4 & 0.495 & 44 \\
 & DeepSeek-V3.2 & Modality & 11 & 90.9 & 0.577 & 16 \\
 &  & Task type & 8 & 50.0 & 0.434 & 16 \\
 &  & Domain & 13 & 53.8 & 0.768 & 16 \\
 &  & Annotation & 7 & 14.3 & 1.000 & 7 \\
 &  & Size & 4 & 75.0 & 0.667 & 5 \\
 &  & Granularity & 7 & 71.4 & -0.100 & 11 \\
 &  & Linguistic scope & 6 & 66.7 & 1.000 & 6 \\
 &  & Cognitive/affective & 1 & 100.0 & 1.000 & 1 \\
 &  & Data quality & 15 & 93.3 & 0.636 & 16 \\
 & GPT-5.1 & Modality & 15 & 86.7 & 0.709 & 17 \\
 &  & Task type & 13 & 30.8 & 0.691 & 18 \\
 &  & Domain & 9 & 55.6 & 0.417 & 18 \\
 &  & Annotation & 7 & 14.3 & 0.561 & 9 \\
 &  & Size & 7 & 85.7 & 0.421 & 11 \\
 &  & Granularity & 11 & 72.7 & 0.365 & 16 \\
 &  & Linguistic scope & 7 & 85.7 & 0.591 & 9 \\
 &  & Data quality & 15 & 100.0 & 0.000 & 18 \\
 & Gemini~3~Pro & Modality & 9 & 77.8 & 0.667 & 11 \\
 &  & Task type & 4 & 0.0 & 0.318 & 10 \\
 &  & Domain & 7 & 42.9 & 0.569 & 11 \\
 &  & Annotation & 4 & 25.0 & 0.667 & 5 \\
 &  & Size & 5 & 100.0 & 0.000 & 8 \\
 &  & Granularity & 6 & 83.3 & 0.310 & 10 \\
 &  & Linguistic scope & 7 & 100.0 & 0.704 & 8 \\
 &  & Data quality & 9 & 100.0 & 0.737 & 10 \\
\addlinespace
Model & Reference & Architecture & 16 & 25.0 & 0.813 & 17 \\
 &  & Training paradigm & 10 & 10.0 & 0.339 & 17 \\
 &  & Provider & 17 & 94.1 & 1.000 & 17 \\
 &  & Openness & 17 & 94.1 & 1.000 & 17 \\
 &  & Size & 8 & 87.5 & 1.000 & 8 \\
 & LLM & Architecture & 32 & 40.6 & 1.000 & 32 \\
 &  & Training paradigm & 20 & 10.0 & 0.329 & 32 \\
 &  & Provider & 29 & 89.7 & 0.923 & 31 \\
 &  & Openness & 31 & 96.8 & 1.000 & 31 \\
 &  & Size & 17 & 47.1 & 0.885 & 18 \\
 & DeepSeek-V3.2 & Architecture & 10 & 20.0 & 1.000 & 10 \\
 &  & Training paradigm & 7 & 14.3 & 0.516 & 10 \\
 &  & Provider & 9 & 66.7 & 1.000 & 9 \\
 &  & Openness & 9 & 100.0 & 1.000 & 9 \\
 &  & Size & 5 & 60.0 & 1.000 & 5 \\
 & GPT-5.1 & Architecture & 12 & 58.3 & 1.000 & 12 \\
 &  & Training paradigm & 7 & 0.0 & 0.268 & 12 \\
 &  & Provider & 11 & 100.0 & 0.893 & 12 \\
 &  & Openness & 12 & 91.7 & 1.000 & 12 \\
 &  & Size & 8 & 62.5 & 1.000 & 8 \\
 & Gemini~3~Pro & Architecture & 10 & 40.0 & 1.000 & 10 \\
 &  & Training paradigm & 6 & 16.7 & 0.130 & 10 \\
 &  & Provider & 9 & 100.0 & 0.865 & 10 \\
 &  & Openness & 10 & 100.0 & 1.000 & 10 \\
 &  & Size & 4 & 0.0 & 0.000 & 5 \\
\addlinespace
Metric & Reference & Evaluation type & 7 & 85.7 & 0.061 & 17 \\
 & LLM & Evaluation type & 29 & 89.7 & 0.588 & 42 \\
 & DeepSeek-V3.2 & Evaluation type & 10 & 100.0 & 0.762 & 12 \\
 & GPT-5.1 & Evaluation type & 7 & 71.4 & 0.198 & 18 \\
 & Gemini~3~Pro & Evaluation type & 12 & 91.7 & 1.000 & 12 \\

\end{longtable}
\endgroup

\begin{table}[H]
\centering
\caption{\textbf{Blinded cross-model audit: normalisation validation.} Merge precision is the fraction of merged pairs that both systems judge should merge; near-miss error rate is the fraction of 80--89 similarity pairs that both systems judge should merge. The overall $\kappa$ is computed on all 100 pairwise-complete rows; value percentages use consensus rows only.}
\label{tab:annotation_normalization}
\begin{adjustbox}{max width=\textwidth}
\begin{tabular}{l r rr l}
\toprule
Stratum & $n_{\mathrm{cons.}}$ & Value (\%) & $\kappa$ & Interpretation \\
\midrule
Merged (score >= 90) & 50 & 76.0 & -- & Merge precision \\
Near-miss (score 80-89) & 45 & 6.7 & -- & Error rate (should have merged) \\
\textbf{Overall} & 95 & 84.2 & 0.898 & Decision accuracy \\
\bottomrule
\end{tabular}
\end{adjustbox}
\end{table}

\begin{table}[H]
\centering
\caption{\textbf{Blinded cross-model audit: introducedness label distribution.} Each entity--paper pair is classified as pre-existing reusable, paper-introduced, paper-specific derivative, or unclear by Claude Opus 4.6 and GPT-5.4. Percentages are computed on consensus rows only; $\kappa$ uses all pairwise-complete rows within each entity type. The sample is quota-based across entity types and frequency bands, so percentages are sample-level estimates rather than corpus-weighted distributions.}
\label{tab:annotation_introducedness}
\begin{adjustbox}{max width=\textwidth}
\begin{tabular}{l r rrrr r}
\toprule
Entity type & $n_{\mathrm{cons.}}$ & Pre-existing (\%) & Paper-introduced (\%) & Paper-specific deriv.\ (\%) & Unclear (\%) & $\kappa$ \\
\midrule
Dataset & 81 & 97.5 & 2.5 & 0.0 & 0.0 & 0.143 \\
Model & 103 & 97.1 & 1.9 & 1.0 & 0.0 & 0.205 \\
Metric & 54 & 98.1 & 1.9 & 0.0 & 0.0 & -0.004 \\
\textbf{Overall} & 238 & 97.5 & 2.1 & 0.4 & 0.0 & 0.108 \\
\bottomrule
\end{tabular}
\end{adjustbox}
\end{table}

\begin{table}[H]
\centering
\caption{\textbf{Heuristic calibration against audit-confirmed introducedness labels.} Precision is the fraction of heuristic-flagged entities that are audit-confirmed paper-specific (paper-introduced or paper-specific derivative); recall is the fraction of audit-confirmed paper-specific entities that the heuristic flags; specificity is the fraction of audit-confirmed pre-existing entities correctly not flagged. Effective evaluation counts use consensus introducedness rows only. Because no consensus introducedness row remained labelled unclear, the conservative and liberal analyses coincide numerically.}
\label{tab:annotation_heuristic_calibration}
\begin{adjustbox}{max width=\textwidth}
\begin{tabular}{l l r rrrr}
\toprule
Heuristic & Assumption & $n_{\mathrm{cons.}}$ & Precision (\%) & Recall (\%) & Specificity (\%) & F1 (\%) \\
\midrule
Singleton filter & Conservative & 238 & 7.5 & 100.0 & 68.1 & 14.0 \\
Singleton filter & Liberal & 238 & 7.5 & 100.0 & 68.1 & 14.0 \\
\addlinespace
Title-match filter & Conservative & 238 & 100.0 & 16.7 & 100.0 & 28.6 \\
Title-match filter & Liberal & 238 & 100.0 & 16.7 & 100.0 & 28.6 \\
\addlinespace
Combined filter & Conservative & 238 & 100.0 & 16.7 & 100.0 & 28.6 \\
Combined filter & Liberal & 238 & 100.0 & 16.7 & 100.0 & 28.6 \\
\bottomrule
\end{tabular}
\end{adjustbox}
\end{table}

\clearpage
\section{Paper-side entity extraction prompt}
\label{sec:prompt_extraction}

The following system prompt was sent to GPT-5.1 via the OpenAI Batch API for each of the 1,000 papers. The model received the paper's title, abstract, and full text as user input under the non-public scientific TDM and API data-handling procedure described in \Cref{sec:copyright_tdm,sec:api_data_handling}; the batch input/output file objects under our control were deleted after processing.

\begin{promptbox}[System prompt]
You are an academic assistant. Given the title, abstract, and full text of a paper:\par
\medskip
1. Generate a single concise research question.\par
2. Extract:\par
\hspace*{1em}- datasets used\par
\hspace*{1em}- models used\par
\hspace*{1em}- evaluation metrics used\par
\medskip
Respond in JSON with:\par
\{\par
\hspace*{1em}"research\_question": "...",\par
\hspace*{1em}"GroundTruthDatasets": ["..."],\par
\hspace*{1em}"GroundTruthModels": ["..."],\par
\hspace*{1em}"GroundTruthMetrics": ["..."]\par
\}\par
\medskip
Only report the datasets, models, and metrics used in the experiments and not from the literature review or related work sections.\par
Each dataset, model and evaluation metric name must be composed from one to three words tops.
\end{promptbox}

This is the requested schema. In the saved raw batch outputs, some responses instead returned generic \texttt{datasets/models/metrics} keys; before analysis, the parsing step harmonized both forms into the standardized \texttt{GroundTruth...} JSON fields used by the analysis pipeline (CSV headers and internal schema). The narrative and visible table labels use ``paper-derived reference inventory''/``Reference'' throughout.

\section{LLM suggestion prompt}
\label{sec:prompt_suggestion}

The following system prompt was sent identically to GPT-5.1, Gemini~3~Pro, and DeepSeek-V3.2, with only the JSON key prefixes varying per model (e.g., \texttt{GPT51\_suggested\_dataset}, \texttt{gemini\_suggested\_dataset}, \texttt{deepseek\_suggested\_dataset}). Each model received only the research question extracted from the paper; Gemini~3~Pro and DeepSeek-V3.2 did not receive the paper PDF, abstract, or extracted full text.

\begin{promptbox}[System prompt]
You are an expert AI research assistant.\par
Given the following research question:\par
\medskip
"\{research\_question\}"\par
\medskip
Please suggest one or more:\par
1. suitable datasets to address the question.\par
2. appropriate machine learning models, architectures, or Large Language Models to use.\par
3. relevant evaluation metrics for measuring the model's performance.\par
4. A straightforward pipeline or methodology for solving it.\par
\medskip
Only respond with one or more specific dataset names, one or more specific model names, one or more specific evaluation metric names and keep the pipeline structured and short.\par
Each dataset, model and evaluation metric name must be composed from one to three words tops.\par
In the pipeline, explain how you want to run the experiment to solve the research question in bullet points.\par
\medskip
Respond in valid JSON with keys:\par
\{\par
\hspace*{1em}"[LLM]\_suggested\_dataset": ["..."],\par
\hspace*{1em}"[LLM]\_suggested\_model": ["..."],\par
\hspace*{1em}"[LLM]\_suggested\_evaluation\_metric": ["..."],\par
\hspace*{1em}"[LLM]\_suggested\_pipeline": "..."\par
\}
\end{promptbox}

\clearpage
\section{Taxonomy classification schema}
\label{sec:prompt_taxonomy}

All paper-derived reference-inventory and LLM-suggested entities were classified by GPT-5.1 via the OpenAI Batch API using the following taxonomy schema. This classification step operated on extracted entity names and, where applicable, generated pipeline text, rather than on the original PDFs or full extracted paper texts. For each entity, the classifier was instructed to assign one or more values from each applicable dimension and return the result in structured JSON. The schema below lists the intended allowed values for each dimension.

\begin{enumerate}
    \item \textbf{Dataset dimensions.} For each dataset, classify under:
    \begin{itemize}
        \item \textbf{Modalities:} Text, Audio, Image, Video, Time series, Graph, Spatial, Multimodal.
        \item \textbf{Task types:} Classification, Regression, Sequence labeling, Generation, Summarization, Translation, Question answering, Reasoning, Dialogue, Object detection, Forecasting, Retrieval, Alignment, Multimodal integration, Clustering, Reinforcement learning.\footnote{In practice, the classifier occasionally assigned labels from adjacent dimensions (e.g., \emph{Decision Making} and \emph{Problem Solving} from the cognitive/affective schema; \emph{Safety}, \emph{Ranking}, and \emph{Evaluation} from metric evaluation types; \emph{Segmentation} adjacent to object detection) to the task-type field in $\leq$0.4\% of assignments per source. A single off-schema modality label (\emph{Vision}, 0.04\% of reference-inventory modality assignments) also appears. These were retained in the analysis rather than discarded, which is why some appendix figures show categories beyond the lists above.}
        \item \textbf{Domains:} General, Media, Scientific/academic, Healthcare, Legal, Economics, Social, Geospatial, Robotics, Vision, Entertainment, Education, Infrastructure, Ontology, Biology, Chemistry, Environmental.
        \item \textbf{Annotation:} Fully supervised, Weakly supervised, Self-supervised, Semi-supervised, Reinforcement feedback, Crowdsourced, Expert annotations.
        \item \textbf{Size:} Small ($<$10K items), Medium (10K--100K items), Large ($>$100K items).
        \item \textbf{Granularity:} Document-level, Sentence-level, Token-level, Frame-level, Pixel-level, Object-level.
        \item \textbf{Linguistic scope:} Monolingual (with language specification from: English, Chinese, Spanish, French, German, Russian, Portuguese, Italian, Dutch, Arabic, Japanese, Korean, Turkish, Polish, Vietnamese, Indonesian, Hebrew, Swedish, Czech, Hungarian, Other), Multilingual, Cross-lingual.
        \item \textbf{Cognitive/affective:} Attention, Memory, Problem solving, Reasoning, Decision making, Perception, Learning, Cognitive load, Emotion, Empathy, Theory of mind, Social reasoning, Moral cognition, Personality.
        \item \textbf{Data quality:} Noisy, Curated.
    \end{itemize}

    \item \textbf{Model dimensions.} For each model, classify under:
    \begin{itemize}
        \item \textbf{Architecture:} Transformer (Encoder/Decoder/Encoder--Decoder), Generative, CNN, RNN/LSTM, GNN, Tree-based, Linear, Kernel models, Probabilistic, Reinforcement learning.\footnote{In figures and text, ``Transformer (all)'' aggregates the three Transformer subtypes (Encoder, Decoder, Encoder--Decoder) into a single category.}
        \item \textbf{Training paradigm:} Supervised learning, Self-supervised learning, Unsupervised learning, Reinforcement learning, Multi-task learning, Few-shot learning, Zero-shot learning, Fine-tuning, RAG.
        \item \textbf{Provider:} OpenAI, Anthropic, Meta~AI, Google DeepMind, Mistral~AI, Alibaba/Qwen, Cohere, Hugging Face, Stability~AI, Microsoft Research, NVIDIA/NeMo, Databricks/MosaicML, DeepSeek, Other/Academic.
        \item \textbf{Openness:} Closed, Open.
        \item \textbf{Size:} Small ($<$1B), Medium (1--10B), Large (10--100B), Extra-large ($>$100B parameters).
    \end{itemize}

    \item \textbf{Metric dimensions.} For each evaluation metric, classify under:
    \begin{itemize}
        \item \textbf{Evaluation type:} Accuracy, Ranking, Regression, Continuous prediction, Probability, Uncertainty, Fairness, Safety, Efficiency/latency, Explainability, Robustness, User experience.
    \end{itemize}
\end{enumerate}

\end{document}